\documentclass[twoside,11pt]{article}

\usepackage{jmlr2e}

\usepackage{hyperref}       

\usepackage{url}            
\usepackage{booktabs}       
\usepackage{amsfonts}       
\usepackage{nicefrac}       %
\usepackage{microtype}      %
\usepackage{subfigure}
\usepackage{booktabs}      
\usepackage{amsmath}
\usepackage{arydshln}
\usepackage{xcolor}
\usepackage[utf8]{inputenc}
\usepackage{url}
\usepackage[demo]{rotating}
\usepackage{amssymb}
\usepackage{rotating}
\usepackage{mathrsfs}
\usepackage{lastpage}
\usepackage{footnote}
\makesavenoteenv{tabular}
\makesavenoteenv{table}

\newcommand\Tstrut{\rule{0pt}{2.6ex}}         
\newcommand\Bstrut{\rule[-0.9ex]{0pt}{0pt}}   

\newcommand{\graph}[1]{\ensuremath{\mathcal{#1}}}
\newcommand{\vertices}[1]{\ensuremath{\mathcal{#1}}}
\newcommand{\edges}[1]{\ensuremath{\mathcal{#1}}}
\newcommand{\vertex}[1]{\ensuremath{\mathsf{#1}}}

\newcommand{\relations}[1]{\ensuremath{\mathcal{#1}}}
\newcommand{\relation}[1]{\ensuremath{\mathsf{#1}}}
\newcommand{\events}[1]{\ensuremath{\mathcal{#1}}}
\newcommand{\embedding}[1]{\ensuremath{\mathtt{#1}}}
\newcommand{\function}[1]{\ensuremath{\mathtt{#1}}}
\newcommand{\model}[1]{\ensuremath{\mathcal{#1}}}
\newcommand{\cdf}[1]{\ensuremath{\mathcal{#1}}}
\newcommand{\history}[1]{\ensuremath{\mathcal{#1}}}
\newcommand{\Th}[1]{\ensuremath{#1^{\text{th}}}}
\newcommand{\transpose}[1]{\ensuremath{#1'}}
\newcommand{\neighbour}{\ensuremath{\mathcal{N}}}
\newcommand{\horizon}[1]{\ensuremath{#1}}

\newcommand*{\savedfootnotes}{}
\newcommand*{\resetsavedfootnotes}{\global\let\savedfootnotes\empty}

\DeclareMathOperator*{\E}{\mathbb{E}}

\newcommand{\vctr}[1]{\ensuremath{\mathbf{#1}}}
\newcommand{\mtrx}[1]{\ensuremath{\mathbf{#1}}}
\newcommand{\imtrx}[1]{\ensuremath{\mathbf{I_{#1}}}}

\begin{document}

\jmlrheading{21}{2020}{1-\pageref{LastPage}}{6/19; Revised 2/20}{3/20}{19-447}{Seyed Mehran Kazemi, Rishab Goel, Kshitij Jain, Ivan Kobyzev, Akshay Sethi,  Peter Forsyth,  Pascal Poupart} \ShortHeadings{Representation Learning for Dynamic Graphs: A Survey}{Kazemi, Goel, Jain, Kobyzev, Sethi, Forsyth, Poupart}


\ShortHeadings{Representation Learning for Dynamic Graphs}{Kazemi, Goel, Jain, Kobyzev, Sethi, Forsyth, and Poupart}
\firstpageno{1}

\title{Representation Learning for Dynamic Graphs: A Survey}

\author{\name Seyed Mehran Kazemi \email mehran.kazemi@borealisai.com \\
       \name Rishab Goel \email rishab.goel@borealisai.com \\
       \addr Borealis AI, 310-6666 Saint Urbain, Montreal, QC, Canada \\
    \AND
       \name Kshitij Jain \email kshitij.jain@borealisai.com\\
       \name Ivan Kobyzev \email ivan.kobyzev@borealisai.com\\
       \name Akshay Sethi \email akshay.sethi@borealisai.com\\
       \name Peter Forsyth \email peter.forsyth@borealisai.com\\
       \name Pascal Poupart \email pascal.poupart@borealisai.com\\
       \addr Borealis AI, 301-420 West Graham Way, Waterloo, ON, Canada\\
       }

\editor{Karsten Borgwardt}

\maketitle

\begin{abstract}
Graphs arise naturally in many real-world applications including social networks, recommender systems, ontologies, biology, and computational finance.  Traditionally, machine learning models for graphs have been mostly designed for static graphs. However, many applications involve evolving graphs.  This introduces important challenges for learning and inference since nodes, attributes, and edges change over time. In this survey, we review the recent advances in representation learning for dynamic graphs, including dynamic knowledge graphs. We describe existing models from an encoder-decoder perspective, categorize these encoders and decoders based on the techniques they employ, and analyze the approaches in each category. We also review several prominent applications and widely used datasets and highlight directions for future research.
\end{abstract}


\section{Introduction}
In the era of big data, a challenge is to leverage data as effectively as possible to extract patterns, make predictions, and more generally unlock value.  In many situations, the data does not consist only of vectors of features, but also relations that form graphs among entities. Graphs arise naturally in social networks (users with friendship relations, emails, text messages), recommender systems (users and products with transactions and rating relations), ontologies (concepts with relations), computational biology (protein-protein interactions), computational finance (web of companies with competitor, customer, subsidiary relations, supply chain graph, graph of customer-merchant transactions), etc. While it is often possible to ignore relations and use traditional machine learning techniques based on vectors of features, relations provide additional valuable information that permits inference among nodes. Hence, graph-based techniques have emerged as leading approaches in the industry for application domains with relational information.

Traditionally, research has been done mostly on static graphs where nodes and edges are fixed and do not change over time. Many applications, however, involve dynamic graphs.  For instance, in social media, communication events such as emails and text messages are streaming while friendship relations evolve.  In recommender systems, new products, new users and new ratings appear every day.  In computational finance, transactions are streaming and supply chain relations are continuously evolving.  As a result, the last few years have seen a surge of works on dynamic graphs.  This survey focuses precisely on dynamic graphs.  Note that there are already many good surveys on static graphs (see, e.g., \cite{hamilton2017representation,zhang2018network,cai2018comprehensive,cui2018survey,nickel2016review,shi2016survey,wang2017knowledge}). There are also several surveys on techniques for dynamic graphs (see, e.g., \cite{bilgin2006dynamic,zhang2010survey,spiliopoulou2011evolution,aggarwal2014evolutionary,al2011survey}), but they do not review recent advances in neural representation learning.

We present a survey that focuses on recent representation learning techniques for dynamic graphs.  More precisely, we focus on reviewing techniques that either produce time-dependent embeddings that capture the essence of the nodes and edges of evolving graphs or use embeddings to answer various questions such as node classification, event prediction/interpolation, and link prediction.  Accordingly, we use an encoder-decoder framework to categorize and analyze techniques that encode various aspects of graphs into embeddings and other techniques that decode embeddings into predictions.  We survey techniques that deal with discrete- and/or continuous-time events. 

The survey is structured as follows.  Section~\ref{sec:background} introduces the notation and provides some background about static/dynamic graphs, inference tasks, and learning techniques. Section~\ref{sec:static-graphs} provides an overview of representation learning techniques for static graphs.  This section is not meant to be a survey, but rather to introduce important concepts that will be extended for dynamic graphs. Section~\ref{sec:dynamic-encoders} describes encoding techniques that aggregate temporal observations and static features, use time as a regularizer, perform decompositions, traverse dynamic networks with random walks, and model observation sequences with various types of processes (e.g., recurrent neural networks).
Section~\ref{sec:dynamic-decoders} categorizes decoders for dynamic graphs into time-predicting and time-conditioned decoders and surveys the decoders in each category.  Section~\ref{sec:other-models} describes briefly other lines of work that do not conform to the encoder-decoder framework such as statistical relational learning, and topics related to dynamic (knowledge) graphs such as spatiotemporal graphs and the construction of dynamic knowledge graphs from text.  Section~\ref{sec:datasets} reviews important applications of dynamic graphs with representative tasks.  A list of static and temporal datasets is also provided with a summary of their properties. Section~\ref{sec:conclusion} concludes the survey with a discussion of several open problems and possible research directions. 

\section{Background and Notation}
\label{sec:background}
In this section, we define our notation and provide the necessary background for readers to follow the rest of the survey. A summary of the main notation and abbreviations can be found in Table~\ref{tab:notation}.

We use lower-case letters to denote scalars, bold lower-case letters to denote vectors, and bold upper-case letters to denote matrices. For a vector \vctr{z}, we represent the $\Th{i}$ element of the vector as $\vctr{z}[i]$. For a matrix $\mtrx{A}$, we represent the $\Th{i}$ row of $\mtrx{A}$ as $\mtrx{A}[i]$, and the element at the $\Th{i}$ row and $\Th{j}$ column as $\mtrx{A}[i][j]$. $|| \vctr{z} ||_i$ represents norm $i$ of a vector $\vctr{z}$ and $|| \mtrx{Z} ||_F$ represents the Frobenius norm of a matrix $\mtrx{Z}$. For two vectors $\vctr{z_1}\in\mathbb{R}^{d_1}$ and $\vctr{z_2}\in\mathbb{R}^{d_2}$, we use $[\vctr{z}_1;\vctr{z}_2]\in\mathbb{R}^{d_1+d_2}$ to represent the concatenation of the two vectors. When $d_1=d_2=d$, we use $[\vctr{z}_1~\vctr{z}_2]\in\mathbb{R}^{d\times 2}$ to represent a $d\times 2$ matrix whose two columns correspond to $\vctr{z}_1$ and $\vctr{z}_2$ respectively. We use $\odot$ to represent element-wise (Hadamard) multiplication. We represent by $\imtrx{d}$ the identity matrix of size $d\times d$. 
$\function{vec}(\mtrx{A})$ vectorizes $\mtrx{A}\in\mathbb{R}^{d_1\times d_2}$ into a vector of size $d_1d_2$. $\function{diag}(\vctr{z})$ turns $\vctr{z}\in\mathbb{R}^d$ into a diagonal matrix of size $d\times d$ that has the values of $\vctr{z}$ on its main diagonal. We denote the transpose of a matrix $\mtrx{A}$ as $\transpose{\mtrx{A}}$.

\subsection{Static Graphs}
\label{defs_static}
A \emph{(static) graph} is represented as $\graph{G}=(\vertices{V}, \edges{E})$ where $\vertices{V}=\{\vertex{v}_1, \vertex{v}_2, \dots, \vertex{v}_{|\vertices{V}|}\}$ is the set of vertices and $\edges{E}\subseteq \vertices{V}\times\vertices{V}$ is the set of edges. Vertices are also called \emph{nodes} and we use the two terms interchangeably. Edges are also called \emph{links} and we use the two terms interchangeably.

Several matrices can be associated with a graph. An \emph{adjacency matrix} $\mtrx{A}\in\mathbb{R}^{|\vertices{V}|\times |\vertices{V}|}$ is a matrix where $\mtrx{A}[i][j]=0$ if $(\vertex{v}_i, \vertex{v}_j)\not\in \edges{E}$; otherwise $\mtrx{A}[i][j]\in \mathbb{R}_{+}$ represents the weight of the edge. For unweighted graphs, all non-zero $\mtrx{A}[i][j]$s are $1$. A \emph{degree matrix} $\mtrx{D}\in\mathbb{R}^{|\vertices{V}|\times |\vertices{V}|}$ is a diagonal matrix where $\mtrx{D}[i][i] = \sum_{j=1}^{|\vertices{V}|}\mtrx{A}[i][j]$ represents the degree of $\vertex{v}_i$. A \emph{graph Laplacian} is defined as $\mtrx{L} = \mtrx{D} - \mtrx{A}$.

A graph is \emph{undirected} if the order of the nodes in the edges is \emph{not} important.
For an undirected graph, the adjacency matrix is symmetric, i.e. $\mtrx{A}[i][j]=\mtrx{A}[j][i]$ for all $i$ and $j$ (in other words, $\mtrx{A}=\transpose{\mtrx{A}}$). 
A graph is \emph{directed} if the order of the nodes in the edges is important. 
Directed graphs are also called \emph{digraphs}. For an edge $(\vertex{v}_i, \vertex{v}_j)$ in a digraph, we call $\vertex{v}_i$ the \emph{source} and $\vertex{v}_j$ the \emph{target} of the edge. 
A graph is \emph{bipartite} if the nodes can be split into two groups where there is no edge between any pair of nodes in the same group. A \emph{multigraph} is a graph where multiple edges can exist between two nodes. A graph is \emph{attributed} if each node is associated with some properties representing its characteristics. For a node $\vertex{v}$ in an attributed graph, we let $\vctr{x}_\vertex{v}$ represent the attribute values of $\vertex{v}$. When all nodes have the same attributes, we represent all attribute values of the nodes by a matrix $\mtrx{X}$ whose $\Th{i}$ row corresponds to the attribute values of $\vertex{v}_i$.

A \emph{knowledge graph (KG)} corresponds to a multi-digraph with labeled edges, where the labels represent the types of the relationships. Let $\relations{R}=\{\relation{r_1}, \relation{r_2}, \dots, \relation{r_{|\relations{R}|}}\}$ be a set of relation types. Then $\edges{E}\subseteq \vertices{V}\times\relations{R}\times\vertices{V}$. That is, each edge is a triple of the form $(source, relation, target)$. A KG can be attributed in which case each node $\vertex{v}\in\vertices{V}$ is associated with a vector $\vctr{x}_\vertex{v}$ of attribute values. 
A digraph is a special case of a KG with only one relation type. An undirected graph is a special case of a KG with only one symmetric relation type.

Closely related to KGs are heterogeneous information networks. A \emph{heterogeneous information network} (HIN) is typically defined as a digraph $\graph{G}=(\vertices{V}, \edges{E})$ with two additional functions: one mapping each node $\vertex{v}\in\vertices{V}$ to a node type and one mapping each edge $(\vertex{v}_i, \vertex{v}_j)\in\edges{E}$ to an edge type (\cite{shi2016survey,sun2013mining}). Compared to KGs, HINs define node types explicitly using a mapping function whereas KGs typically define node types using triples, e.g., $(\vertex{Zootopia}, \relation{type}, \vertex{Movie})$. Moreover, the definition of HINs implies the possibility of only one edge between two nodes whereas KGs allow multiple edges with different labels. However, other definitions have been considered for HINs which allow multiple edges between two entities as well (see, e.g., \cite{yang2012predicting}). Despite slight differences in definition, the terms KG and HIN have been used interchangeably in some works (see, e.g., \cite{nickel2016review}). In this work, we mainly adopt the term KG.

\begin{figure*}[t]
  \centering
\includegraphics[width=\textwidth]{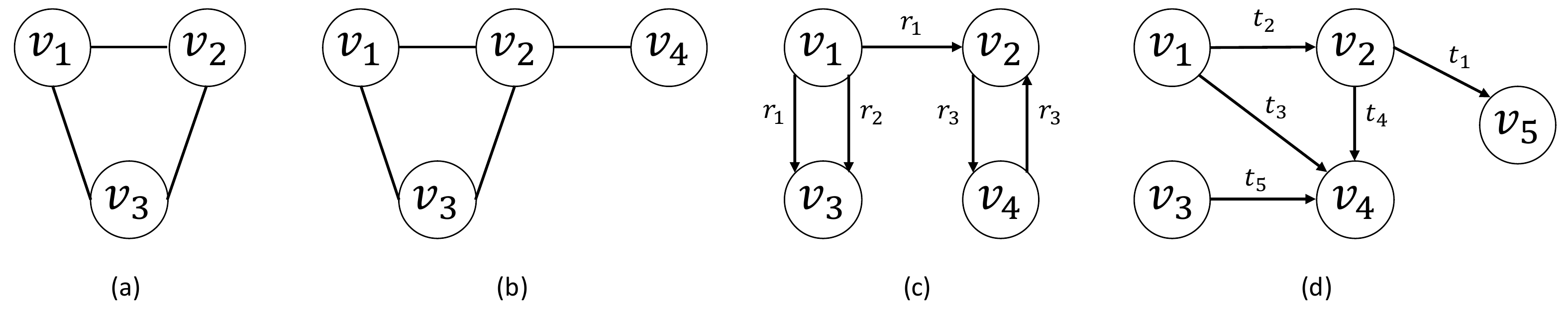}
  \caption{Four graphs to be used as running examples throughout the survey. (a) and (b) are two examples of undirected graphs. They can be also considered as two snapshots of a discrete-time dynamic graph. (c) is an example of a knowledge graph. (d) is an example of a continuous-time dynamic graph where the only possible event/observation is edge addition.}
  \label{graphs-fig}
\end{figure*}

\begin{example}
Figure~\ref{graphs-fig}(a) represents an undirected graph with three nodes $\vertex{v}_1$, $\vertex{v}_2$ and $\vertex{v}_3$ and three edges $(\vertex{v}_1, \vertex{v}_2)$, $(\vertex{v}_1, \vertex{v}_3)$ and $(\vertex{v}_2, \vertex{v}_3)$. Figure~\ref{graphs-fig}(b) represents a graph with four nodes and four edges. The adjacency, degree, and Laplacian matrices for the graph in Figure~\ref{graphs-fig}(b) are as follows:
\[
\mtrx{A}
=
\begin{bmatrix}
    0 & 1 & 1 & 0 \\
    1 & 0 & 1 & 1 \\
    1 & 1 & 0 & 0 \\
    0 & 1 & 0 & 0
\end{bmatrix}
\quad
\mtrx{D}
=
\begin{bmatrix}
    2 & 0 & 0 & 0 \\
    0 & 3 & 0 & 0 \\
    0 & 0 & 2 & 0 \\
    0 & 0 & 0 & 1
\end{bmatrix}
\quad
\mtrx{L}
=
\begin{bmatrix}
    ~2 & -1 & -1 & ~0 \\
    -1 & ~3 & -1 & -1 \\
    -1 & -1 & ~2 & ~0 \\
    ~0 & -1 & ~0 & ~1
\end{bmatrix}
\]

where the $\Th{i}$ row (and the $\Th{i}$ column) corresponds to $\vertex{v}_i$. Since the graph is undirected, $\mtrx{A}$ is symmetric. Figure~\ref{graphs-fig}(c) represents a KG with four nodes $\vertex{v}_1$, $\vertex{v}_2$, $\vertex{v}_3$ and $\vertex{v}_4$, three relation types $\relation{r}_1$, $\relation{r}_2$, and $\relation{r}_3$, and five labeled edges as follows:
\begin{align*}
    (\vertex{v}_1, \relation{r}_1, \vertex{v}_2) \quad
    (\vertex{v}_1, \relation{r}_1, \vertex{v}_3) \quad
    (\vertex{v}_1, \relation{r}_2, \vertex{v}_3) \quad
    (\vertex{v}_2, \relation{r}_3, \vertex{v}_4) \quad
    (\vertex{v}_4, \relation{r}_3, \vertex{v}_2)
\end{align*}
The KG in Figure~\ref{graphs-fig}(c) is directed and is a multigraph as there are, e.g., two edges (with the same direction) between $\vertex{v}_1$ and $\vertex{v}_3$.
\end{example}

\subsection{Dynamic Graphs}
We represent a \emph{continuous-time dynamic graph (CTDG)} as a pair $(\graph{G},\events{O})$ where $\graph{G}$ is a static graph representing an initial state of a dynamic graph at time $t_0$ and \events{O} is a set of observations/events where each observation is a tuple of the form $(event~type, event, timestamp)$. An event type can be an edge addition, edge deletion, node addition, node deletion, node splitting, node merging, etc. At any point $t\geq t_0$ in time, a snapshot $\graph{G}^t$ (corresponding to a static graph) can be obtained from a CTDG by updating $\graph{G}$ sequentially according to the observations $\events{O}$ that occurred before (or at) time $t$ (sometimes, the update may require aggregation to handle multiple edges between two nodes).

A \emph{discrete-time dynamic graph (DTDG)} is a sequence of snapshots from a dynamic graph sampled at regularly-spaced times. Formally, we define a DTDG as a set $\{\graph{G}^1, \graph{G}^2, \dots, \graph{G}^T\}$ where $\graph{G}^t=\{\vertices{V}^t,\edges{E}^t\}$ is the graph at snapshot $t$, $\vertices{V}^t$ is the set of nodes in $\graph{G}^t$, and $\edges{E}^t$ is the set of edges in $\graph{G}^t$. We use the term \emph{dynamic graph} to refer to both DTDGs and CTDGs. Compared to a CTDG, a DTDG may lose information by looking only at some snapshots of the graph over time, but developing models for DTDGs may be generally easier. In particular, a model developed for CTDGs may be used for DTDGs, but the reverse is not necessarily true.

An \emph{undirected dynamic graph} is a dynamic graph where at any time $t$, $\graph{G}^t$ is an undirected graph.
A \emph{directed dynamic graph} is a dynamic graph where at any time $t$, $\graph{G}^t$ is a digraph.
A \emph{bipartite dynamic graph} is a dynamic graph where at any time $t$, $\graph{G}^t$ is a bipartite graph.
A \emph{dynamic KG} is a dynamic graph where at any time $t$, $\graph{G}^t$ is a KG.

\begin{example}
Consider a CTDG as $(\graph{G}, \events{O})$ where $\graph{G}$ is a graph with five nodes $\vertex{v}_1$, $\vertex{v}_2$, $\vertex{v}_3$, $\vertex{v}_4$ and $\vertex{v}_5$ and with no edges between any pairs of nodes, and \events{O} is:
\begin{align*}
    \{
    (AddEdge, (\vertex{v}_2,&~ \vertex{v}_5), t_1),
    (AddEdge, (\vertex{v}_1, \vertex{v}_2), t_2),
    (AddEdge, (\vertex{v}_1, \vertex{v}_4), t_3),\\
    &~(AddEdge, (\vertex{v}_2, \vertex{v}_4), t_4),
    (AddEdge, (\vertex{v}_3, \vertex{v}_4), t_5)
    \}
\end{align*}
This CTDG may be represented graphically as in Figure~\ref{graphs-fig}(d). The only type of observation in this CTDG is the addition of new edges. The second element of each observation corresponding to an edge addition represents the source and the target nodes of the new edge. The third element of each observation represents the timestamp at which the observation was made.
\end{example}
\begin{example}
Consider an undirected CTDG whose initial state is as in Figure~\ref{graphs-fig}(a). Suppose $\events{O}$ is:
\begin{align*}
    \{(AddNode,\vertex{v}_4,~t_1),
    (AddEdge,(\vertex{v}_2,\vertex{v}_4),~t_2)\}
\end{align*}
where $t_1 \leq t_2$. Now consider a DTDG that takes two snapshots from this CTDG, one snapshot at time $t_0$ and one snapshot at time $t_2$. The two snapshots of this DTDG look like the graphs in Figure~\ref{graphs-fig}(a) and Figure~\ref{graphs-fig}(b) respectively.
\end{example}

\begin{table}[t]
    \centering
    \footnotesize
    \begin{tabular}{c|l}
        Symbols and abbreviations & Meaning \\ \hline
        DTDG, CTDG & Discrete-Time and Continuous-Time Dynamic Graph \\
        KG & Knowledge Graph \\
        \graph{G}, \vertices{V}, \edges{E} & Graph, nodes, and edges. \\
        \mtrx{A}, \mtrx{L}, \mtrx{D}, \mtrx{X} & Adjacency, Laplacian, degree, and attribute matrices of a graph \\
        $\events{O}$ & Set of observations for a CTDG \\
        $\mtrx{W}$ & Matrix of learnable weights \\
        $\graph{G}^t, \vertices{V}^t, \edges{E}^t, \mtrx{A}^t$ & Graph, nodes, edges, and adjacency matrix at time $t$. \\
        $\vertex{v}, \vertex{u}$ & Two generic nodes in a graph. \\
        $T$ & The number of snapshots in a DTDG \\
        $\embedding{EMB}$ & The embedding function \\
        $[\vctr{z}_1;\vctr{z}_2]$ & Concatenation of two vectors $\vctr{z}_1$ and $\vctr{z}_2$ \\
        $\phi, \sigma$ & A generic and the Sigmoid activation function \\
        \function{vec}(.) & Vectorized view of the input matrix or tensor \\
        $|| \vctr{z} ||_i$, $|| \mtrx{Z} ||_F$ & Norm $i$ of $\vctr{z}$, and Frobenius norm of $\mtrx{Z}$. \\
        $\transpose{\mtrx{A}}, \transpose{\vctr{z}}$ & Transpose of a matrix and a vector
    \end{tabular}
    \caption{Summary of the main notation and abbreviations.}
    \label{tab:notation}
\end{table}

\subsection{Prediction Problems} \label{sec:prediction-problems}
In this survey, we mainly study three general problems for dynamic graphs: \emph{node classification}, \emph{edge prediction}, and \emph{graph classification}. Node classification is the problem of classifying each node into one class from a set of predefined classes. Link prediction is the problem of predicting new links between the nodes. Graph classification is the problem of classifying a whole graph into one class from a set of predefined classes. A high-level description of some other prediction problems can be found in Section~\ref{subsec: application}.

Reasoning over dynamic graphs typically falls under two settings: \emph{interpolation} and \emph{extrapolation}. Consider a dynamic graph that has incomplete information from the time interval $[t_0, t_T]$. The \emph{interpolation} problem is to make predictions at some time $t$ such that $t_0 \leq t \leq t_T$. The interpolation problem is also known as the \emph{completion} problem and is mainly used for completing (dynamic) KGs (\cite{jiang2016towards,leblay2018deriving,garcia2018learning,dasgupta2018hyte,goel2020diachronic}). An example of the interpolation problem is to predict which country won the world cup in 2002 assuming this information is missing in the KG.
The \emph{extrapolation} problem is to make predictions at time $t$ such that $t \geq t_T$, i.e., predicting future based on the past. Extrapolation is usually a more challenging problem than the interpolation problem. An example of the extrapolation problem is predicting which country will win the next world cup. 

\paragraph{Streaming scenario:} In the streaming scenario, new observations are being streamed to the model at a fast rate and the model needs to update itself based on these observations in real-time so it can make informed predictions immediately after each observation arrives. For this scenario, a model may not have enough time to retrain completely or in part when new observations arrive. Streaming scenarios are often best handled by CTDGs and often give rise to extrapolation problems.

\subsection{The Encoder-Decoder Framework} \label{sec:enc-dec-framework}
Following \citet{hamilton2017representation}, to deal with the large notational and methodological diversity of the existing approaches and to put the various methods on an equal notational and conceptual footing, we develop an encoder-decoder framework for dynamic graphs. Before describing the encoder-decoder framework, we define one of the main components in this framework known as \emph{embedding}.

\begin{definition} \label{embedding-dfnt}
An \emph{embedding} is a function that maps every node $\vertex{v}\in\vertices{V}$ of a graph, and every relation type $\relation{r}\in\relations{R}$ in case of a KG, to a hidden representation where the hidden representation is typically a tuple of one or more scalars, vectors, and/or matrices of numbers. The vectors and matrices in the tuple are supposed to contain the necessary information about the nodes and relations to enable making predictions about them.
\end{definition}

For each node $\vertex{v}$ and relation $\relation{r}$, we refer to the hidden representation of $\vertex{v}$ and $\relation{r}$ as the embedding of $\vertex{v}$ and the embedding of $\relation{r}$ respectively. When the main goal is link prediction, some works define the embedding function as mapping each pair of nodes into a hidden representation. In these cases, we refer to the hidden representation of a pair $(\vertex{v}, \vertex{u})$ of nodes as the embedding of the pair $(\vertex{v}, \vertex{u})$.

Having the above definition, we can now formally define an encoder and a decoder.

\begin{definition} \label{dfnt:encoder}
An \emph{encoder} takes as input a dynamic graph and outputs an embedding function that maps nodes, and relations in case of a KG, to hidden representations.
\end{definition}

\begin{definition}
A \emph{decoder} takes as input an embedding function and makes predictions (such as node classification, edge prediction, etc.) based on the embedding function.
\end{definition}

In many cases (e.g., \cite{kipf2017semi,hamilton2017inductive,yang2015embedding,bordes2013translating,nickel2016holographic,dong2014knowledge}), the embedding function $\embedding{EMB}(.)$ maps each node, and each relation in the case of a KG, to a tuple containing a single vector; that is $\embedding{EMB}(\vertex{v})=(\vctr{z}_\vertex{v})$ where $\vctr{z}_\vertex{v}\in\mathbb{R}^{d_1}$ and $\embedding{EMB}(\relation{r})=(\vctr{z}_\relation{r})$ where $\vctr{z}_\relation{r}\in\mathbb{R}^{d_2}$. Other works consider different representations. For instance, \citet{kazemi2018simple} define $\function{EMB}(\vertex{v})=(\vctr{z}_\vertex{v},\overline{\vctr{z}}_\vertex{v})$ and $\function{EMB}(\relation{r})=(\vctr{z}_\relation{r},\overline{\vctr{z}}_\relation{r})$, i.e. mapping each node and each relation to two vectors where each vector has a different usage. \citet{StransE} define $\function{EMB}(\vertex{v})=(\vctr{z}_\vertex{v})$ and $\function{EMB}(\relation{r})=(\vctr{z}_\vertex{r}, \mtrx{P}_\relation{r},\mtrx{Q}_\relation{r})$, i.e. mapping each node to a single vector but mapping each relation to a vector and two matrices. We will describe these approaches (and many others) in the upcoming sections.

A \emph{model} corresponds to an encoder-decoder pair. One of the benefits of describing models in an encoder-decoder framework is that it allows for creating new models by combining the encoder from one model with the decoder from another model when the hidden representations produced by the encoder conform to the hidden representations consumed by the decoder.

\subsubsection{Training}
For many choices of an encoder-decoder pair, it is possible to train the two components end-to-end. In such cases, the parameters of the encoder and the decoder are typically initialized randomly. Then, until some criterion is met, several epochs of stochastic gradient descent are performed where in each epoch, the embedding function is produced by the encoder, predictions are made based on the embedding function by the decoder, the error in predictions is computed with respect to a loss function, and the parameters of the model are updated based on the loss. 

For node classification and graph classification, the loss function can be any classification loss (e.g., cross-entropy loss). For link prediction, typically one only has access to positive examples corresponding to the links already in the graph. A common approach in such cases is to generate a set of negative samples where negative samples correspond to edges that are believed to have a low probability of being in the graph. Then, having a set of positive and a set of negative samples, the training of a link predictor turns into a classification problem and any classification loss can be used. The choice of the loss function depends on the application.

\subsection{Expressivity}
The expressivity of the models for (dynamic) graphs can be thought of as the diversity of the graphs they can represent. Depending on the problem at hand (e.g., node classification, link prediction, graph classification, etc.), the expressivity can be defined differently. We first provide some intuition on the importance of expressivity using the following example.

\begin{example} \label{expressivity-example}
Consider a model $\model{M}$ for binary classification (with labels $True$ and $False$) in KGs. Suppose the encoder of $\model{M}$ maps every node to a tuple containing a single scalar representing the number of incoming edges to that node (regardless of the labels of the edges). For the KG in Figure~\ref{graphs-fig}(c), for instance, this encoder will output an embedding function as:
\begin{align*}
    \embedding{EMB}(\vertex{v}_1) = (0) \quad\quad\quad
    \embedding{EMB}(\vertex{v}_2) = (2) \quad\quad\quad
    \embedding{EMB}(\vertex{v}_3) = (2) \quad\quad\quad
    \embedding{EMB}(\vertex{v}_4) = (1)
\end{align*}
No matter what the decoder of $\model{M}$ is, 
Since $\embedding{EMB}(\vertex{v}_2)$ and $\embedding{EMB}(\vertex{v}_3)$ are identical, any deterministic decoder will assign the same class to $\vertex{v}_2$ and $\vertex{v}_3$. Therefore, no matter what decoder $\model{M}$ uses, $\model{M}$ is not expressive enough to assign different classes to $\vertex{v}_2$ and $\vertex{v}_3$.
\end{example}

From Example~\ref{expressivity-example}, we can see why the expressivity of a model may be important. A model that is not expressive enough is doomed to underfitting. Expressivity of the representation learning models for graphs has been the focus of several studies. It has been studied from different perspectives and for different classes of models. \citet{xu2019powerful}, \citet{morris2019weisfeiler}, \citet{Maron2019OnTU}, \citet{Keriven2019UniversalIA} and \citet{Chen2019OnTE} study the expressivity of a class of models called graph convolutional networks (see Section~\ref{sec:gcn}). \citet{kazemi2018simple}, \citet{trouillon2017knowledge}, \citet{fatemi2019improved}, \citet{balavzevic2019tucker} and several other works provide expressivity results for models operating on KGs (see Section~\ref{sec:static-kg-decoders}). \citet{goel2020diachronic} provide expressivity results for their model developed for temporal KGs (see Section~\ref{sec:time-conditioned-decoders}). We will refer to several of these works in the next sections when describing different (classes of) models.

In what follows, we provide general definitions for the expressivity of representation learning models for graphs. Before giving the definitions, we describe symmetric nodes. Two nodes in a graph are \emph{symmetric} if there exists the same information about them (i.e. they have the same attribute values and the same neighbors). Recall that a model corresponds to an encoder-decoder pair. 

\begin{definition} \label{dfnt:full-exp-classify}
A model $\model{M}$ with parameters $\Theta$ is fully expressive with respect to node classification if given any graph $\graph{G}=(\vertices{V}, \edges{E})$ and any function $\Omega: \vertices{V} \rightarrow \mathcal{C}$ mapping nodes to classes (where symmetric nodes are mapped to the same class), there exists an instantiation of $\Theta$ such that $\model{M}$ classifies the nodes in $\vertices{V}$ according to $\Omega$.
\end{definition}

\begin{example}
Consider a model $\model{M}$ for binary classification (with labels $True$ and $False$) whose encoder is the one introduced in Example~\ref{expressivity-example} and whose decoder is a logistic regression model. It can be verified that the encoder has no parameters so model parameters $\Theta$ correspond to the parameters of the decoder (i.e. the logistic regression).
We disprove the full expressivity of $\model{M}$ using a counterexample.
According to Definition~\ref{dfnt:full-exp-classify}, a counterexample corresponds to a a pair $(\graph{G}_{ce}, \Omega_{ce}$) (where $ce$ stands for \emph{counterexample}) of a specific graph and a specific function $\Omega_{ce}$ such that there exists no instantiation of $\Theta$ that classifies the nodes of $\graph{G}_{ce}$ according to $\Omega_{ce}$. Let $\graph{G}_{ce}$ be the graph in Figure~\ref{graphs-fig}(c) and let $\Omega_{ce}$ be a mapping function defined as: $\Omega_{ce}(\vertex{v}_1)=True$, $\Omega_{ce}(\vertex{v}_2)=True$, 
$\Omega_{ce}(\vertex{v}_3)=False$, and
$\Omega_{ce}(\vertex{v}_4)=True$. From Example~\ref{expressivity-example}, we know that the encoder gives the same embedding for $\vertex{v}_2$ and $\vertex{v}_3$ so there cannot exist an instantiation of $\Theta$ which classifies $\vertex{v}_2$ as $True$ and $\vertex{v}_3$ as $False$. Hence, the above pair is a counterexample.
\end{example}

A similar definition can be given for the full expressivity of a model with respect to link prediction and graph classification.

\begin{definition} \label{dfnt:full-exp-link}
A model $\model{M}$ with parameters $\Theta$ is fully expressive with respect to link prediction if given any graph $\graph{G}=(\vertices{V}, \edges{E})$ and any function $\Omega:\edges{E}\rightarrow \{True, False\}$ indicating the existence or non-existence of (labeled) edges for all node-pairs in the graph, there exists an instantiation of $\Theta$ such that $\model{M}$ classifies the edges in $\edges{E}$ according to $\Omega$.
\end{definition}

\begin{definition} \label{dfnt:full-exp-iso}
A model $\model{M}$ with parameters $\Theta$ is fully expressive with respect to graph classification if given any set $\mathcal{S}=\{\graph{G}_1, \graph{G}_2, \dots, \graph{G}_n \}$ of non-isomorphic graphs and any function $\Omega: \mathcal{S}\rightarrow \mathcal{C}$ mapping graphs to classes, there exists an instantiation of $\Theta$ such that $\model{M}$ classifies the graphs according to $\Omega$.
\end{definition}

\subsection{Sequence Models} \label{sec:sequence-models}

In dynamic environments, data often consists of sequences of observations of varying lengths.  There is a long history of models to handle sequential data without a fixed length. This includes auto-regressive models~(\cite{akaike1969fitting}) that predict the next observations based on a window of past observations.  Alternatively, since it is not always clear how long the window of part observations should be, hidden Markov models~(\cite{rabiner1986introduction}), Kalman filters~(\cite{welch1995introduction}), dynamic Bayesian networks~(\cite{murphy2002dynamic}) and dynamic conditional random fields~(\cite{sutton2007dynamic}) use hidden states to capture relevant information that might be arbitrarily far in the past.  Today, those models can be seen as special cases of recurrent neural networks, which allow rich and complex hidden dynamics. 

Recurrent neural networks (RNNs) (\cite{elman1990finding,cho2014learning}) have achieved impressive results on a range of sequence modeling problems.
The core principle of the RNN is that its input is a function of the current data point as well as the history of the previous inputs. A simple RNN model can be formulated as follows:
\begin{align}
    \vctr{h}^t = \phi(\mtrx{W}_i \vctr{x}^t + \mtrx{W}_h \vctr{h}^{t-1} + \vctr{b}_i)
\end{align}
where $\vctr{x}^t\in\mathbb{R}^{d_{in}}$ is the input at position $t$ in the sequence, $\vctr{h}^{t-1}\in\mathbb{R}^d$ is a hidden representation containing information about the sequence of inputs until time $t-1$, $\mtrx{W}_i\in\mathbb{R}^{d\times d_{in}}$ and $\mtrx{W}_h\in\mathbb{R}^{d\times d}$ are weight matrices, $\vctr{b}_i\in\mathbb{R}^d$ represents the vector of biases, $\phi$ is an activation function, and $\vctr{h}^t\in\mathbb{R}^d$ is an updated hidden representation containing information about the sequence of inputs until time $t$. We use $\vctr{h}^t = \function{RNN}(\vctr{h}^{t-1}, \vctr{x}^t)$ to represent the output of an RNN operation on a previous state $\vctr{h}^{t-1}$ and a new input $\vctr{x}^t$.

Long short term memory (LSTM)~(\cite{hochreiter1997long}) is considered one of the most successful RNN architectures.
The original LSTM model can be neatly defined with the following equations:
\begin{align}
&\vctr{i}^t= \sigma\left(\mtrx{W}_{ii}
\vctr{x}^t+\mtrx{W}_{ih}\vctr{h}^{t-1}+\vctr{b}_i\right) \label{lstm-eq-i}\\
&\vctr{f}^t= \sigma\left(\mtrx{W}_{fi}\vctr{x}^t+\mtrx{W}_{fh}\vctr{h}^{t-1}+\vctr{b}_f\right)\label{lstm-eq-f}\\
&\vctr{c}^t = \vctr{f}^t\odot\vctr{c}^{t-1} + \vctr{i}^t\odot \function{Tanh}\left(\mtrx{W}_{ci} \vctr{x}^t+\mtrx{W}_{ch} \vctr{h}^{t-1}+\vctr{b}_c\right) \label{lstm-eq-c}\\
&\vctr{o}^t= \sigma\left(\mtrx{W}_{oi} \vctr{x}^t+\mtrx{W}_{oh} \vctr{h}^{t-1}+\vctr{b}_o\right)\label{lstm-eq-o}\\
&\vctr{h}^t= \vctr{o}^t\odot\function{Tanh}\left(\vctr{c}^t\right)\label{lstm-eq-h}
\end{align}
Here $\vctr{i}^t$, $\vctr{f}^t$, and $\vctr{o}^t$ represent the input, forget and output gates respectively, while $\vctr{c}^t$ is the memory cell and $\vctr{h}^t$ is the hidden state. $\sigma$ and $\function{Tanh}$ represent the Sigmoid and hyperbolic tangent activation functions respectively. Gated recurrent units (GRUs)~(\cite{cho2014learning}) is another successful RNN architecture. In the context of dynamic graphs, sequence models such as LSTMs and GRUs can be used to, e.g., provide node representations based on the history of the node (see Sections~\ref{sec:rnn-dtdg}~and~\ref{sec:rnn-encoders-ctdg}).

Fully attentive models have recently demonstrated on-par or superior performance compared to RNN variants for a variety of tasks (see, e.g., \cite{vaswani2017attention,dehghani2018universal,krantz2018abstractive,shaw2018self}). These models rely only on (self-)attention and abstain from using recurrence. Let $\mtrx{X_{in}}\in\mathbb{R}^{T\times d}$ represent a sequence containing $T$ elements each with $d$ features. The idea behind a self-attention layer is to update each row of $\mtrx{X_{in}}$ by allowing it to attend to itself and all other rows. For this purpose,  \citet{vaswani2017attention} first create $\mtrx{\bar{X}_{in}}=\mtrx{X_{in}}+\mtrx{P}$ where $\mtrx{P}\in\mathbb{R}^{T\times d}$ is called the \emph{positional encoding} matrix and carries information about the position of each element in the sequence. Then, they
project the matrix $\mtrx{\bar{X}_{in}}$ into a matrix $\mtrx{Q}=\mtrx{\bar{X}_{in}}\mtrx{W}_{Q}\in\mathbb{R}^{T\times d_k}$ dubbed \emph{queries matrix}, a matrix $\mtrx{K}=\mtrx{\bar{X}_{in}}\mtrx{W}_{K}\in\mathbb{R}^{T\times d_k}$ dubbed \emph{keys matrix}, and a matrix $\mtrx{V}=\mtrx{\bar{X}_{in}}\mtrx{W}_{V}\in\mathbb{R}^{T\times d_v}$ dubbed \emph{values matrix}, where $\mtrx{W}_Q, \mtrx{W}_K\in\mathbb{R}^{d\times d_k}$ and $\mtrx{W}_V\in\mathbb{R}^{d\times d_v}$ are matrices with learnable parameters. Then, each row of the matrix $\mtrx{\bar{X}_{in}}$ is updated by taking a weighted sum of the rows in $\mtrx{V}$. The weights are computed using the query and key matrices. The updated matrix $\mtrx{X_{out}}\in\mathbb{R}^{T\times d_v}$ is computed as follows:
\begin{align}
    \mtrx{X_{out}}=\function{Attention}(\mtrx{Q},\mtrx{K},\mtrx{V}) = \function{softmax}(\frac{\mtrx{Q}\mtrx{K}'}{\sqrt{d_k}})\mtrx{V} \label{eq:scaled_attention}
\end{align}
where $\function{softmax}$ performs a row-wise normalization of the input matrix and $\function{softmax}(\frac{\mtrx{Q}\mtrx{K}'}{\sqrt{d_k}})$ gives the weights. A mask can be added to Equation~\eqref{eq:scaled_attention} to make sure that at time $t$, the mechanism only allows a sequence model to attend to the points before time $t$. \citet{vaswani2017attention} also define a \emph{multi-head} self-attention mechanism by considering multiple self-attention blocks (as defined in Equation~\eqref{eq:scaled_attention}) each having different weight matrices and then concatenating the results. In the context of static graphs, the initial $\mtrx{X_{in}}$ may correspond to the representations of the neighbors of a node, and in the context of dynamic graphs where node representations keep evolving, the initial $\mtrx{X_{in}}$ may correspond to a node's representations at different points in time (see Sections~\ref{sec:gcn}~and~\ref{sec:attention-encoder}).

\subsection{Temporal Point Processes} \label{sec:tpp}
Temporal point processes (TPPs)~(\cite{cox1972multivariate}) are stochastic processes which are used for modeling sequential asynchronous discrete events occurring in continuous time.
A typical realization of a TPP is a sequence of discrete events occurring at time points $t_1, t_2, t_3, \dots$ for $t_{i}\leq\horizon{T}$, where $\horizon{T}$ represents the time horizon of the process. A TPP is generally characterized using a conditional intensity function $\lambda(t)$ such that $\lambda(t)dt$ represents the probability of an event happening in the interval $[t, t+dt]$ given the history $t_1, \dots, t_n$ of the process and given that no event occurred until $t_n < t \leq T$. The conditional density function $\function{f}(t)$, indicating the density of the occurrence of the next event at some time point $t_n < t \leq \horizon{T}$, can be obtained as $\function{f}(t)=\lambda(t)\function{S}(t)$. Here, $\function{S}(t)=\exp\big(-\int_{t_n}^{t}\lambda(\tau)d\tau\big)$, called the \emph{survival function} of the process, is the probability that no event happens during $[t_n, t)$. The time for the next event can be predicted by taking an expectation over $\function{f}(t)$.

Traditionally, intensity functions were hand-designed to model how future/present events depend on the past events in the TPP. 
Some of the well-known TPPs include Hawkes process~(\cite{hawkes1971spectra,mei2017neural}), Poisson processes~(\cite{kingman2005p}), self-correcting processes~(\cite{isham1979self}), and autoregressive conditional duration processes~(\cite{engle1998autoregressive}). Depending on the application, one may use the intensity function in one of these TPPs or design new ones. Recently, there has been growing interest in learning the intensity function entirely from the data (see, e.g., \cite{du2016recurrent}). In the context of dynamic graphs, a TPP with an intensity function parameterized by the node representations in the graph can be constructed to predict when something will happen to a single node or to a pair of nodes (see Section~\ref{sec:time-pred-decoders}).

\section{Representation Learning for Static Graphs}
\label{sec:static-graphs}
In this section, we provide an overview of representation learning approaches for static graphs. The main purpose of this section is to provide enough information for the descriptions and discussions in the next sections on dynamic graphs. Readers interested in learning more about representation learning on static graphs can refer to several existing surveys specifically written on this topic (e.g., see \cite{hamilton2017representation,zhang2018network,cai2018comprehensive,cui2018survey} for graphs and \cite{nickel2016review,wang2017knowledge} for KGs).

\subsection{Encoders} \label{sec:static-encoders}
As described in Subsection~\ref{sec:enc-dec-framework}, a model can be viewed as a combination of an encoder and a decoder. In this section, we describe different approaches for creating encoders.

\subsubsection{High-Order Proximity Matrices}\label{sec:high-order-encoder}
While the adjacency matrix of a graph only represents local proximities, one can also define \emph{high-order proximity} matrices~(\cite{Ou2016AsymmetricTP}) also known as \emph{graph similarity metrics}~(\cite{da2012time}). Let $\mtrx{S}$ be a high-order proximity matrix. A simple approach for creating an encoder is to let $\embedding{EMB}(\vertex{v}_i)=(\mtrx{S}[i])$ (or $\embedding{EMB}(\vertex{v}_i)=(\transpose{\mtrx{S}}[i])$) corresponding to the $\Th{i}$ row (or the $\Th{i}$ column) of matrix $\mtrx{S}$. Encoders based on high-order proximity matrices are typically parameter-free and do not require learning (although some of them have hyper-parameters that need to be tuned). In what follows, we describe several of these matrices.

\emph{Common neighbors} matrix is defined as $\mtrx{S}_{CN} = \mtrx{A}\mtrx{A}$. $\mtrx{S}_{CN}[i][j]$ corresponds to the number of nodes that are connected to both $\vertex{v}_i$ and $\vertex{v}_j$. For a directed graph, $\mtrx{S}_{CN}[i][j]$ counts how many nodes $\vertex{v}$ are simultaneously the target of an edge starting at $\vertex{v}_i$ and the source of an edge ending at $\vertex{v}_j$.

\emph{Jaccard's coefficient} is a slight modification of $\mtrx{S}_{CN}$ where one divides the number of common neighbors of $\vertex{v}_i$ and $\vertex{v}_j$ by the total number of distinct nodes that are the targets of edges starting at  $\vertex{v}_i$ or the sources of edges ending at $\vertex{v}_j$. Formally, Jaccard's coefficient is defined as $\mtrx{S}_{JC}[i][j] =  \mtrx{S}_{CN}[i][j]/(\sum_{k=1}^{|\vertices{V}|} (\mtrx{A}[i][k]+\mtrx{A}[k][j]) - \mtrx{S}_{CN}[i][j])$.

\emph{Adamic-Adar} is defined as $\mtrx{S}_{AA} = \mtrx{A}\hat{\mtrx{D}}\mtrx{A}$, where $\hat{\mtrx{D}}[i][i] = 1/\sum_{k=1}^{|\vertices{V}|} (\mtrx{A}[i][k]+\mtrx{A}[k][i]))$. $\mtrx{S}_{AA}$ computes the weighted sum of common neighbors where the weight is inversely proportional to the degree of the neighbor.

\emph{Katz index} is defined as $\mtrx{S}_{Katz} = \sum_{k=1}^\infty (\beta \mtrx{A})^k$. $\mtrx{S}_{Katz}[i][j]$ corresponds to a weighted sum of all the paths between two nodes $\vertex{v}_i$ and $\vertex{v}_j$. $\beta$ controls the depth of the connections: the closer $\beta$ is to $1$, the longer paths one wants to consider. One can rewrite the formula recursively as $\beta \mtrx{A} \mtrx{S}_{Katz} + \beta \mtrx{A} = \mtrx{S}_{Katz}$ and, as a corollary, obtain $\mtrx{S}_{Katz} = (\mtrx{I}_N - \beta \mtrx{A})^{-1}\beta \mtrx{A}$.

\emph{Preferential Attachment} is simply a product of in- and out- degrees of nodes: $\mtrx{S}_{PA}[i][j] = (\sum_{k=1}^{|\vertices{V}|} \mtrx{A}[i][k] ) (\sum_{k=1}^{|\vertices{V}|} \mtrx{A}[k][j])$.

\subsubsection{Shallow Encoders} \label{sec:shallow-encoders}
Shallow encoders first decide on the number and shape of the vectors and matrices for node and relation embeddings. Then, they consider each element in these vectors and matrices as a parameter to be directly learned from the data. A shallow encoder can be viewed as a lookup function that finds the hidden representation corresponding to a node or a relation given their id. Shallow encoders are commonly used for KG embedding (see e.g., \cite{nickel2011three,yang2015embedding,trouillon2016complex,bordes2013translating,StransE,kazemi2018simple,dettmers2018convolutional}).

\subsubsection{Decomposition Approaches}
\label{decomposition}
Decomposition methods are among the earliest attempts for developing encoders for graphs. They learn node embeddings similar to shallow encoders but in an unsupervised way: the node embeddings are learned in a way that connected nodes are close to each other in the embedded space. Once the embeddings are learned, they can be used for purposes other than reconstructing the edges (e.g., for clustering). Formally, for an undirected graph $\graph{G}$, learning node embeddings  $\embedding{EMB}(\vertex{v}_i)=(\vctr{z}_{\vertex{v}_i})$, where $\vctr{z}_{\vertex{v}_i }\in \mathbb{R}^d$, such that connected nodes are close in the embedded space can be done through solving the following optimization problem:
\begin{align}
    \label{eq:decomposition-close-embeddings}
    \min_{\{\vctr{z}_{\vertex{v}_i}\}_{ i = 1}^N}\sum_{i,j} \mtrx{A}[i][j]||\vctr{z}_{\vertex{v}_i} - \vctr{z}_{\vertex{v}_j}||^2
\end{align}
This loss ensures that connected nodes are close to each other in the embedded space.
One needs to impose some constraints to get rid of a scaling factor and to eliminate the trivial solution where all nodes are set to a single vector. For that let us consider a new matrix $\mtrx{Y} \in \mathbb{R}^{|\vertices{V}| \times d}$, such that its rows give the embedding: $\mtrx{Y}[i] = \transpose{\vctr{z}_{\vertex{v}_i}}$. Then one can add the constraints to the optimization problem~\eqref{eq:decomposition-close-embeddings}: $\transpose{\mtrx{Y}} \mtrx{D} \mtrx{Y} = \mtrx{I} $, where $\mtrx{D}$ is a diagonal matrix of degrees as defined in Subsection~\ref{defs_static}.
 As was proved by 
\cite{belkin}, this constrained optimization is equivalent to solving a generalized eigenvalue decomposition:
\begin{align}
    \label{eq_eigen}
    \mtrx{L} \vctr{y} = \lambda \mtrx{D} \vctr{y},   
\end{align}
where $\mtrx{L}$ is a graph Laplacian; the matrix $\mtrx{Y}$ can be obtained by considering the $|\vertices{V}| \times d$  matrix of top-$d$ generalized eigenvectors:  $\mtrx{Y} =  [\vctr{y}_1 \dots \vctr{y}_d]$. 

\citet{ase} suggested to use a slightly different embedding based on the eigenvalue decomposition of the adjacency matrix $\mtrx{A} = \mtrx{U}  \mtrx{\Sigma} \mtrx{U}' $ (this matrix is symmetric for an undirected graph). Then one can choose the top $d$ eigenvalues $\{ \lambda_1, \dots, \lambda_d \} $ and the corresponding eigenvectors $\{ \vctr{u}_1, \dots, \vctr{u}_d \} $  and construct a new matrix 
\begin{align}
\label{ASE}
    \mtrx{Z} = \mtrx{U}_{\le d} \sqrt{ \mtrx{\Sigma}_{ \le d} } \in \mathbb{R}^{|\vertices{V}|\times d},
\end{align}
where $\mtrx{\Sigma}_{\le d} = \text{diag}(\lambda_1, \dots, \lambda_d ) $, and $\mtrx{U}_{\le d} = [\vctr{u}_1 \dots \vctr{u}_d ] $. Rows of this matrix can be used as node embedding: $\vctr{z}_{\vertex{v}_i}= \transpose{\mtrx{Z}[i]} \in \mathbb{R}^d $. This is the so called \emph{adjacency spectral embedding}, see also \cite{levin2018out}. 

For directed graphs, because of their asymmetric nature, keeping track of the \Th{n}-order neighbors where $n > 1$ becomes difficult. For this reason, working with a high-order proximity matrix $\mtrx{S}$ is preferable (see Section~\ref{sec:high-order-encoder} for a description of high-order proximity matrices). Moreover, for directed graphs, it may be preferable to learn two vector representations per node, one to be used when the node is the source and the other to be used when the node is the target of an edge. One may learn embeddings for directed graphs by solving the following: 
\begin{align}
    \label{eq:optimization-problem}
    \min_{\mtrx{Z}_s,\mtrx{Z}_t} \; ||\mtrx{S} - \mtrx{Z}_s\transpose{\mtrx{Z}_t} ||^2_F ,
\end{align}
 where $||.||_F$ is the Frobenius norm and  $\mtrx{Z}_s, \mtrx{Z}_t \in \mathbb{R}^{|\vertices{V}|\times d}$. Given the solution, one can define the \emph{source features} of a node $\vertex{v}_i$ as  $\transpose{\mtrx{Z}_s[i]} $ and the \emph{target features} as $\transpose{\mtrx{Z}_t[i]}$. A single-vector embedding of a node $\vertex{v}_i$ can be defined as a concatenation of these features. 
 The Eckart–Young–Mirsky theorem~(\cite{ey}) from linear algebra indicates that the solution is equivalent to finding the singular value decomposition of $\mtrx{S}$:
\begin{align}
\label{svd}
    \mtrx{S} = \mtrx{U}_s \mtrx{\Sigma} (\mtrx{U}_t)',
\end{align}
where $\mtrx{\Sigma} = \text{diag}(\sigma_1, \dots, \sigma_{|\vertices{V}|})$ is a matrix of singular values and $\mtrx{U}_s$ and $\mtrx{U}_t$ are matrices of left and right singular vectors respectively (stacked as columns). Then using the top $d$ singular vectors one gets the solution of the optimization problem in (\ref{eq:optimization-problem}): 
\begin{align}
 \mtrx{Z}_s = (\mtrx{U_s})_{\le d}\sqrt{\mtrx{\Sigma}}_{\le d} \\
    \mtrx{Z}_t = (\mtrx{U_t})_{\le d} \sqrt{\mtrx{\Sigma}}_{\le d}  .
\end{align}

\subsubsection{Random Walk Approaches} \label{static-random-walk-sec}
A popular class of approaches for learning an embedding function for graphs is the class of random walk approaches. Similar to decomposition approaches, encoders based on random walks also learn embeddings in an unsupervised way. However, compared to decomposition approaches, these embeddings may capture longer-term dependencies. To describe the encoders in this category, first we define what a random walk is and then describe the encoders that leverage random walks to learn an embedding function.

\begin{definition}
A \emph{random walk} for a graph $\graph{G}=(\vertices{V}, \edges{E})$ is a sequence of nodes $\vertex{v}_1, \vertex{v}_2, \dots, \vertex{v}_l$ where $\vertex{v}_i\in\vertices{V}$ for all $1 \leq i \leq l$ and $(\vertex{v}_i, \vertex{v}_{i+1})\in\edges{E}$ for all $1 \leq i \leq l-1$. $l$ is called the \emph{length} of the walk.
\end{definition}

A random walk of length $l$ can be generated by starting at a node $\vertex{v}_i$ in the graph, then transitioning to a neighbor $\vertex{v}_j$ of $\vertex{v}_i$ ($j \neq i$), then transitioning to a neighbor of $\vertex{v}_j$ and continuing this process for $l$ steps. The selection of the first node and the node to transition to in each step can be uniformly at random or based on some distribution/strategy.

\begin{example}
Consider the graph in Figure~\ref{graphs-fig}(b). The following are three examples of random walks on this graph with length $4$: $1)~\vertex{v}_1, \vertex{v}_3, \vertex{v}_2, \vertex{v}_3$, $2)~\vertex{v}_2, \vertex{v}_1, \vertex{v}_2, \vertex{v}_4$ and $3)~\vertex{v}_4, \vertex{v}_2, \vertex{v}_4, \vertex{v}_2$.
In the first walk, the initial node has been selected to be $\vertex{v}_1$. Then a transition has been made to $\vertex{v}_3$, which is a neighbor of $\vertex{v}_1$. Then a transition has been made to $\vertex{v}_2$, which is a neighbor of $\vertex{v}_3$ and then a transition back to $\vertex{v}_3$, which is a neighbor of $\vertex{v}_2$. The following are two examples of invalid random walks: $1)~\vertex{v}_1, \vertex{v}_4, \vertex{v}_2, \vertex{v}_3$ and $2)~\vertex{v}_1, \vertex{v}_3, \vertex{v}_4, \vertex{v}_2$.
The first one is not a valid random walk since a transition has been made from $\vertex{v}_1$ to $\vertex{v}_4$ when there is no edge between $\vertex{v}_1$ and $\vertex{v}_4$. The second one is not valid because a transition has been made from $\vertex{v}_3$ to $\vertex{v}_4$ when there is no edge between $\vertex{v}_3$ and $\vertex{v}_4$.
\end{example}

Random walk encoders perform multiple random walks of length $l$ on a graph and consider each walk as a sentence, where the nodes are considered as the words of these sentences.  Then they use the techniques from natural language processing for learning word embeddings (e.g., \cite{mikolov2013distributed,pennington2014glove}) to learn a vector representation for each node in the graph. One such approach is to create a matrix $\mtrx{S}$ from these random walks such that $\mtrx{S}[i][j]$ corresponds to the number of times $\vertex{v}_i$ and $\vertex{v}_j$ co-occurred in random walks and then factorize the matrix (see Section~\ref{decomposition}) to get vector representations for nodes. 

Random walk encoders typically differ in the way they perform the walk, the distribution they use for selecting the initial node, and the transition distribution they use. For instance, DeepWalk~(\cite{perozzi2014deepwalk}) selects both the initial node and the node to transition to uniformly at random. \citet{perozzi2016walklets} extends DeepWalk by allowing random walks to skip over multiple nodes at each transition. Node2Vec~(\cite{grover2016node2vec}) selects the node to transition to based on a combination of breadth-first search (to capture local information) and depth-first search (to capture global information).

Random walk encoders have been extended to KGs (and HINs) through constraining the walks to conform to some meta-paths. A meta-path can be considered as a sequence of relations in $\relations{R}$. \citet{dong2017metapath2vec} propose \emph{metapath2vec} where each random walk is constrained to conform to a meta-path $\relation{r_1}, \relation{r_2}, \dots, \relation{r_k}$ by starting randomly at a node $\vertex{v}_1$ whose type is compatible with the source type of $\relation{r_1}$. Then the walk transitions to a node $\vertex{v}_2$ where $\vertex{v}_2$ is selected uniformly at random among the nodes having relation $\relation{r}_1$ with $\vertex{v}_1$, then the walk transitions to a node $\vertex{v}_3$ where $\vertex{v}_3$ is selected uniformly at random among the nodes having relation $\relation{r}_2$ with $\vertex{v}_2$, and so forth. Each meta-path provides a semantic relationship between the start and end nodes. 

\citet{shi2018heterogeneous} take a similar approach as metapath2vec but aim at learning node embeddings that are geared more towards improving recommendation performance. Both \citet{dong2017metapath2vec} and \citet{shi2018heterogeneous} use a set of hand-crafted meta-paths to guide the random walks. Instead of hand-crafting meta-paths, \citet{chen2017task} propose a greedy approach to select the meta-paths based on performance on a validation set. \citet{zhang2018metagraph2vec} identify some limitations for models restricting random walks to conform to meta-paths. They propose meta-graphs as an alternative to meta-paths in which relations are connected as a graph (instead of a sequence) and at each node, the walk can select to conform to any outgoing edge in the meta-graph. \cite{ristoski2016rdf2vec} extend random walk approaches to general RDF data.

\subsubsection{Autoencoder Approaches} \label{sec:static-autoencoder}
Another class of models for learning an embedding function for static graphs is by using autoencoders. Similar to the decomposition approaches, these approaches are also unsupervised. However, instead of learning shallow embeddings that reconstruct the edges of a graph, the models in this category create a deep encoder that compresses a node's neighborhood to a vector representation, which can be then used to reconstruct the node's neighborhood. The model used for compression and reconstruction is referred to as an autoencoder. Similar to the decomposition approaches, once the node embeddings are learned, they may be used for purposes other than predicting a node's neighborhood. 

In its simplest form, an autoencoder~(\cite{hinton2006reducing}) contains two components called the \emph{encoder} and \emph{reconstructor}\footnote{Reconstructor is also called \emph{decoder} but we use the name \emph{reconstructor} to avoid confusion with graph decoders.}, where each component is a feed-forward neural network. 
The encoder takes as input a vector $\vctr{a}\in\mathbb{R}^{N}$ (e.g., corresponding to $N$ numerical features of an object) and passes it through several feed-forward layers producing  $\vctr{z}\in\mathbb{R}^{d}$ such that $d << N$. The reconstructor receives $\vctr{z}$ as input and passes it through several feed-forward layers aiming at reconstructing $\vctr{a}$. That is, assuming the output of the reconstructor is $\hat{\vctr{a}}$, the two components are trained such that $||\vctr{a}-\hat{\vctr{a}}||$ is minimized. $\vctr{z}$ can be considered a compression of $\vctr{a}$.

Let $\graph{G}=(\vertices{V}, \edges{E})$ be a graph with adjacency matrix $\mtrx{A}$. For a node $\vertex{v}_i\in\vertices{V}$, let  $\mtrx{A}[i]$ represent the $\Th{i}$ row of the adjacency matrix corresponding to the neighbors of $\vertex{v}_i$. To use autoencoders for generating node embeddings, \citet{wang2016structural} train an autoencoder (named \emph{SDNE}) that takes a vector $\mtrx{A}[i]\in\mathbb{R}^{|\vertices{V}|}$ as input, feeds the input vector through an encoder and produces $\vctr{z}_i\in\mathbb{R}^d$, and then feeds $\vctr{z}_i$ into a reconstructor to reconstruct $\mtrx{A}[i]$. After training, the $\vctr{z}_i$ vectors corresponding to the output of the encoder of the autoencoder can be considered as embeddings for the nodes $\vertex{v}_i$. A graph decoder can be applied to these embeddings to make predictions. $\vctr{z}_i$ and $\vctr{z}_j$ may further be constrained to be close in Euclidean space if $\vertex{v}_i$ and $\vertex{v}_j$ are connected. 
For the case of attributed graphs, \citet{tran2018learning} concatenates the attribute values $\vctr{x}_i$ of node $\vertex{v}_i$ to $\mtrx{A}[i]$ and feeds the concatenation $[\vctr{x}_i;\mtrx{A}[i]]$ into an autoencoder. \citet{cao2016deep} propose an autoencoder approach (named \emph{RDNG}) that is similar to SDNE, but they first compute a high-order proximity matrix $\mtrx{S}\in\mathbb{R}^{|\vertices{V}|\times |\vertices{V}|}$ based on node co-occurrences on random walks (any other matrix from Section~\ref{sec:high-order-encoder} may also be used), and then feed $\mtrx{S}[i]$s into an autoencoder. 

\subsubsection{Graph Convolutional Network Approaches} \label{sec:gcn}
Yet another class of models for learning node embeddings in a graph are graph convolutional networks (GCNs). As the name suggests, graph convolutions generalize convolutions to arbitrary graphs. Graph convolutions have spatial (see, e.g., \cite{hamilton2017inductive,hamilton2017representation,schlichtkrull2018modeling,gilmer2017neural}) and spectral constructions (see, e.g., \cite{liao2019lanczosnet,kipf2017semi,defferrard2016convolutional,levie2017cayleynets}). Here, we describe the spatial (or message passing) view and refer the reader to \cite{bronstein2017geometric} for the spectral view.

A GCN consists of multiple layers where each layer takes node representations (a vector per node) as input and outputs transformed representations. Let $\textbf{z}_{\vertex{v}, l}$ be the representation for a node $\vertex{v}$ after passing it through the $\Th{l}$ layer. A very generic forward pass through a GCN layer transforms the representation of each node \vertex{v} as follows: 
\begin{align}
    \vctr{z}_{\vertex{v}, l+1} = \function{transform}(\{\vctr{z}_{\vertex{v}, j}\}_{0 \leq j \leq l}, \{\vctr{z}_{\vertex{u}, k}\}_{\vertex{u} \in \neighbour(\vertex{v}), 0 \leq k \leq l}, \Theta)\label{eq:gcn}
\end{align}
where $\neighbour(\vertex{v})$ represents the neighbors of $\vertex{v}$ and $\function{transform}$ is a function parametrized by $\Theta$ which aggregates the information from the previous representations of the neighbors of $\vertex{v}$ and combines it with the previous representations of $\vertex{v}$ itself to compute $\vctr{z}_{\vertex{v}, l+1}$. The $\function{transform}$ function should be invariant to the order of the nodes in $\neighbour(\vertex{v})$ because there is no specific ordering to the nodes in an arbitrary graph. Moreover, it should be able to handle a variable number of neighbors.
If the graph is attributed, for each node \vertex{v}, $\vctr{z}_{\vertex{v},0}$ can be initialized to $\vctr{x}_\vertex{v}$ corresponding to the attribute values of $\vertex{v}$ (see, e.g., \cite{kipf2017semi}). Otherwise, they can be initialized using a one-hot encoding of the nodes (see, e.g.,\cite{schlichtkrull2018modeling}). 
In a GCN with $L$ layers, each node receives information from the nodes at most $L$ hops away from it.

There is a large literature on the design of the $\function{transform}$ function (see, e.g., \cite{li2015gated,kipf2017semi,hamilton2017inductive,dai2018learning}). 
\citet{kipf2017semi} formulate it as:
\begin{align}
    \mtrx{Z}_{l+1} = \sigma(\Tilde{\mtrx{D}}^{-\frac{1}{2}}\Tilde{\mtrx{A}}\Tilde{\mtrx{D}}^{-\frac{1}{2}}\mtrx{Z}_l\mtrx{W}_{l+1})
\end{align}
where $\Tilde{\mtrx{A}} = \mtrx{A} + \mtrx{I}_N$ is adjacency matrix with self-connections for input graph, $N$ is the number of nodes in the graph, $\mtrx{I}_N$ is the identity matrix, $\mtrx{W}_{l+1}$ is a parameter matrix for the $\Th{(l+1)}$ layer and $\sigma(.)$ is a non-linearity.  $\Tilde{\mtrx{D}}^{-\frac{1}{2}}\Tilde{\mtrx{A}}\Tilde{\mtrx{D}}^{-\frac{1}{2}}\mtrx{Z}_l$ corresponds to taking a normalized average of the features of $\vertex{v}$ and its neighbors (treating the features of $\vertex{v}$ and its neighbors identically).
Other formulations for the $\function{transform}$ function can be found in several recent surveys (see, e.g., \cite{zhou2018graph,cai2018comprehensive}). 

For a node $\vertex{v}$, not all the neighboring nodes may be equally important. 
\citet{velivckovic2018graph} propose an adaptive attention mechanism that learns to weigh the neighbors depending on their importance when aggregating information from the neighbors. The mechanism is adaptive in the sense that the weight of a node is not fixed and depends on the current representation of the node for which the aggregation is performed. 
Following \citet{vaswani2017attention}, \citet{velivckovic2018graph} also use multi-headed attention.
\emph{GaAN}~(\cite{zhang2018gaan}) extends this idea and introduces adaptive attention weights for different attention heads, i.e., the weights for different attention heads depend on the node for which the multi-head attention is being applied.

In graphs like social networks, there can be nodes that have a large number of neighbors. This can make the $\function{transform}$ function computationally prohibitive. \citet{hamilton2017inductive} propose to use a uniform sampling of the neighbors to fix the neighborhood size to a constant number. Not only the sampling helps reduce computational complexity and speed up training, but also it acts as a regularizer.
\citet{ying2018graph} propose an extension of this idea according to which the neighborhood of a node $\vertex{v}$ is formed by repeatedly starting truncated random walks from $\vertex{v}$ and choosing the nodes most frequently hit by these truncated random walks. In this way, the neighborhood of a node consists of the nodes most relevant to it, regardless of whether they are connected with an edge or not.

\textbf{Expressivity:} There are currently two approaches for measuring the expressivity of GCNs. \citet{xu2019powerful} study the expressiveness of certain GCN models with respect to graph classification (see Definition~\ref{dfnt:full-exp-iso}) and show that in terms of distinguishing non-isomorphic graphs, these GCNs are \emph{at most} as powerful as the Weisfeiler-Lehman isomorphism test~(\cite{weisfeiler1968reduction}) \textemdash~a test which is able to distinguish a broad class of graphs~(\cite{babai1979canonical}) but also known to fail in some corner cases~(\cite{cai1992optimal}). In a concurrent work, a similar result has been reported by \citet{morris2019weisfeiler}. \citet{xu2019powerful} provide the necessary conditions under which these GCNs become as powerful as the Weisfeiler-Lehman test.
On the other hand, \citet{Maron2019OnTU} and \citet{Keriven2019UniversalIA} study how well certain GCN models can approximate any continuous function which is invariant to permutation of its input. They proved that a certain class of networks, called \emph{G-invariant}, are universal approximators. \citet{Chen2019OnTE} demonstrate that these two approaches to the expressivity of GCNs are closely related. 

\textbf{GCNs for KGs:} Several works extend GCNs to KG embedding. One notable example is called relational GCN (RGCN)~(\cite{schlichtkrull2018modeling}). 
The core operation that RGCN does differently is the application of a relation specific transformation (i.e., the transformation depends on the direction and the label of the edge) to the neighbors of the nodes in the aggregation function. In RGCN, the $\function{transform}$ function is defined as follows:
\begin{align}\label{eq:rgcn}
    \vctr{z}_{\vertex{v},l+1} = \sigma(\sum_{\relation{r}\in \relations{R}}\sum_{\vertex{u}\in \neighbour(\vertex{v}, \relation{r})}\frac{1}{c_{\vertex{v},\relation{r}}}\mtrx{W}_{r,l}\vctr{z}_{\vertex{u},l} + \mtrx{W}_{0,l}\vctr{z}_{\vertex{v},l})
\end{align}
where $\relations{R}$ is the set of relation types, $\neighbour(\vertex{v}, \relation{r})$ is the set of neighboring nodes connected to $\vertex{v}$ via relation $\relation{r}$, $c_{\vertex{v},\relation{r}}$ is a normalization factor that can either be learned or fixed (e.g., to $|\neighbour(\vertex{v}, \relation{r})|$), $\mtrx{W}_{\relation{r},l}$ is a transformation matrix for relation $\relation{r}$ at the $\Th{l}$ layer, and $\mtrx{W}_{0,l}$ is a self-transformation matrix at the $\Th{l}$ layer. \citet{sourek2018lifted} and \citet{kazemi2018relnn} propose other variants for Equation~\eqref{eq:rgcn} where (roughly) the transformations are done using soft first-order logic rules. \citet{wang2019kgat} and \citet{nathani2019learning} propose attention-based variants of Equation~\eqref{eq:rgcn}.

\subsection{Decoders} \label{sec:static-decoders}
We divide the discussion on decoders into those used for graphs and those used for KGs.

\subsubsection{Decoders for Static Graphs}
For static graphs, the embedding function usually maps each node to a single vector; that is, $\embedding{EMB}(\vertex{v})=(\vctr{z}_{\vertex{v}})$ where $\vctr{z}_{\vertex{v}}\in\mathbb{R}^{d}$ for any $\vertex{v}\in\vertices{V}$. To classify a node $\vertex{v}$, a decoder can be any classifier on $\vctr{z}_{\vertex{v}}$ (e.g., logistic regression or random forest). 

To predict a link between two nodes $\vertex{v}$ and $\vertex{u}$, for undirected (and bipartite) graphs, the most common decoder is based on the \emph{dot-product} of the vectors for the two nodes, i.e.,  $\transpose{\vctr{z}_{\vertex{v}}}\vctr{z}_{\vertex{u}}$. The dot-product gives a score that can then be fed into a sigmoid function whose output can be considered as the probability of a link existing between $\vertex{v}$ and $\vertex{u}$. \citet{grover2016node2vec} propose several other decoders for link prediction in undirected graphs. Their decoders are based on defining a function $\function{f}(\vctr{z}_{\vertex{v}}, \vctr{z}_{\vertex{u}})$ that combines the two vectors $\vctr{z}_{\vertex{v}}$ and $\vctr{z}_{\vertex{u}}$ into a single vector. The resulting vector is then considered as the edge features that can be fed into a classifier. These combining functions include average $\frac{\vctr{z}_{\vertex{v}}+\vctr{z}_{\vertex{u}}}{2}$, Hadamard multiplication $\vctr{z}_{\vertex{v}}\odot\vctr{z}_{\vertex{u}}$, absolute value of the difference $\function{abs}(\vctr{z}_{\vertex{v}}-\vctr{z}_{\vertex{u}})$, and squared value of the difference $(\vctr{z}_{\vertex{v}}-\vctr{z}_{\vertex{u}})^2$.
Instead of computing the distance between $\vctr{z}_{\vertex{v}}$ and $\vctr{z}_{\vertex{u}}$ in the Euclidean space, the distance can be computed in other spaces such as the hyperbolic space (see, e.g., \cite{chamberlain2017neural}). Different spaces offer different properties. Note that all these four combination functions are symmetric, i.e., $\function{f}(\vctr{z}_{\vertex{v}}, \vctr{z}_{\vertex{u}})=\function{f}(\vctr{z}_{\vertex{u}}, \vctr{z}_{\vertex{v}})$ where $\function{f}$ is any of the above functions. This is an important property when the graph is undirected. 

For link prediction in digraphs, it is important to treat the source and target of the edge differently. Towards this goal, one approach is to \emph{concatenate} the two vectors as $[\vctr{z}_{\vertex{v}};\vctr{z}_{\vertex{u}}]$ and feed the concatenation into a classifier (see, e.g., \cite{pareja2019evolvegcn}). Another approach used by \cite{ma2018dynamic} is to project the source and target vectors to another space as $\hat{\vctr{z}}_{\vertex{v}}=\mtrx{W}_1\vctr{z}_{\vertex{v}}$ and $\hat{\vctr{z}}_{\vertex{u}}=\mtrx{W}_2\vctr{z}_{\vertex{u}}$, where $\mtrx{W}_1$ and $\mtrx{W}_2$ are matrices with learnable parameters, and then take the dot-product in the new space (i.e., $\transpose{\hat{\vctr{z}}_{\vertex{v}}}\hat{\vctr{z}}_{\vertex{u}}$). A third approach is to take the vector representation $\vctr{z}_{\vertex{v}}$ of a node $\vertex{v}\in\vertices{V}$ and send it through a feed-forward neural network with $|\vertices{V}|$ outputs where each output gives the score for whether $\vertex{v}$ has a link with one of the nodes in the graph or not. This approach is used mainly in graph autoencoders (see, e.g., \cite{wang2016structural,cao2016deep,tran2018learning,goyal2017dyngem,chen2018gc}) and is used for both directed and undirected graphs. 

The decoder for a graph classification task needs to compress node representations into a single representation which can then be fed into a classifier to perform graph classification.  \citet{duvenaud2015convolutional} simply average all the node representations into a single vector. \citet{gilmer2017neural} consider the node representations of the graph as a set and use the DeepSet aggregation~(\cite{zaheer2017deep}) to get a single representation. \citet{li2015gated} add a virtual node to the graph which is connected to all the nodes and use the representation of the virtual node as the representation of the graph. Several approaches perform a deterministic hierarchical graph clustering step and combine the node representations in each cluster to learn hierarchical representations (\cite{defferrard2016convolutional,fey2018splinecnn,simonovsky2017dynamic}). Instead of performing a deterministic clustering and then running a graph classification model, \citet{ying2018hierarchical} learn the hierarchical structure jointly with the classifier in an end-to-end fashion.

\subsubsection{Decoders for Link Prediction in Static KGs} \label{sec:static-kg-decoders}
We provide an overview of the \emph{translational}, \emph{bilinear}, and \emph{deep learning} decoders for KGs. When we discuss the expressivity of the decoders in this subsection, we assume the decoder is combined with a shallow encoder (see Section~\ref{sec:shallow-encoders}).

\paragraph{Translational decoders} usually assume the encoder provides an embedding function such that $\embedding{EMB}(\vertex{v})=(\vctr{z}_\vertex{v})$ for every $\vertex{v}\in\vertices{V}$ where $\vctr{z}_\vertex{v}\in\mathbb{R}^{d_1}$, and $\embedding{EMB}(\relation{r})=(\vctr{z}_\relation{r}, \mtrx{P}_\relation{r},\mtrx{Q}_\relation{r})$ for every $\vertex{r}\in\relations{R}$ where $\vctr{z}_\relation{r}\in\mathbb{R}^{d_2}$ and $\mtrx{P}_\relation{r},\mtrx{Q}_\relation{r}\in\mathbb{R}^{d_1\times d_2}$. That is, the embedding for a node contains a single vector whereas the embedding for a relation contains a vector and two matrices. For an edge $(\vertex{v}, \relation{r}, \vertex{u})$, these models use:
\begin{align}
||\mtrx{P}_\relation{r}\vctr{z}_\relation{v}+\vctr{z}_\relation{r}-\mtrx{Q}_\relation{r}\vctr{z}_\relation{u}||_i
\end{align}
as the dissimilarity score for the edge where $||.||_i$ represents norm $i$ of a vector. $i$ is usually either $1$ or $2$. Translational decoders differ in the restrictions they impose on $\mtrx{P}_\relation{r}$ and $\mtrx{Q}_\relation{r}$. TransE~(\cite{bordes2013translating}) constrains $\mtrx{P}_\relation{r}=\mtrx{Q}_\relation{r}=\imtrx{d}$. So the dissimilarity function for TransE can be simplified to $||\vctr{z}_\relation{v}+\vctr{z}_\relation{r}-\vctr{z}_\relation{u}||_i$.
In TransR~(\cite{lin2015learning}), $P_r=Q_r$. In STransE~(\cite{StransE}), no restrictions are imposed on the matrices. \citet{kazemi2018simple} proved that regardless of the encoder, TransE, TransR, STransE, and many other variants of translational approaches are not fully expressive for link prediction (see Definition~\ref{dfnt:full-exp-link} for a definition of fully expressive for link prediction) and identified severe restrictions on the type of relations these approaches can model.

\begin{figure}[t]
\centering
\includegraphics[width=0.8\textwidth]{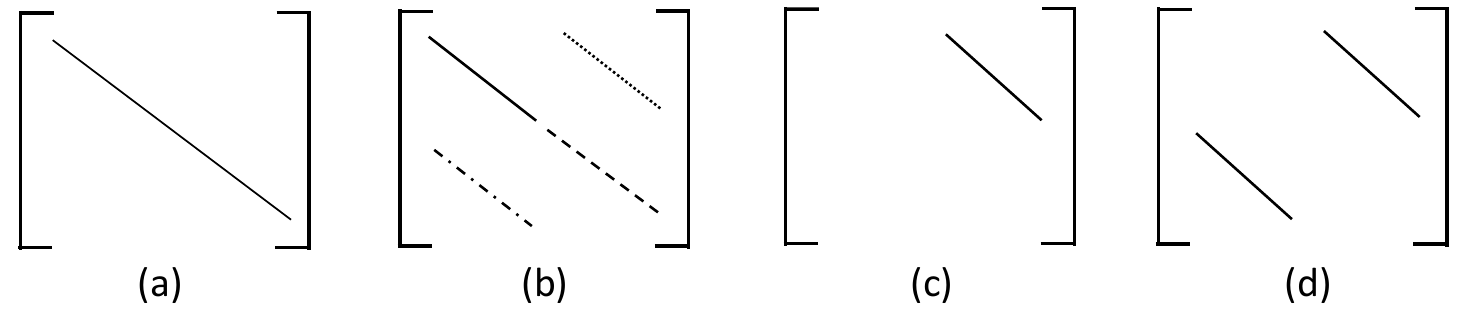}
\caption{A graphical representation of the constraints over the $P_r$ matrices for bilinear models (a) DistMult, (b) ComplEx, (c) CP, and (d) SimplE taken from \cite{kazemi2018simple} where lines represent the non-zero elements of the matrices. In ComplEx, the parameters represented by the dashed line are tied (i.e., equal) to the parameters represented by the solid line and the parameters represented by the dotted line are tied to the negative of the dotted-and-dashed line.}
\label{bilinear-fig}
\end{figure}

\paragraph{Bilinear decoders} usually assume the encoder provides an embedding function such that $\embedding{EMB}(\vertex{v})=(\vctr{z}_\vertex{v})$ for every $\vertex{v}\in\vertices{V}$ where $\vctr{z}_\vertex{v}\in\mathbb{R}^{d}$, and $\embedding{EMB}(\relation{r})=(\mtrx{P}_\relation{r})$ for every $\vertex{r}\in\relations{R}$ where $\mtrx{P}_\relation{r}\in\mathbb{R}^{d\times d}$. For an edge $(\vertex{v}, \relation{r}, \vertex{u})$, these models use:
\begin{align}\label{eq:bilinear}
\transpose{\vctr{z}_\vertex{v}}\mtrx{P}_\relation{r}\vctr{z}_\vertex{u}
\end{align}
as the similarity score for the edge. Bilinear decoders differ in the restrictions they impose on $\mtrx{P}_\relation{r}$ matrices~(see \cite{wang2018multi}). In RESCAL~(\cite{nickel2011three}), no restrictions are imposed on the $\mtrx{P}_\relation{r}$ matrices. RESCAL is fully expressive with respect to link prediction, but the large number of parameters per relation makes RESCAL prone to overfitting. To reduce the number of parameters in RESCAL, DistMult~(\cite{yang2015embedding}) constrains the $\mtrx{P}_\relation{r}$ matrices to be diagonal. This reduction in the number of parameters, however, comes at a cost: DistMult loses expressivity and is only able to model symmetric relations as it does not distinguish between the source and target vectors.

ComplEx~(\cite{trouillon2016complex}), CP~(\cite{hitchcock1927expression}) and SimplE~(\cite{kazemi2018simple}) reduce the number of parameters in RESCAL without sacrificing full expressivity. ComplEx extends DistMult by assuming the embeddings are complex (instead of real) valued, i.e. $\vctr{z}_\vertex{v}\in\mathbb{C}^{d}$ and $\mtrx{P}_\relation{r}\in\mathbb{C}^{d\times d}$ for every $\vertex{v}\in\vertices{V}$ and $\relation{r}\in\relations{R}$. Then, it slightly changes the score function to $\function{Real}(\transpose{\vctr{z}_\vertex{v}}\mtrx{P}_\relation{r}\function{conjugate}(\vctr{z}_\vertex{u}))$ where $\function{Real}$ returns the real part of an imaginary number and $\function{conjugate}$ takes an element-wise conjugate of the vector elements. By taking the conjugate of the target vector, ComplEx differentiates between source and target nodes and does not suffer from the symmetry issue of DistMult.
CP defines $\embedding{EMB}(\vertex{v})=(\vctr{z}_\vertex{v}, \overline{\vctr{z}}_\vertex{v})$, i.e. the embedding of a node consists of two vectors, where $\vctr{z}_\vertex{v}$ captures the $\vertex{v}$'s behaviour when it is the source of an edge and $\overline{\vctr{z}}_\vertex{v}$ captures $\vertex{v}$'s behaviour when it is the target of an edge. For relations, CP defines $\embedding{EMB}(\relation{r})=(\vctr{z}_\relation{r})$. The similarity function of CP for an edge $(\vertex{v}, \relation{r}, \vertex{u})$ is then defined as $\transpose{\vctr{z}_\vertex{v}}\function{diag}(\vctr{z}_\relation{r})\overline{\vctr{z}}_\vertex{u}$. Realizing the information may not flow well between the two vectors of a node, SimplE adds another vector to the relation embeddings as $\embedding{EMB}(\relation{r})=(\vctr{z}_\relation{r}, \overline{\vctr{z}}_\relation{r})$ where $\overline{\vctr{z}}_\relation{r}$ models the behaviour of the inverse of the relation. Then, it changes the score function to be the average of $\transpose{\vctr{z}_\vertex{v}}\function{diag}(\vctr{z}_\relation{r})\overline{\vctr{z}}_\vertex{u}$ and $\transpose{\vctr{z}_\vertex{u}}\function{diag}(\overline{\vctr{z}}_\relation{r})\overline{\vctr{z}}_\vertex{v}$. 

For ComplEx, CP, and SimplE, it is possible to view the embedding for each node $\vertex{v}$ as a single vector in $\mathbb{R}^{2d}$ by concatenating the two vectors (for ComplEx, the two vectors correspond to the real and imaginary parts of the embedding vector). Then, the $\mtrx{P}_\relation{r}$ matrices can be viewed as being restricted according to Figure~\ref{bilinear-fig}.

Other bilinear approaches include HolE~(\cite{sadilek2010recognizing})
whose equivalence to ComplEx has been established (see \cite{hayashi2017equivalence}), and Analogy~(\cite{liu2017analogical}) where the $\mtrx{P}_\relation{r}$ matrices are constrained to be block-diagonal.

\paragraph{Deep learning-based decoders:} Deep learning approaches typically use feed-forward or convolutional neural networks for scoring edges in a KG. \citet{dong2014knowledge} and \citet{santoro2017simple} consider $\embedding{EMB}(\vertex{v})=(\vctr{z}_\vertex{v})$ for every node $\vertex{v}\in\vertices{V}$ such that $\vctr{z}_\vertex{v}\in\mathbb{R}^{d_1}$ and $\embedding{EMB}(\relation{r})=(\vctr{z}_\relation{r})$ for every relation $\vertex{r}\in\relations{R}$ such that $\vctr{z}_\relation{r}\in\mathbb{R}^{d_2}$. Then for an edge $(\vertex{v}, \relation{r}, \vertex{u})$, they feed $[\vctr{z}_\vertex{v};\vctr{z}_\relation{r};\vctr{z}_\vertex{u}]$ (i.e., the concatenation of the three vector representations) into a feed-forward neural network that outputs a score for this edge. \citet{dettmers2018convolutional} develop a score function based on convolutions. They consider $\embedding{EMB}(\vertex{v})=(\vctr{Z}_\vertex{v})$ for each node $\vertex{v}\in\vertices{V}$ such that $\mtrx{Z}_\vertex{v}\in\mathbb{R}^{d_1\times d_2}$ and $\embedding{EMB}(\relation{r})=(\mtrx{Z}_\relation{r})$ for each relation $\vertex{r}\in\relations{R}$ such that $\vctr{Z}_\relation{r}\in\mathbb{R}^{d_1\times d_2}$\footnote{Alternatively, the matrices can be viewed as vectors of size $d_1 d_2$.}. For an edge (\vertex{v}, \relation{r}, \vertex{u}), first they combine $\mtrx{Z}_\vertex{v}$ and $\mtrx{Z}_\relation{r}$ into a matrix $\mtrx{Z}_{vr}\in\mathbb{R}^{2d_1\times d_2}$ by concatenating the two matrices on the rows, or by adding the $\Th{i}$ row of each matrix in turn. Then 2D convolutions with learnable filters are applied on $\mtrx{Z}_{\vertex{v}\relation{r}}$ generating multiple matrices and the matrices are vectorized into a vector $\vctr{c}_{\vertex{v}\relation{r}}$, where $|\vctr{c}_{\vertex{v}\relation{r}}|$ depends on the number of convolution filters. Then the score for the edge is computed as $(\transpose{\vctr{c}_{\vertex{v}\relation{r}}} \mtrx{W}) \function{vec}(\mtrx{Z}_\vertex{u})$ where $\mtrx{W}\in\mathbb{R}^{|\vctr{c}_{\vertex{v}\relation{r}}|\times (d_1 d_2)}$ is a weight matrix. Other deep learning approaches include HypER~(\cite{balazevic2018hypernetwork}) which is another score function based on convolutions, and neural tensor networks (NTN)~(\cite{socher2013reasoning}) which contains feed-forward components as well as several bilinear components.

\section{Encoders for Dynamic Graphs}
\label{sec:dynamic-encoders}
In Section~\ref{sec:static-encoders}, we described different encoders for static graphs. In this section, we describe several general categories of encoders for dynamic graphs.
Recall that reasoning problems for dynamic graphs can be for extrapolation or interpolation (see Section~\ref{sec:prediction-problems}). 
Although some encoders may be used for both problems, the extrapolation and interpolation problems typically require different types of encoders. For extrapolation, one needs an encoder that provides node and relation embeddings based only on the observations in the past. For interpolation, however, at any time $t$, one needs an encoder that provides node and relation embeddings based on the observations before, at, and after $t$.

\subsection{Aggregating Temporal Observations} \label{sec:aggregate-temp-obs}
A simple approach for dealing with the temporal aspect of a dynamic graph is through collapsing the dynamic graph into a static graph by aggregating the temporal observations (or the adjacency matrices) over time. Once an aggregated static graph is produced, a static encoder can be used to generate an embedding function. 

\citet{liben2007link} follow a simple aggregation approach for DTDGs by ignoring the timestamps and taking the sum (or union) of the entries of the adjacency matrices across all snapshots. That is, assuming $\mtrx{A}^1, \dots, \mtrx{A}^T$ represent the adjacency matrices for $T$ timestamps, \citet{liben2007link} first aggregate these adjacency matrices into a single matrix as follows:
\begin{align}\label{aggregate1-eq}
    \mtrx{A}_{sum}[i][j] = \sum_{t=1}^T \mtrx{A}^t[i][j]
\end{align}
Then a static decoder can be applied on $\mtrx{A}_{sum}$ to learn an embedding function. \citet{hisano2018semi} also follows a similar aggregation scheme where he takes the union of the previous $k$ formation and dissolution matrices of a DTDG. He defines the formation matrix for snapshot $t$ as a matrix representing which edges have been added to the graph since $\Th{(t-1)}$ snapshot and the dissolution matrix as a matrix representing which edges have been removed from the graph since $\Th{(t-1)}$ snapshot. These simple approaches lose the timing information and may not perform well when timing information are of high importance. 

An alternative to taking a uniform average of the adjacency matrices is to give more weights to snapshots that are more recent~(\cite{sharan2008temporal,ibrahim2015link,ahmed2016efficient,ahmed2016sampling}). Below is one such aggregation:
\begin{align}\label{aggregate2-eq}
    \mtrx{A}_{wsum}[i][j] = \sum_{t=1}^T \theta^{T-t}\mtrx{A}^t[i][j]
\end{align}
where $0 \leq \theta \leq 1$ controls the importance of recent snapshots. Larger values for $\theta$ put more emphasis on the more recent adjacency matrices.

\begin{example}
Let $\{\graph{G}^1, \graph{G}^2, \graph{G}^3\}$ be a DTDG with three snapshots. Let all $\graph{G}^i$s have the same set $\{\vertex{v}_1, \vertex{v}_2, \vertex{v}_3\}$ of nodes and the adjacency matrices be as follows:
\[
\mtrx{A}^1
=
\begin{bmatrix}
    0 & 1 & 0 \\
    1 & 0 & 1 \\
    0 & 1 & 0 
\end{bmatrix}
\quad
\mtrx{A}^2
=
\begin{bmatrix}
    0 & 1 & 1 \\
    1 & 0 & 1 \\
    1 & 1 & 0 
\end{bmatrix}
\quad
\mtrx{A}^3
=
\begin{bmatrix}
    0 & 1 & 1 \\
    1 & 0 & 0 \\
    1 & 0 & 0 
\end{bmatrix}
\]
The aggregation scheme in Equation~\eqref{aggregate1-eq} and Equation~\eqref{aggregate2-eq} (assuming $\theta=0.5$) respectively aggregate the three adjacency matrices into $\mtrx{A}_{sum}$ and $\mtrx{A}_{wsum}$ as follows:
\[
\mtrx{A}_{sum}
=
\begin{bmatrix}
    0 & 3 & 2 \\
    3 & 0 & 2 \\
    2 & 2 & 0 
\end{bmatrix}
\quad\quad\quad
\mtrx{A}_{wsum}
=
\begin{bmatrix}
    0 & \frac{7}{4} & \frac{3}{2} \\
    \frac{7}{4} & 0 & \frac{3}{4} \\
    \frac{3}{2} & \frac{3}{4} & 0 
\end{bmatrix} \label{example:temporal-aggr}
\]
Then an embedding function can be learned using $\mtrx{A}_{sum}$ or $\mtrx{A}_{wsum}$ (e.g., by using decomposition approaches). Although the interaction evolution between $\vertex{v}_1$ and $\vertex{v}_3$ (which were not connected at the beginning, but then formed a connection) is different from the interaction evolution between $\vertex{v}_2$ and $\vertex{v}_3$ (which were connected at the beginning and then got disconnected), $\mtrx{A}_{sum}$ assigns the same number to both these pairs. $\mtrx{A}_{wsum}$ contains more temporal information compared to $\mtrx{A}_{sum}$, but still loses large amounts of information. For instance, it is not possible to realize from $\mtrx{A}_{wsum}$ that $\vertex{v}_2$ and $\vertex{v}_3$ got disconnected only recently.

\end{example}

The approaches based on aggregating temporal observations typically enjoy advantages such as simplicity, scalability, and the capability to directly use a large body of literature on learning from static graphs. The aggregation in Equation~\eqref{aggregate1-eq} can be potentially used for both interpolation and extrapolation. The aggregation in Equation~\eqref{aggregate1-eq} may be more suited for extrapolation as it weighs recent snapshots more than the old ones. However, it can be easily adapted for interpolation by, e.g., changing $\theta^{T-t}$ in the equation with $\theta^{|t_q-t|}$ where $t_q$ is the timestamp for which we wish to make a prediction and $|.|$ returns the absolute value. Note that the aggregation approaches may lose large amounts of useful information hindering them from making accurate predictions in many scenarios.

\subsection{Aggregating Static Features}\label{sec:aggr-static-feats}
Rather than first aggregating a dynamic graph over time to produce a static graph and then running static encoders on the aggregated graph, in the case of DTDGs, one may first apply a static encoder to each snapshot and then aggregate the results over time. Let $\{\graph{G}^1, \dots, \graph{G}^T\}$ be a DTDG. The main idea behind the approaches in this category is to first use a static encoder (e.g., an encoder from Section~\ref{sec:high-order-encoder}) to compute/learn node features $\vctr{z}^t_\vertex{v}$ for each node $\vertex{v}$ at each timestamp $t$. The features for each timestamp are computed/learned independently of the other timestamps. Then, these features are aggregated into a single feature vector that can be fed into a decoder.

\citet{yaowangpan} aggregate features into a single feature vector as follows:
\begin{align} \label{eq:aggr-static-feats}
    \vctr{z}_{\vertex{v}} = \sum_{t=1}^T \function{exp}(-\theta (T-t)) \vctr{z}^t_\vertex{v}
\end{align} 
thus exponentially decaying older features. 
\citet{zhu2012hybrid} follow a similar strategy where they compute features for each pair of nodes and take a weighted sum (with prefixed weights) of the features, giving higher weights to the features coming from more recent snapshots. 

Rather than using an explicitly defined aggregator (e.g., exponential decay) that assigns prefixed weights to previous snapshots, one can fit a time-series model to the features from previous snapshots and use this model to predict the values of the features for the next snapshot. For the time-series model, \citet{huang2009time} and \citet{gunecs2016link} use the ARIMA model~(\cite{box2015time}), \citet{da2012time} use ARIMA and other models such as moving averages, and \citet{lats} use an approach based on some basic reinforcement learning.

\begin{example}
Consider the DTDG in Example~\ref{example:temporal-aggr}. A simple example of creating node embeddings by aggregating static features is to use the common neighbor static encoder (see Section~\ref{sec:high-order-encoder} for details) to obtain node embeddings at each timestamp and then combine these embeddings using Equation~\eqref{eq:aggr-static-feats}. The common neighbors encoder applied to $\mtrx{A}^1, \mtrx{A}^2$ and $\mtrx{A}^3$ gives the following embeddings:
\[
\mtrx{Z}^1 = \mtrx{S}_{CN}^1
=
\begin{bmatrix}
    1 & 0 & 1 \\
    0 & 2 & 0 \\
    1 & 0 & 1 
\end{bmatrix}
\quad
\mtrx{Z}^2 = \mtrx{S}_{CN}^2
=
\begin{bmatrix}
    2 & 1 & 1 \\
    1 & 2 & 1 \\
    1 & 1 & 2 
\end{bmatrix}
\quad
\mtrx{Z}^3 = \mtrx{S}_{CN}^3
=
\begin{bmatrix}
    2 & 0 & 0 \\
    0 & 1 & 1 \\
    0 & 1 & 1 
\end{bmatrix}
\]
In the above matrices, $\mtrx{Z}^t[i]$ corresponds to the embedding of $\vertex{v}_i$ at timestamp $t$. Notice how the embeddings for different timestamps are computed independently of the other timestamps. Also note that while we used the common neighbors encoder, any other static encoder from Section~\ref{sec:static-encoders} could potentially be used. Considering a value of $\theta=0.5$, the above embeddings are then aggregated into:
\[
\mtrx{Z}^{aggr}
\approx
\begin{bmatrix}
    3.58 & 0.61 & 0.97 \\
    0.61 & 2.95 & 1.61 \\
    0.97 & 1.61 & 2.58 
\end{bmatrix}
\]
where $\mtrx{Z}^{aggr}[i]$ corresponds to the embedding for $\vertex{v}_i$.
\end{example}

\textbf{Scalability:} Depending on the number of snapshots and the static encoder used for feature generation, the approaches that compute node features/embeddings at each snapshot independently of the other snapshots and then aggregate these features may be computationally expensive. In the upcoming subsections, for some choices of static encoders (e.g., for decomposition and random-walk approaches), we will see some techniques to save computations in later snapshots by leveraging the computations from the previous snapshots. 
\subsection{Time as a Regularizer} \label{sec:time-as-regularizer}
A common approach to leverage the temporal aspect of DTDGs is to use time as a regularizer to impose a smoothness constraint on the embeddings of each node over consecutive snapshots~(\cite{chakrabarti2006evolutionary,chi2009evolutionary,kim2009particle,gupta2011evolutionary,yaowangpan, zhu2016scalable,zhou2018dynamic}).
Consider a DTDG as $\{\graph{G}^1, \dots, \graph{G}^T\}$. For a node $\vertex{v}$, let $\embedding{EMB}^{t-1}(\vertex{v})=(\vctr{z}^{t-1}_{\vertex{v}})$ represent the vector representation learned for this node at the $\Th{(t-1)}$ snapshot. To learn the vector representation for \vertex{v} at the $\Th{t}$ snapshot, the approaches in this class typically use a static encoder to learn an embedding function for $\graph{G}^t$ with the additional constraint that for any node $\vertex{v}\in\vertices{V}^t$ such that $\vertex{v}\in\vertices{V}^{t-1}$ (i.e. for any node $\vertex{v}$ that has been in the graph in the previous and current snapshots), $\function{dist(\vctr{z}^{t-1}_{\vertex{v}}, \vctr{z}^{t}_{\vertex{v}})}$ should be small, where $\function{dist}$ is a distance function. This constraint is often called the \emph{smoothness} constraint.
A common choice for the distance function is the Euclidean distance:
\begin{align} \label{eq:time-as-reg}
    \function{dist}(\vctr{z}^{t}_{\vertex{v}}, \vctr{z}^{t-1}_{\vertex{v}})=||\vctr{z}^{t}_{\vertex{v}} - \vctr{z}^{t-1}_{\vertex{v}}||
\end{align}
but distance in other spaces may also be used (see, e.g., \cite{chi2009evolutionary}). \citet{singer2019node} add a rotation projection to align the embedding $\vctr{z}^{t}_\vertex{v}$s with the embedding $\vctr{z}^{t-1}_\vertex{v}$s before taking the Euclidean distance. Their distance function can be represented as follows:
\begin{align}
    \function{dist}(\vctr{z}^{t}_\vertex{v},\vctr{z}^{t-1}_\vertex{v})=||\mtrx{R}^{t}\vctr{z}^{t}_\vertex{v}-\vctr{z}^{t-1}_\vertex{v}||
\end{align}
where $\mtrx{R}^{t}$ is a rotation matrix. Instead of Euclidean distance, \citet{fard2019relationship} define the $\function{dist}$ function based on the angle between the two vectors. Their distance function can be written as follows:
\begin{align}
    \function{dist}(\vctr{z}^{t}_\vertex{v},\vctr{z}^{t-1}_\vertex{v})=1-(\vctr{z}^{t}_\vertex{v})^\prime \vctr{z}^{t-1}_\vertex{v}
\end{align}
where all embedding vectors are restricted to have a norm of $1$. Note that the smaller the angle between $\vctr{z}^{t}_\vertex{v}$ and $\vctr{z}^{t-1}_\vertex{v}$, the closer $(\vctr{z}^{t}_\vertex{v})^\prime \vctr{z}^{t-1}_\vertex{v}$ is to $1$ and so the closer $\function{dist}(\vctr{z}^{t}_\vertex{v},\vctr{z}^{t-1}_\vertex{v})=1-(\vctr{z}^{t}_\vertex{v})^\prime \vctr{z}^{t-1}_\vertex{v}$ is to $0$. \citet{liu2019streaming} also use time as a regularizer, but they turn the representation learning problem into a constrained optimization problem that can be approximated in a reasonable amount of time. As new observations are made, their representations can be updated in a short amount of time and so their model may be used for streaming scenarios. The model they propose also handles addition of new nodes to the graph. \citet{pei2016node} propose a dynamic factor graph model for node classification in which they use the temporal information in a similar way as the other approaches in this section: they impose factors that decrease the probability of the worlds where the label of a node at the $\Th{t}$ snapshot is different from the previous snapshots (exponentially decaying the importance of the labels for the older snapshots). Using time as a regularizer can be useful for both interpolation and extrapolation problems.

\begin{example}
Consider the DTDG in Example~\ref{example:temporal-aggr}. Suppose we want to provide node embeddings by using a static autoencoder approach (see Section~\ref{sec:static-autoencoder} for details) while using time as a regularizer. In the first timestamp, we train an autoencoder whose encoder takes as input $\mtrx{A}^1[i]$, feeds it through its encoder and generates $\vctr{z}^1_{\vertex{v}_i}$, and then feeds $\vctr{z}^1_{\vertex{v}_i}$ through its reconstructor to output $\hat{\mtrx{A}}^1[i]$ with the loss function being $\sum_{i=1}^{|\vertices{V}|} ||\mtrx{A}^1[i] - \hat{\mtrx{A}}^1[i]||$. In the second timestamp, we follow a similar approach but instead of the loss function being $\sum_{i=1}^{|\vertices{V}|} ||\mtrx{A}^2[i] - \hat{\mtrx{A}}^2[i]||$, we define the loss function to be $\sum_{i=1}^{|\vertices{V}|} ||\mtrx{A}^2[i] - \hat{\mtrx{A}}^2[i]||+||\vctr{z}^2_{\vertex{v}_i}-\vctr{z}^1_{\vertex{v}_i}||$. We continue a similar procedure in the third timestamp by defining the loss function as $\sum_{i=1}^{|\vertices{V}|} ||\mtrx{A}^3[i] - \hat{\mtrx{A}}^3[i]||+||\vctr{z}^3_{\vertex{v}_i}-\vctr{z}^2_{\vertex{v}_i}||$. Note that here we are using the distance function from Equation~\eqref{eq:time-as-reg} but other distance function can be used as well.
\end{example}

Imposing smoothness constraints through penalizing the distance between the vector representations of a node at consecutive snapshots stops the vector representation from having sharp changes. While this may be desired for some applications, in some other applications a node may change substantially from one snapshot to the other. As an example, if a company gets acquired by a large company, it is expected that its vector representation in the next snapshot makes sharp changes.  Instead of penalizing the distance of the vector representations for a node at consecutive snapshots, one may simply initialize the representations (or the model) for time $t$ with the learned representations (or model) at time $t-1$ and then allow the static encoder to further optimize the representation at time $t$ (see, e.g., \cite{goyal2017dyngem}). This procedure implicitly imposes the smoothness constraint while also allowing for sharp changes when necessary.

Another notable work where time is used as a regularizer is an extension of a well-known model for static graphs, named \emph{LINE}~(\cite{tang2015line}), to DTDGs by \citet{du2018dynamic}. 
Besides using time as regularizer, the authors propose a way of recomputing the node embeddings only for the nodes that have been influenced greatly from the last snapshot.

\subsection{Decomposition-based Encoders} \label{sec:temporal-decomposition}
A good application of decomposition methods to dynamic graphs is to use them as an alternative to aggregating temporal observations described in Section~\ref{sec:aggregate-temp-obs}. 
Consider a DTDG as $\{\graph{G}^1, \dots, \graph{G}^T\}$ such that $\vertices{V}^1=\vertices{V}^2=\dots=\vertices{V}^T=\vertices{V}$ (i.e. nodes are not added or removed). As was proposed by \cite{dunlavy2011temporal}, the adjacency matrices $\mtrx{A}^1, \dots, \mtrx{A}^T$ for $T$ timestamps can be stacked
into an order $3$ tensor   $\mathcal{A} \in \mathbb{R}^{|\vertices{V}|\times |\vertices{V}|\times T}$. Then one can do a $d$-component tensor decomposition (e.g., CP decomposition, see \cite{Harshman1970FOUNDATIONSOT}):
\begin{align}
\mathcal{A} \approx \sum_{k=1}^d \lambda_k \vctr{a}_k \otimes \vctr{b}_k \otimes \vctr{c}_k 
\end{align}
where $\lambda_k \in \mathbb{R}_+$, $\vctr{a}_k, \vctr{b}_k \in \mathbb{R}^{|\vertices{V}|}$, $\vctr{c}_k \in \mathbb{R}^T$, and $\otimes$ is a tensor product of vector spaces. The temporal pattern is captured in the $\vctr{c}_k$s, and a combination of $\vctr{a}_k$s and $ \vctr{b}_k$s can be used as the node (or edge) embeddings. These embeddings can be used to make predictions about any time $1\leq t\leq T$, i.e. for interpolation. For extrapolation, \citet{dunlavy2011temporal} used the Holt-Winters method (see, \cite{holtwinters}): given the input $\vctr{c}_k$, it predicts an $L$-dimensional vector $\vctr{c}'_k $, which is the prediction of the temporal factor for the next $L$ timesteps. Then they predict the adjacency tensor for the next $L$ snapshots as $\hat{\mathcal{A}} = \sum_{k=1}^d \lambda_k \vctr{a}_k \otimes \vctr{b}_k \otimes \vctr{c}'_k$. One can also use other forms of tensor decomposition, \emph{e.g.} Tucker decomposition or HOSVD~(\cite{tensor_decomposition}). \citet{xiong2010temporal} propose a probabilistic factorization of $\mathcal{A}$ where the nodes are represented as normal distributions with the means coming from $\vctr{a}_k$s and $\vctr{b}_k$s. They also impose a smoothness prior over the temporal vectors corresponding to using time as a regularizer (see Section~\ref{sec:time-as-regularizer}). After some time steps, one needs to update the tensor decomposition for more accurate future predictions. The recomputation can be quite costly so one can try incremental updates  (see \cite{gujral2018sambaten} and \cite{incremental-cp2018}).

\citet{yucheng} present another way of incorporating temporal dependencies into the embeddings with decomposition methods. As above, let $\mtrx{A}^1, \dots, \mtrx{A}^T$ be the adjacency matrices for $T$ timestamps. \citet{yucheng} predict $\hat{\mtrx{A}}^{T+l}$, where $l \in \mathbb{N}$, as follows. First, they solve the optimization problem:
\begin{align}
    \text{min} \sum_{t = T-\omega}^T e^{-\theta (T-t)}
    ||\mtrx{A}^t - \mtrx{U}\transpose{(\mtrx{V}^t)} \transpose{(\mtrx{P}^t)}||^2_F   ,
\end{align}
where $\mtrx{P}^t = (1-\alpha)(\mathbf{I} - \alpha \sqrt{\mtrx{D}^t} \mtrx{A}^t \sqrt{\mtrx{D}^t})^{-1} $ is the projection onto feature space that ensures neighboring nodes have similar feature vectors (see \cite{yucheng} for details), $\omega$ is a window of timestamps into consideration, $\alpha \in (0,1) $ is a regularization parameter, $\theta$ is a decay parameter, $\mtrx{U} \in \mathbb{R}^{|\vertices{V}|\times d}$ is a  matrix that does not depend on time, and $\mtrx{V}^t \in \mathbb{R}^{|\vertices{V}|\times d}$ is a matrix with explicit time dependency (in the paper, it is a polynomial in time with matrix coefficients). The optimization problem can be slightly rewritten using the sparsity of $\mtrx{A}$ and then solved with stochastic gradient descent. The prediction can be obtained as $\hat{\mtrx{A}}^{T+l} = \mtrx{U}\transpose{(\mtrx{V}^{T+l})} $. From the point of view of the encoder-decoder framework, $\mtrx{U} $ can be interpreted as static node features and $\mtrx{V}^t$s are time-dependent features (one takes the $\Th{i}$ row of the matrices as an embedding of the $\Th{i}$ node). 

One can extend the tensor decomposition idea to the case of temporal KGs by modelling the KG as an order 4 tensor $\mathcal{A}\in\mathbb{R}^{|\vertices{V}|\times|\relations{R}|\times|\vertices{V}|\times T}$ and decomposing it using CP, Tucker, or some other decomposition approach to obtain entity, relation, and timestamp embeddings (see \cite{tresp2015learning,esteban2016predicting,tresp2017embedding}). \citet{tresp2015learning} and \citet{tresp2017embedding} study the connection between these order 4 tensor decomposition approaches and the human cognitive memory functions.

\paragraph{The streaming scenario.} As was discussed in Subsection~\ref{decomposition}, one can learn node embedding using either eigen-decomposition or SVD for graph matrices for each timestamp. Then one can aggregate these features as in Section~\ref{sec:aggr-static-feats} for predictions. However, recalculating decomposition at every timestamp may be computationally expensive. So one needs to come up with incremental algorithms that will update the current state in the streaming case. 

Incremental eigenvalue decomposition~(\cite{chentong, li2017attributed, wanghezhou}) is based on perturbation theory. Consider a generalized eigenvalue problem as in Equation~\eqref{eq_eigen}. Then assume that in the next snapshot we add a few new edges to the graph $\graph{G}^T$. In this case, the Laplacian and the degree matrix change by a small amount: $\Delta \mtrx{L}$ and $\Delta \mtrx{D}$ respectively.   Assume that we have solved Equation~\eqref{eq_eigen} and $\{(\lambda_i, \vctr{y}_i)\}_{i=1}^{|\vertices{V}|}$ is the solution. Then one can find the solution to the new generalized eigenvalue problem for the graph $\graph{G}^{T+1}$ in the form: updated eigenvalues $ \approx \lambda + \Delta \lambda $ and updated eigenvectors $ \approx \vctr{y} + \Delta \vctr{y} $, where $\Delta \lambda$ and $\Delta \vctr{y}$ can be efficiently computed. For example,
\begin{align}
    \Delta \lambda_i = \frac{\transpose{\vctr{y}_i} \Delta \mtrx{L} \vctr{y}_i  - \lambda_i\transpose{\vctr{y}_i} \Delta \mtrx{D} \vctr{y}_i}{\transpose{\vctr{y}_i} \mtrx{D} \vctr{y}_i}.
\end{align}
An analogous formula could be written for $\Delta \vctr{y}_i$. The Davis-Kahan theorem~(\cite{daviskahan}) gives an approximation error for the top $d$ eigen-pairs. 

As shown by \citet{levin2018out}, one can recalculate the adjacency spectral embedding (see Section~\ref{decomposition} for the construction) in case of addition of a new node $\vertex{v}$ to a graph $\graph{G}$. Denote $\vctr{a}_\vertex{v} \in \mathbb{R}^{|\vertices{V}|}$ a binary vector where each entry indicates whether there is an edge between the added node and an already existing node. One can find $\vctr{z}_\vertex{v}$ as the solution to the maximum likelihood problem to fit $\vctr{a}_\vertex{v} \sim \text{Bernoulli}(\transpose{\mtrx{Z}} \vctr{z}_\vertex{v})$, where $\mtrx{Z}$ is as in Formula \eqref{ASE}. 

\citet{svdincremental} proposes an efficient way to update the singular value decomposition of a matrix $\mtrx{S}$ when another lower rank matrix of the same size $\Delta \mtrx{S}$ is added to it. Consider the problem in Equation~\eqref{svd}. If one knows the solution $(\mtrx{U}_s, \mtrx{U}_t, \mtrx{\Sigma}  )$ and  $\Delta \mtrx{S}$ is an update of the matrix, one can find a general formula for the update of the SVD using some basic computations with block matrices.  However, this becomes especially efficient if we approximate the increment as a rank one matrix: $\Delta \mtrx{S} = \vctr{a}\vctr{b}'$ (see also \cite{svd2}). \citet{svd3} studied how SVD can be updated when a row or column is added to or removed from a matrix $\mtrx{S}$. This can be applied to get the encoding for a DTDG in the case of node addition or deletion. 

One problem with incremental updates is that the approximation error keeps accumulating gradually. As a solution, one needs to recalculate the model from time to time. However, since the recalculation is expensive, one needs to find a proper time when the error becomes intolerable. Usually in applications people use heuristic methods (e.g., restart after a certain time); however, ideally a timing should depend on the graph dynamics.  \citet{zhang2018timers} propose a new method where given a tolerance threshold, it notifies at what timestamp the approximation error exceeds the threshold.  

\subsection{Random Walk Encoders}
Recently, several approaches have been proposed to leverage or extend the random walk models for static graphs to dynamic graphs. In this section, we provide an overview of these approaches.

Consider a DTDG as $\{\graph{G}^1, \dots, \graph{G}^T\}$. \citet{mahdavi2018dynnode2vec} first generate random walks on $\graph{G}^1$ similar to the random walk models on static graphs and then feed those random walks to a model $\model{M}^1$ that learns to produce vector representations for nodes given the random walks. For the $\Th{t}$ snapshot ($t>1$), instead of generating random walks from scratch, they keep the \emph{valid} random walks from $\Th{(t-1)}$ snapshot, where they define a random walk as valid if all its nodes and the edges taken along the walk are still in the graph in the $\Th{t}$ snapshot. They generate new random walks only
starting from the affected nodes, where affected nodes are the nodes that have been either added in this snapshot, or are involved in one or more edge addition or deletion. Having obtained the updated random walks, they initialize $\model{M}^t$ with the learned parameters from $\model{M}^{t-1}$ and then allow $\model{M}^t$ to be optimized and produce the node embeddings for the  $\Th{t}$ snapshot.

\citet{bian2019network} take a strategy similar to that of \citet{mahdavi2018dynnode2vec} but for KGs. They use metapath2vec (explained in Section~\ref{static-random-walk-sec}) to generate random walks on the initial KG. Then, at each snapshot, they use metapath2vec to generate random walks for the affected nodes and re-compute the embeddings for these nodes.

\citet{sajjad2019efficient} observed that by keeping the valid random walks from the previous snapshot and naively generating new random walks starting from the affected nodes, the resulting random walks may be biased. That is, the random walks obtained by following this procedure may have a different distribution than generating random walks for the new snapshot from scratch. Example~\ref{random-walk-bias-example} demonstrates one such example. 

\begin{example}\label{random-walk-bias-example}
Consider Figure~\ref{graphs-fig}(a) as the first snapshot of a DTDG and assume the following random walks have been generated for this graph (two random walks starting from each node) following a uniform transition:
\begin{align*}
    1)~\vertex{v}_1, \vertex{v}_2, \vertex{v}_1 \quad\quad\quad
    2)~\vertex{v}_1, \vertex{v}_2, \vertex{v}_3 \quad\quad\quad
    3)~\vertex{v}_2, \vertex{v}_1, \vertex{v}_3 \\
    4)~\vertex{v}_2, \vertex{v}_3, \vertex{v}_1 \quad\quad\quad
    5)~\vertex{v}_3, \vertex{v}_2, \vertex{v}_1 \quad\quad\quad
    6)~\vertex{v}_3, \vertex{v}_1, \vertex{v}_2
\end{align*}
Now assume the graph in Figure~\ref{graphs-fig}(b) represents the next snapshot. The affected nodes are $\vertex{v}_2$, which has a new edge, and $\vertex{v}_4$, which has been added in this snapshot. A naive approach for updating the above set of random walks is to remove random walks 3 and 4 (since they start from an affected node) and add two new random walks from $\vertex{v}_2$ and two from $\vertex{v}_4$. This may give the following eight walks:
\begin{align*}
    1)~\vertex{v}_1, \vertex{v}_2, \vertex{v}_1 \quad\quad\quad
    2)~\vertex{v}_1, \vertex{v}_2, \vertex{v}_3 \quad\quad\quad
    3)~\vertex{v}_2, \vertex{v}_4, \vertex{v}_2 \quad\quad\quad
    4)~\vertex{v}_2, \vertex{v}_3, \vertex{v}_1 \\
    5)~\vertex{v}_3, \vertex{v}_2, \vertex{v}_1 \quad\quad\quad
    6)~\vertex{v}_3, \vertex{v}_1, \vertex{v}_2 \quad\quad\quad
    7)~\vertex{v}_4, \vertex{v}_2, \vertex{v}_3 \quad\quad\quad
    8)~\vertex{v}_4, \vertex{v}_2, \vertex{v}_1 
\end{align*}
In the above $8$ random walks, the number of times a transition from $\vertex{v}_2$ to $\vertex{v}_4$ has been made is $1$ and the number of times a transition from $\vertex{v}_2$ to $\vertex{v}_1$ (or $\vertex{v}_3$) has been made is $3$, whereas, if new random walks are generated from scratch, the two numbers are expected to be the same. The reason for this bias is that in random walks 1, 2, 5, and 6, the walk could not go from $\vertex{v}_2$ to $\vertex{v}_4$ as $\vertex{v}_4$ did not exist when these walks were generated. Note that performing more random walks from each node does not solve the bias problem.
\end{example}

\citet{sajjad2019efficient} propose an algorithm for generating unbiased random walks for a new snapshot while reusing the valid random walks from the previous snapshot. NetWalk~(\cite{yu2018netwalk}) follows a similar approach as the previous two approaches. However, rather than relying on natural language processing techniques to generate vector representations for nodes given random walks, they develop a customized autoencoder model that learns the vector representations for nodes while minimizing the pairwise distance among the nodes in each random walk.

The previous three approaches mainly leverage the temporal aspect of DTDGs to reduce the computations. They can be useful in the case of feature aggregation (see Section~\ref{sec:aggr-static-feats}) when random walk encoders are used to learn features at each snapshot. However, they may fail at capturing the evolution and the temporal patterns of the nodes. \citet{nguyen2018continuous,nguyen2018dynamic} propose an extension of the random walk models for CTDGs that also captures the temporal patterns of the nodes. 

Consider a CTDG as $(\graph{G}, \events{O})$ where the only type of event in $\events{O}$ is the addition of new edges. Therefore, the nodes are fixed and each element of \events{O} can be represented as $(AddEdge, (\vertex{v}, \vertex{u}), t_{(\vertex{v}, \vertex{u})})$ indicating an edge was added between \vertex{v} and \vertex{u} at time $t_{(\vertex{v}, \vertex{u})}$. \citet{nguyen2018continuous,nguyen2018dynamic} constrain the random walks to respect time, where they define a random walk on a CTDG that respects time as a sequence $\vertex{v}_1, \vertex{v}_2, \dots, \vertex{v}_l$ of nodes where:
\begin{align}
    \vertex{v}_i\in\vertices{V}&~\text{for all}~1\leq i\leq l \\
     (AddEdge, (\vertex{v}_i, \vertex{v}_{i+1}), t_{(\vertex{v}_i, \vertex{v}_{i+1})}) \in \events{O}&~\text{for all}~1\leq i \leq l-1\\
     t_{(\vertex{v}_i, \vertex{v}_{i+1})} \leq t_{(\vertex{v}_{i+1}, \vertex{v}_{i+2})}&~\text{for all}~ 1\leq i \leq l-2
\end{align}
That is, the sequence of edges taken by each random walk only moves forward in time. Similar to the random walks on static graphs, the initial node to start a random walk from and the next node to transition to can come from a distribution. Unlike the static graphs, however, these distributions can be a function of time. For instance, consider a walk that has currently reached a node $\vertex{u}$ by taking an edge $(\vertex{v}, \vertex{u})$ that has been added at time $t$. The edge for the next transition (to be selected from the outgoing edges of $\vertex{u}$ that have been added after $t$) can be selected with a probability proportional to how long after $t$ they were added to the graph. 

\begin{example}
Assume $t_1 < t_2 < t_3 < t_4 < t_5$ for the CTDG in Figure~\ref{graphs-fig}(d). Consider a random walk that has started at $\vertex{v}_1$, then transitioned to $\vertex{v}_2$, and is now deciding the next node to transition to. According to \citet{nguyen2018continuous}'s strategy, even though both $\vertex{v}_4$ and $\vertex{v}_5$ are neighbors of $\vertex{v}_2$, only the transition to $\vertex{v}_4$ is valid as the edge between $\vertex{v}_2$ and $\vertex{v}_4$ has been added after the edge between $\vertex{v}_1$ and $\vertex{v}_2$ whereas the edge between $\vertex{v}_2$ and $\vertex{v}_5$ has been added before the edge between $\vertex{v}_1$ and $\vertex{v}_2$.
\end{example}

\citet{de2018combining} follow a similar approach as \cite{nguyen2018continuous} but for DTDGs instead of CTDGs. Their experiments show that in some cases, discretizing a CTDG into a DTDG and then running the random walks that respect time on the DTDG results in better performance. \citet{bastas2019evolve2vec} also follow a similar approach as \cite{nguyen2018continuous}, but they divide the time horizon into two intervals one corresponding to observations before some time $t_s$ and the other corresponding to the observations after $t_s$. They aggregate the observations until time $t_s$ into a static graph (see Section~\ref{sec:aggregate-temp-obs}) following the intuition that older observations mainly contain topological information and not temporal information. They run static random walks on the static graph from the first interval and temporal random walks that respect time (with custom distributions for selecting the initial node and the node to transition to) on the second interval. Both the static and temporal walks are then used to learn node embeddings.

\subsubsection{Analysis of Random Walk Encoders}
\textbf{Supervised and unsupervised learning:} Similar to decomposition and autoencoder-based approaches, one of the major advantages of the encoders based on random walks is that they provide an embedding function without needing to be combined with a decoder. Therefore, the encoder can be used for unsupervised learning approaches such as clustering and community detection~(\cite{xin2016adaptive,yu2017overlapping}). However, the disconnect between the encoder and the decoder typically prevents these models from being trained end-to-end. Therefore, for supervised prediction tasks, the embedding learned for nodes are not optimized for the prediction problem.

\textbf{The streaming scenario:} When new observations are made, random walk approaches typically require to perform new walks that take the new observations into account and then update the node embeddings based on the new walks. This update usually requires a few rounds of computing gradients, which, depending on the size of the dynamic graph, can be quite time-consuming. This makes random walk approaches not an ideal option for the streaming scenario.

\textbf{Random walk for attributed graphs:} Using random walks for representation learning has been mostly done for non-attributed graphs. 
An interesting direction for future research is to extend the approaches discussed in this section to attributed graphs.

\subsection{Sequence-Model Encoders}
A natural choice for modeling dynamic graphs is by extending sequence models to graph data. With the success of RNNs in several synchronous sequence modeling problems~(\cite{mikolov2010recurrent,bahdanau2014neural,hermann2015teaching,mesnil2015using,huang2015bidirectional,seo2016bidirectional}), where the duration between any two consecutive items in the sequence is considered equal, and several asynchronous sequence modeling problems~(\cite{choi2016doctor,li2017time,du2016recurrent,neil2016phased,zhu2017next,hu2017state}),  RNNs have been a common choice for extending sequence models to DTDGs and CTDGs. In the next subsections, we describe the RNN-based models for DTDGs, which can be considered a synchronous sequence modeling problem, and CTDGs, which can be considered an asynchronous sequence modeling problem. We also describe some other sequence modeling approaches that have been extended to dynamic graphs. Note that sequence model approaches are mainly designed for extrapolation as they go through the observations sequentially and provide embeddings at the current time based on the past. However, one may use them for interpolation by, e.g., using bi-directional sequence models one running forward and providing embeddings based on everything before some time $t$ and the other running backward and providing embeddings based on everything after time $t$. 

\subsubsection{RNN-based Encoders for DTDGs} \label{sec:rnn-dtdg}
Consider a DTDG as $\{\graph{G}^1, \graph{G}^2, \dots, \graph{G}^T\}$. Let $\model{M}$ be a (differentiable) encoder, which, given a static graph $\graph{G}^t$, outputs a vector representation for each node. As an example, $\model{M}$ can be a GCN (see Section~\ref{sec:gcn} for details).

One way of leveraging RNNs for DTDGs is as follows. We run $\model{M}$ on each $\graph{G}^t$ and obtain a sequence $\vctr{z}^1_{\vertex{v}}, \vctr{z}^2_{\vertex{v}}, \dots, \vctr{z}^T_{\vertex{v}}$ of vector representations for each node $\vertex{v}$. This sequence is then fed into an RNN that produces a vector representation $\vctr{z}_{\vertex{v}}$ for $\vertex{v}$ containing information from \vertex{v}'s history and evolution. These vector representations of the nodes can then be fed into a decoder to make predictions about the nodes. The idea behind this approach is similar to the static feature aggregation idea described in Section~\ref{sec:aggr-static-feats} except that the weights of the RNN and the model $\model{M}$ are learned simultaneously and over all the snapshots. The $\Th{t}$ step of the encoder for this architecture can be represented as:
\begin{align}
    \vctr{z}^t_{\vertex{v}_1}, \dots, \vctr{z}^t_{\vertex{v}_{|\vertices{V}^t|}} &= \model{M}(\graph{G}^t)\\
    \vctr{h}^t_{\vertex{v}_j} &= \function{RNN}(\vctr{h}^{t-1}_{\vertex{v}_j}, \vctr{z}^t_{\vertex{v}_j})~\text{for}~j\in [1,|\vertices{V}^t|]
\end{align}
which can be equivalently represented as:
\begin{align}
    \mtrx{Z}^t &= \model{M}(\graph{G}^t)\\
    \mtrx{H}^t &= \function{RNN}(\mtrx{H}^{t-1}, \mtrx{Z}^t)
\end{align}
where $\mtrx{Z}^t\in\mathbb{R}^{|\vertices{V}^t|\times d}$ represents the vector representations of size $d$ for the $|\vertices{V}^t|$ nodes in the graph at snapshot $t$ and $\mtrx{H}^t\in\mathbb{R}^{|\vertices{V}^t|\times d}$ represents the hidden state of the RNN corresponding to vector representations of size $d$ for the $|\vertices{V}^t|$ nodes in the graph that captures the history of the nodes as well. In this architecture, $\model{M}$ aims at capturing the structural information for each node at each snapshot and the RNN aims at capturing the temporal information. The approach described above has been proposed and used in different works. Model 1 of \cite{seo2018structured} uses this approach where $\model{M}$ is the GCN proposed by \cite{defferrard2016convolutional} and the RNN is a standard LSTM. \citet{narayan2018learning} also use this approach with $\model{M}$ being the GCN proposed by \cite{niepert2016learning} and the RNN being a standard LSTM. \citet{manessi2017dynamic} modify this approach slightly by (mainly) adding skip-connections in the GCN part. Another similar architecture is proposed by \cite{mohantygraph}. Instead of obtaining $\vctr{z}^t_{\vertex{v}_j}$s by running a GCN that aggregates the features of neighboring nodes only at the $\Th{t}$ snapshot, \citet{yu20193d} propose a 3D GCN that aggregates the features of  neighboring nodes on a window of previous snapshots (i.e., the aggregation is both spatial and temporal rather than just being spatial). 

In the above approach, $\model{M}$ is independent of the RNN. That is, the vector representations for nodes provided by $\model{M}$ are independent of the node histories captured in $\vctr{h}^t_{\vertex{v}_j}$s. The \emph{embedded approaches} aim at embedding the model(s) $\model{M}$ into the RNN so that $\model{M}$ can also use the node histories. 

One such embedded approach has been proposed by \cite{chen2018gc}, where the authors combine the GCN proposed by \cite{defferrard2016convolutional} with LSTMs. Consider a DTDG as $\{\graph{G}^1, \graph{G}^2, \dots, \graph{G}^T\}$, let $\mtrx{A}^t$ represent the adjacency matrix for $\graph{G}^t$, and let $\mtrx{A}^t[j]$ represent the $\Th{j}$ row of $\mtrx{A}^t$ corresponding to the neighborhood of node $\vertex{v}_j$. Let $\mtrx{C}^{t-1}[j]$ and $\mtrx{H}^{t-1}[j]$ represent the memory and hidden state of the LSTM at time $t-1$ for node $\vertex{v}_j$. Let $\model{M}_1$, $\model{M}_2$, $\model{M}_3$, $\model{M}_4$ and $\model{M}_5$ be five GCN models (same model with different parameters), where the node representations for $\model{M}_1$, $\model{M}_2$, $\model{M}_4$ and $\model{M}_5$ are initialized according to $\mtrx{H}^{t-1}$ and for $\model{M}_3$ are initialized according to $\mtrx{C}^{t-1}$. Let $\model{M}_i(\graph{G})[j]$ represent the vector representation provided by $\model{M}_i$ for node $\vertex{v}_j$ when applied to $\graph{G}$. The embedded model of \cite{chen2018gc} can be formulated as:
\begin{align}
\vctr{i}^t &~= \sigma\left(\mtrx{W}_{ii}
\mtrx{A}^t[j]+\model{M}_1(\graph{G}^t)[j]+\vctr{b}_i\right)\\
\vctr{f}^t &~= \sigma\left(\mtrx{W}_{fi}\mtrx{A}^t[j]+\model{M}_2(\graph{G}^t)[j]+\vctr{b}_f\right)\\
\mtrx{C}^t[j] &~= \vctr{f}_t\odot\model{M}_3(\graph{G}^t)[j] + \vctr{i}^t\odot \function{Tanh}\left(\mtrx{W}_{ci} \mtrx{A}^t[j]+\model{M}_4(\graph{G}^t)[j]+\vctr{b}_c\right) \\
\vctr{o}^t &~= \sigma\left(\mtrx{W}_{oi} \mtrx{A}^t[j]+\model{M}^5(\graph{G}^t)[j]+\vctr{b}_o\right)\\
\mtrx{H}^t[j] &~= \vctr{o}^t\odot\function{Tanh}\left(\mtrx{C}^t[j]\right)
\end{align}
where the above formulation is done for all $1 \leq j \leq |\vertices{V}^t|$. The $\model{M}_i$ models are embedded into the LSTM gates and use its memory and hidden state  in their computations. The above formulation can be considered as a standard LSTM taking a sequence of adjacency matrices as input, where the gate computations for the memory and hidden states have been replaced with GCN operations to take the graph structure into account. Other similar ways of embedding the $\model{M}$ models into RNNs can be seen in \cite{li2017diffusion} and \cite{seo2018structured}. Instead of embedding a GCN into an RNN, \citet{pareja2019evolvegcn} embed an RNN into a GCN by running a GCN with different parameters at each snapshot where the RNN provides the weights of the GCN at the $\Th{t}$ snapshot based on the weights of the GCNs at the previous snapshots.

\subsubsection{Other Sequence-Model Encoders for DTDGs} \label{sec:attention-encoder}
Besides RNNs, other sequence models have been also used as encoders for DTDGs. \citet{sarkar2007latent} use Kalman filters to track the embeddings of the nodes through time for a bipartite graph. Each timestep of the Kalman filter corresponds to a snapshot of the DTDG. The $\Th{t}$ timestep observes the adjacency matrix $\mtrx{A}^t$ corresponding to the $\Th{t}$ snapshot $\graph{G}^t$ of the DTDG and updates the node embeddings accordingly. The observations for each element $\mtrx{A}^t$ are considered to be independent of each other and $\mtrx{A}^t[i][j]$ is defined to be proportional to the distance between the embedding vectors of $\vertex{v}_i$ and $\vertex{v}_j$. To make the computations tractable, this conditional probability of the observation (i.e., the adjacency matrix) given the hidden state (i.e., the embeddings) is approximated by a Gaussian.

Inspired by fully attentive models (see Section~\ref{sec:sequence-models}), \citet{sankar2018dynamic} propose DySAT: a fully attentive model for DTDGs. Consider a DTDG as $\{\graph{G}^1, \dots, \graph{G}^T\}$.  DySAT applies the attention-based GCN model of \cite{velivckovic2018graph} to each $\graph{G}^t$ and obtains $\vctr{z}^1_{\vertex{v}}, \dots, \vctr{z}^T_{\vertex{v}}$ for every node $\vertex{v}$, where $\vctr{z}^t_{\vertex{v}}$ encodes the structural information from $\graph{G}^t$. Similar to \citet{vaswani2017attention}, DySAT adds a positional embedding $\vctr{p}^t$ to each $\vctr{z}^t_{\vertex{v}}$ and obtains $\overline{\vctr{z}}^t_{\vertex{v}}=\vctr{z}^t_{\vertex{v}}+\vctr{p}^t$, where each $\vctr{p}^t$ encodes information about the relative position of the $\Th{t}$ snapshot compared to other snapshots. Finally, DySAT applies a multi-head self-attention on $\overline{\vctr{z}}^1_{\vertex{v}}, \dots, \overline{\vctr{z}}^T_{\vertex{v}}$ to get the final representation of the node to be sent to the decoder. 

\citet{kazemi2019time2vec} extend positional encoding to continuous time encoding through a vector representation for time dubbed \emph{Time2Vec}. An interesting direction for future research is to extend DySAT to CTDGs by replacing positional encoding with Time2Vec. 

\subsubsection{RNN-based Encoders for CTDGs} \label{sec:rnn-encoders-ctdg}
RNN-based approaches for CTDGs ~(\cite{kumar2018learning,trivedi2017know,ma2018dynamic,trivedi2019dyrep}) mainly consider CTDGs where the only possible observation is addition of new edges. They define custom RNNs that update the representations of the source and target nodes forming a new edge (and the representation of the relation between the two nodes in the case of KGs) upon making a new observation $(AddEdge, (\vertex{v}, \vertex{u}), t)$ (or $(AddEdge, (\vertex{v}, \relation{r}, \vertex{u}), t)$ in the case of a KG). One of the main differences in these approaches is in the way they define the embedding function and the way they define their custom RNN.

\citet{kumar2018learning}, for instance, consider bipartite graphs and develop a model named \emph{JODIE} which defines $\embedding{EMB}(\vertex{v})=(\vctr{z}_\vertex{v}, \overline{\vctr{z}}_\vertex{v})$ for each node $\vertex{v}$ in the graph where $\vctr{z}_\vertex{v}\in\mathbb{R}^{d_1}$ and $\overline{\vctr{z}}_\vertex{v}\in\mathbb{R}^{d_2}$. The values of $\vctr{z}_\vertex{v}$ are optimized directly (similar to shallow encoders), but the values of $\overline{\vctr{z}}_\vertex{v}$ are updated using an RNN. In JODIE, there are two different RNNs for updating the source and the target nodes. Upon making a new observation $(AddEdge, (\vertex{v}, \vertex{u}), t)$, the two RNNs update $\overline{\vctr{z}}_\vertex{v}$ and $\overline{\vctr{z}}_\vertex{u}$ as follows:
\begin{align}
    \overline{\vctr{z}}_\vertex{v} = \function{RNN}_{source}(\overline{\vctr{z}}_\vertex{v}, [\overline{\vctr{z}}_\vertex{u}; \Delta t_\vertex{v}; \vctr{f}]) \\
    \overline{\vctr{z}}_\vertex{u} = \function{RNN}_{target}(\overline{\vctr{z}}_\vertex{u}, [\overline{\vctr{z}}_\vertex{v}; \Delta t_\vertex{u}; \vctr{f}]) 
\end{align}
where $\Delta t_\vertex{v}$ represents the time elapsed since $\vertex{v}$'s previous interaction (similarly for $\Delta t_\vertex{u}$), $\vctr{f}$ represents edge features (e.g., it can be the rating a user assigned to a movie), and $[\overline{\vctr{z}}_\vertex{v}; \Delta t_\vertex{u}; \vctr{f}]$ represents the concatenation of $\overline{\vctr{z}}_\vertex{v}$, $\Delta t_\vertex{u}$ and $\vctr{f}$. $\function{RNN}_{source}$ is a standard RNN that takes as input the current state $\overline{\vctr{z}}_\vertex{v}$ and a new input $[\overline{\vctr{z}}_\vertex{u}; \Delta t_\vertex{v}; \vctr{f}]$, and outputs an updated state for $\overline{\vctr{z}}_\vertex{v}$; similarly for $\function{RNN}_{target}$. Note that the two functions are executed simultaneously thus $\overline{\vctr{z}}_\vertex{v}$ is updated based on the previous value of $\overline{\vctr{z}}_\vertex{u}$ and $\overline{\vctr{z}}_\vertex{u}$ is updated based on the previous value of $\overline{\vctr{z}}_\vertex{v}$. The two vectors $\vctr{z}_\vertex{v}$ and $\overline{\vctr{z}}_\vertex{v}$ for a node $\vertex{v}$ are then concatenated as one vector and sent to a decoder for making predictions.

\citet{trivedi2017know} consider KGs and define $\embedding{EMB}(\vertex{v})=(\vctr{z}_\vertex{v})$ for every node $\vertex{v}$ where $\vctr{z}_\vertex{v}\in\mathbb{R}^{d_1}$ and $\embedding{EMB}(\relation{r})=(\vctr{z}_\vertex{r})$ for every relation $\relation{r}$ where $\vctr{z}_\vertex{r}\in\mathbb{R}^{d_2}$. Their model, named \emph{Know-Evlove}, defines two custom RNNs that update $\vctr{z}_\vertex{v}$ and $\vctr{z}_\vertex{u}$ upon making a new observation $(AddEdge, (\vertex{v}, \relation{r}, \vertex{u}), t)$ as follows:
\begin{align}
    \vctr{z}_\vertex{v} = \function{Tanh}(\mtrx{W}_s\Delta t_\vertex{v}+\mtrx{W}_{hh}\function{Tanh}(\mtrx{W}_h[\vctr{z}_\vertex{v};\vctr{z}_\vertex{u};\vctr{r}_{p_\vertex{v}}])) \\
    \vctr{z}_\vertex{u} = \function{Tanh}(\mtrx{W}_t\Delta t_\vertex{u}+\mtrx{W}_{hh}\function{Tanh}(\mtrx{W}_h[\vctr{z}_\vertex{u};\vctr{z}_\vertex{v};\vctr{r}_{p_\vertex{u}}]))
\end{align}
where $\Delta t_\vertex{v}$ and $\Delta t_\vertex{u}$ are defined as before, $\vctr{r}_{p_\vertex{v}}$ is the vector representation for the last relation that $\vertex{v}$ was involved in (similarly for $\vctr{r}_{p_\vertex{u}}$), $\mtrx{W}_s\in\mathbb{R}^{d_1\times 1}$, $\mtrx{W}_t\in\mathbb{R}^{d_1\times 1}$, $\mtrx{W}_{h}\in\mathbb{R}^{l\times (2d_1+d_2)}$, and $\mtrx{W}_{hh}\in\mathbb{R}^{d_1\times l}$ are weight matrices, and $[\vctr{z}_\vertex{u};\vctr{z}_\vertex{v};\vctr{r}_{p_\vertex{u}}]$ is the concatenation of $\vctr{z}_\vertex{u}$, $\vctr{z}_\vertex{v}$ and $\vctr{r}_{p_\vertex{u}}$. The vector representation for relations is optimized directly (similar to shallow encoders). Compared to JODIE, Know-Evolve may be more influenced by $\Delta t$s because Know-Evolve projects each $\Delta t$ to a vector and sums the resulting vector with the influence coming from the representations of source, target, and relation. Unlike JODIE, the dependence of the update rules on $\Delta t$s in Know-Evolve is somewhat separate from the two nodes and the relation involved. 

\citet{trivedi2019dyrep} develop a model that can be used for several types of graphs. They define $\embedding{EMB}(\vertex{v})=(\vctr{z}_\vertex{v})$. Upon making a new observation $(AddEdge, (\vertex{v}, \vertex{u}), t)$, they update the node representation for \vertex{v} using the following custom RNN (and similarly for \vertex{u}):
\begin{align}
    \vctr{z}_\vertex{v} = \phi(\mtrx{W}_1 \vctr{z}_{\neighbour(\vertex{u})} + \mtrx{W}_2 \vctr{z}_\vertex{v} + \mtrx{W}_3 \Delta t_\vertex{v})
\end{align}
where $\vctr{z}_{\neighbour(\vertex{u})}$ is a weighted aggregation of the neighbors of \vertex{u}, $\Delta t_\vertex{v}$ is defined as before, $\mtrx{W}_i$s are weight matrices, and $\phi$ is an activation function. The aggregation $\vctr{z}_{\neighbour(\vertex{u})}$ can be different for different types of graphs (e.g., it can take relations into account in the case of a KG). \citet{trivedi2019dyrep} define a temporal attention mechanism to obtain the neighbor weights for $\vctr{z}_{\neighbour(\vertex{u})}$ at each time.

Other ways of defining the embedding function as well as custom RNNs can be viewed in \cite{ma2018dynamic} for directed graphs, \cite{dai2016deep} for bipartite graphs, and \cite{jin2019recurrent} for interpolation in KGs.

\subsubsection{Discussion and Analysis of RNN-based Encoders}
\textbf{Information propagation:}
Upon observing $(AddEdge, (\vertex{v}, \vertex{u}), t)$ (or $(AddEdge, (\vertex{v}, \relation{r}, \vertex{u}), t)$ in the case of a KG), many existing works only update the nodes directly involved in the new edge. \citet{ma2018dynamic} argue that it is important to propagate this information to the neighboring nodes so that they can update their representations accordingly. Towards this goal, they first compute a vector representation for the new observation as follows:
\begin{align}
    \vctr{z}_{o}=\phi(\mtrx{W}_1\vctr{z}_{s_\vertex{v}} + \mtrx{W}_2\vctr{z}_{t_\vertex{u}} + \vctr{b})
\end{align}
where $\vctr{z}_{s_\vertex{v}}$ and $\vctr{z}_{t_\vertex{u}}$ belong to $\embedding{EMB}(\vertex{v})$ and $\embedding{EMB}(\vertex{u})$ respectively and $\mtrx{W}_1$, $\mtrx{W}_2$ and $\vctr{b}$ are learnable parameters. $\vctr{z}_{o}$ is then  sent to the immediate neighbors of \vertex{v} and \vertex{u} and custom RNNs update the representation of the neighbors based on $\vctr{z}_{o}$ and based on how they are connected to $\vertex{v}$ and/or \vertex{u}.

\textbf{Attributed graphs:} For attributed graphs where nodes have attributes with fixed values, one way to incorporate these attributes into the model is by initializing (part of) the node representations using their attribute values~(\cite{trivedi2019dyrep}). For the case where the attribute values can change over time as well, \citet{seo2018structured} and \citet{feng2018temporal} develop models that take such changes into account for DTDGs. Developing models for attributed CTDGs where the attribute values can change over time is an interesting direction for future research.

\textbf{The streaming scenario:}
For the RNN-based approaches for CTDGs, once the RNN weights are learned during training, the RNN has learned how to take an observation as input and update the node (and relation) embeddings without requiring to compute any further gradients. That is, after training, the RNN weights can be freezed and used for updating the representations as new observations arrive. This makes RNN-based approaches a natural choice for the streaming scenario. Although as the amount of data collected during the test (freezed) time increases (e.g., when it reaches some predefined threshold), the training can run again on all the collected data to learn better weights for the RNN, then the weights can be frozen again and the updated RNN can replace the old one.  

\subsection{Autoencoder-based Encoders}
Consider a DTDG as $\{\graph{G}^1, \dots, \graph{G}^T\}$ and let $\mtrx{A}^1, \dots, \mtrx{A}^T$ be the corresponding adjacency matrices. \citet{goyal2017dyngem} learn an autoencoder $AE^1$ for $\graph{G}^1$ similar to SDNE where the encoder takes as input $\mtrx{A}^1[i]$ and generates a vector representation $\vctr{z}^1_{\vertex{v}_i}$ for node $\vertex{v}_i$. The reconstructor takes $\vctr{z}^1_{\vertex{v}_i}$ as input and reconstructs $\mtrx{A}^1[i]$. $\vctr{z}^1_{\vertex{v}_i}$ and $\vctr{z}^1_{\vertex{v}_j}$ are constrained to be close together if there is an edge between $\vertex{v}_i$ and $\vertex{v}_j$. Having $AE^1$, the embedding function for the first snapshot is defined as $\embedding{EMB}^1(\vertex{v}_i)=(\vctr{z}^1_{\vertex{v}_i})$. For the $\Th{t}$ snapshot ($t>1$), an autoencoder $AE^{t}$ is initialized with the weights from $AE^{t-1}$ and then trained based on $\mtrx{A}^t$ to produce the vector representations for nodes at snapshot $t$. $AE^{t}$ can have a different size (e.g., different number of neurons or layers) compared to $AE^{t-1}$. The authors decide the size of $AE^{t}$ based on heuristic methods that take into account the size of $AE^{t-1}$ and how different $\graph{G}^t$ is from $\graph{G}^{t-1}$. If the size of $AE^{t}$ is different from $AE^{t-1}$, in order to still be able to initialize $AE^{t}$ according to $AE^{t-1}$, the authors use the Net2WiderNet and Net2DeeperNet approaches from \citet{chen2015net2net}, which change the number of neurons and the number of layers in an autoencoder without substantially changing the function it computes.

The approach of \citet{goyal2017dyngem} uses the information within previous snapshots of a DTDG to enable learning an autoencoder for the current snapshot faster. Furthermore, initializing $AE^t$ according to $AE^{t-1}$ implicitly acts as a regularizer imposing a smoothness constraint (see Section~\ref{sec:time-as-regularizer}). However, the embeddings learned in their approach may not capture the evolution of the nodes. To better capture the evolution of the nodes,  
\citet{bonner2018temporal} propose to develop autoencoders that reconstruct a node's neighborhood in the next snapshot(s) given the current snapshot. They use a two-layer GCN as the encoder and dot-product as the reconstructor of the autoencoder. 
The authors also propose a variational autoencoder model where instead of directly learning the embeddings $\mtrx{Z}^t$ in the encoder, they learn a Gaussian distribution from which $\mtrx{Z}^t$ is sampled. The Gaussian distribution is parameterized by a mean vector $\pmb{\mu}$ and a variance vector $\pmb{\gamma}$
that are learned using two separate two-layer GCNs with tied parameters on the first layers.

To take more snapshots into account in learning node embeddings, \citet{goyal2018dyngraph2vec} propose to learn a single autoencoder where at snapshot $t$, the encoder takes as input $\mtrx{A}^{t-l}[i], \mtrx{A}^{t-l+1}[i], \dots, \mtrx{A}^{t-1}[i]$ and produces a vector $\vctr{z}^t_{\vertex{v}_i}$ corresponding to the embedding of $\vertex{v}_i$ at time $t$, and the reconstructor takes as input $\vctr{z}^t_{\vertex{v}_i}$ and reconstructs $\mtrx{A}^t[i]$. \citet{goyal2017dyngem} propose several ways for modeling the encoder and reconstructor. Examples for the encoder include feeding a concatenation $[\mtrx{A}^{t-l}[i];\mtrx{A}^{t-l+1}[i];\dots;\mtrx{A}^{t-1}[i]]$ into a feed-forward neural network or feeding a sequence $\mtrx{A}^{t-l}[i],\mtrx{A}^{t-l+1}[i],\dots,\mtrx{A}^{t-1}[i]$ into an LSTM. Similar architectures are used for the reconstructor.

\citet{rahman2016link} also follow an autoencoder-based approach by mapping each node-pair (instead of each node) to a hidden representation, which can then be used to predict addition or deletion of edges in the next snapshot. Towards this goal, for each snapshot they compute features for each pair of nodes based on the graphlet transitions (see \cite{prvzulj2004modeling,prvzulj2007biological}). Then they concatenate the features for each node-pair from the previous snapshots (similar to \cite{goyal2017dyngem}) and feed the concatenation to an autoencoder that learns a vector representation for the node-pair.

\subsection{Diachronic Encoders} \label{sec:diachronic}
\emph{Diachronic encoders} are encoders that map every pair (node, timestamp) or every pair (relation, timestamp) to a hidden representation. Note that this is different from Definition~\ref{dfnt:encoder} where encoders map every node or every relation to a hidden representation. Such encoders can be used effectively for interpolation as they learn to provide node and relation embeddings at any point in time.  \citet{goel2020diachronic} propose a diachronic encoder for nodes of a KG in which each vector $\vctr{z}_\vertex{v}^t\in\mathbb{R}^d$ in the embedding of $\vertex{v}$ is defined as follows:
\begin{equation}
    \label{eq:demb}
  \vctr{z}^t_\vertex{v}[i]=\begin{cases}
    \vctr{a}_\vertex{v}[i] \phi(\vctr{w}_\vertex{v}[i] t + \vctr{b}_\vertex{v}[i]), & \text{if~~$1 \leq i\leq \gamma d$}. \\
    \vctr{a}_\vertex{v}[i], & \text{if~~$\gamma d < i \leq d$}.
  \end{cases}
\end{equation}
where $\vctr{a}_\vertex{v}\in\mathbb{R}^d$, $\vctr{w}_\vertex{v}\in\mathbb{R}^{\gamma d}$ and $\vctr{b}_\vertex{v}\in\mathbb{R}^{\gamma d}$ are node-specific parameters, $\gamma$ specifies the percentage of features that are a function of time, and $\phi$ is an activation function. Notice how the embedding vector $\vctr{z}^t_\vertex{v}$ is an explicit function of time. To obtain the embedding of a node $\vertex{v}$ at a specific time such as $1999$, one can simple replace $t$ in the equation with $1999$ and use $\vctr{z}^{1999}_\vertex{v}$ as the features of $\vertex{v}$ on $1999$. While \citet{goel2020diachronic} mainly proposed the above embedding for nodes, it can be also used for relations. Moreover, while they propose this approach for interpolation, it can be potentially adapted to extrapolation.

\citet{xu2019temporal} also define $\vctr{z}_\vertex{v}^t$ to be a direct function of time as follows:
\begin{equation} \label{eq:atisee}
\vctr{z}_\vertex{v}^t = \vctr{z}_\vertex{v} + \alpha_\vertex{v}\vctr{w}_\vertex{v}t + \beta_{\vertex{v}}\function{sin}(2\pi \vctr{\omega}_\vertex{v}t) + \mathcal{N}(0, \vctr{\Sigma}_{\vertex{v}})
\end{equation}
where $\vctr{z}_\vertex{v}$ is a shallow embedding for $\vertex{v}$, $\alpha_\vertex{v}\vctr{w}_\vertex{v}t$ aims at modeling trend, $\beta_{\vertex{v}}\function{sin}(2\pi \vctr{\omega}_\vertex{v}t)$ aims at modeling seasonality, and $\mathcal{N}(0, \vctr{\Sigma}_{\vertex{v}})$ turns the embedding into a Gaussian distribution with mean $\vctr{z}_\vertex{v} + \alpha_\vertex{v}\vctr{w}_\vertex{v}t + \beta_{\vertex{v}}\function{sin}(2\pi \vctr{\omega}_\vertex{v}t)$ and covariance matrix $\vctr{\Sigma}_{\vertex{v}}$.

\citet{dasgupta2018hyte} develop a diachronic encoder by mapping every timestamp $t\in T$ into a vector representation $\vctr{z}_t$ (i.e. a shallow encoder for timestamps) and then projecting entity and relation embedding vectors to the space of $t$ as follows:
\begin{align} \label{eq:hyte}
  \vctr{z}^t_\vertex{v}=\vctr{z}_\vertex{v}-(\transpose{\vctr{z}_t} \vctr{z}_\vertex{v})\vctr{z}_t, \quad \quad
  \vctr{z}^t_\vertex{r}=\vctr{z}_\vertex{r}-(\transpose{\vctr{z}}_t \vctr{z}_\vertex{r})\vctr{z}_t
\end{align}
where $\vctr{z}_\vertex{v}$, $\vctr{z}_\relation{r}$ and $\vctr{z}_t$ are node-specific, relation-specific, and timestamp-specific learnable parameters. Note that unlike the encoders in Equation~\eqref{eq:demb} and \eqref{eq:atisee} which provide embeddings at any time $t$, the above encoder can provide embeddings only for a pre-defined set of timestamps.

Instead of considering shallow embeddings for each timestamp as in \cite{dasgupta2018hyte}, \citet{garcia2018learning} consider shallow embeddings for each character in the timestamps. Then, they map each (relation, timestamp) pair into a vector representation $\vctr{z}^t_\relation{r}$ by feeding a shallow embedding of $\relation{r}$ (i.e. $\vctr{z}_\relation{r}$) and the shallow embeddings of the characters in the timestamp into an LSTM\footnote{In cases where there are different time modifiers (e.g., \emph{OccurredAt}, \emph{Since}, and \emph{Until}), they consider a shallow embedding for the time modifier as well and this representation is also fed into the LSTM.}: 
\begin{equation}\label{eq:ta-distmult}
\vctr{z}^t_\relation{r} = \function{LSTM}(\vctr{z}_\relation{r}, \vctr{z}_{c_1}, \dots, \vctr{z}_{c_k})
\end{equation}
where $c_i$ is the $\Th{i}$ character in $t$.
One benefit of this approach is that it can naturally deal with missing values in dates. Note that while \citet{garcia2018learning} mainly proposed the above approach for relation embeddings, it can be used for node embeddings as well.

Diachronic encoders have been also explored for word embeddings to understand how the meaning (or usage) of words has changed over time (see, e.g., \cite{hamilton2016diachronic,bamler2017dynamic}).

\subsection{Staleness}
Consider an extrapolation problem over a CTDG and an encoder that updates the embedding for each node $\vertex{v}$ whenever a new observation involving $\vertex{v}$ is made (e.g., when a new edge is added between $\vertex{v}$ and some other node). Assume the last time the encoder updated the embedding for $\vertex{v}$ was at time $t_\vertex{v}$ and currently we are at time $t~(>t_\vertex{v})$. 
Depending on how long it has passed since $t_\vertex{v}$ (corresponding to $t-t_\vertex{v}$), the embedding for $\vertex{v}$ may be staled. 

To handle the staleness of representations, \citet{kumar2018learning} propose a method to learn how the representation for a node $\vertex{v}$ evolves when no observation involving $\vertex{v}$ (or involving a node that affects $\vertex{v}$) is made. Let $\embedding{EMB}(\vertex{v})=(\vctr{z}_\vertex{v})$ and let $\Delta t_\vertex{v}=t-t_\vertex{v}$. Following the approach proposed by \cite{beutel2018latent}, \citet{kumar2018learning} first create a vector representation $\vctr{z}_{\Delta t_\vertex{v}}\in\mathbb{R}^{d}$ for $\Delta t_\vertex{v}$ where the $\Th{i}$ element of the vector is computed as follows:
\begin{align}
    \vctr{z}_{\Delta t_\vertex{v}}[i] = \vctr{w}[i]\Delta t_\vertex{v}+\vctr{b}[i]
\end{align} 
where $\vctr{w}$ and $\vctr{b}$ are vectors with learnable parameters. Then they compute a new vector representation $\vctr{z}^t_\vertex{v}$ for $\vertex{v}$ at time $t$ as follows:
\begin{align}
    \vctr{z}^t_\vertex{v} = (\vctr{1} + \vctr{z}_{\Delta t_\vertex{v}}) \odot \vctr{z}_\vertex{v}
\end{align}
where $\vctr{1}\in\mathbb{R}^d$ is a vector of ones. Having computed $\vctr{z}^t_\vertex{v}$, instead of using the (potentially) staled representation $\vctr{z}_\vertex{v}$, \citet{kumar2018learning} use $\vctr{z}^t_\vertex{v}$ to make predictions about \vertex{v} at time $t$. Note that while diachronic encoders (see Section~\ref{sec:diachronic}) have been mainly proposed for interpolation, if used for extrapolation, several of them (e.g., see Equations~\eqref{eq:demb}, \eqref{eq:atisee} and \eqref{eq:ta-distmult}) have the benefit of updating node and relation embeddings even when no observations have been made about the node and relation.

\section{Decoders for Dynamic Graphs} \label{sec:dynamic-decoders}
We divide the decoders for dynamic graphs into two categories: time-predicting decoders and time-conditioned decoders. In what follows, we explain each category and provide a summary of the existing approaches for that category.

\subsection{Time-Predicting Decoders}
\label{sec:time-pred-decoders}
Time-predicting decoders can be used for extrapolation or interpolation. In extrapolation settings, these decoders aim at predicting \emph{when} an event will happen. For instance, they aim at predicting when \emph{Bob} will visit \emph{Montreal}. In interpolation settings, they aim at predicting a missing timestamp. For instance, they aim at predicting when \emph{Frank Lampard} became the head coach of \emph{Chelsea} assuming $(Frank Lampard, HeadCoachOf, Chelsea, ?)$ is in a KG, where $?$ shows that the timestamp is missing. 

\citet{sun2012will} were among the first to study the problem of predicting when a particular type of relation will be formed between two nodes. 
To make such a prediction, first they find all paths between two nodes. These paths are matched with a set of pre-defined path templates and the number of paths matching each template is counted. These counts, which can be roughly considered as node-pair embeddings, are fed into a generalized linear model (GLM) and the score of this model is used to define the parameters of a density function.
\citet{sun2012will} use exponential, Weibull~(\cite{weibull1951statistical}), and geometric distributions for defining the density function. Their exponential distribution for the formation of a relation between two nodes is as follows:
\begin{align}
    \function{f}(t) = \frac{1}{\theta}\function{\exp}(-\frac{t}{\theta})
\end{align}
where $\theta$ is the output of the GLM model. An expectation of $t \sim \function{f}(t)$ can be used to predict when the relation will be formed between the two nodes, as described in Section~\ref{sec:tpp}. 

Recently there has been growing interest towards time predicting decoders (\cite{trivedi2019dyrep,trivedi2017know, zuo2018embedding}).
 \citet{trivedi2017know} consider an encoder that provides an embedding function such that given a dynamic graph until time $t$ gives $\embedding{EMB}(\vertex{v})=(\vctr{z}^t_{\vertex{v}})$ and $\embedding{EMB}(\relation{r})=(\mtrx{P}_\relation{r})$. 
They first compute a score for the formation of a relation $\relation{r}$ between two nodes $\vertex{v}$ and $\vertex{u}$ as follows:
\begin{align}
    \function{s}_{\vertex{v},\relation{r},\vertex{u}}(t) = \vctr{z}^{t'}_\vertex{v} \mtrx{P}_\relation{r} \vctr{z}^t_\vertex{u}
\end{align}
The obtained score is then used to modulate the conditional intensity function ($\lambda_{\vertex{v},\relation{r},\vertex{u}}(t|\history{H}_{t-})$) of a TPP for a given relation $\relation{r}$ and entities $\vertex{v}$ and $\vertex{u}$ as follows:
\begin{align} \label{eq:know-evolve-tpp}
    \lambda_{\vertex{v},\relation{r},\vertex{u}}(t|\history{H}_{t-}) =  \function{exp}(\function{s}_{\vertex{v},\relation{r},\vertex{u}}(t)) (t-\Bar{t})
\end{align}
where $\history{H}_{t-}$ represents the history until time $t$ but not including $t$, $\Bar{t}$ represents the most recent time when either $\vertex{v}$ or $\vertex{u}$ was involved in an observation and $t>\Bar{t}$. Using $\function{exp}$ ensures that the intensity function is always positive. To predict when relation $\relation{r}$ will form between $\vertex{v}$ and $\vertex{u}$, one can convert the conditional intensity function into a conditional density function ($\function{f}_{\vertex{v},\relation{r},\vertex{u}}$) and subsequently take an expectation over the time horizon as described in Section~\ref{sec:tpp}. Since the intensity function in Equation~\eqref{eq:know-evolve-tpp} is piece-wise linear, computing the integral corresponding to the survival function can be done efficiently. During training, however, one needs to compute such an integral $|\vertices{V}|^2|\relations{R}|$ times making the computations intractable. To remedy the intractability, \citet{trivedi2017know} propose an approximation algorithm. 

Note that as identified by \citet{jin2019recurrent}, the intensity function in Equation~\eqref{eq:know-evolve-tpp} may not be suitable for settings involving concurrent events. That is because, in the presence of concurrent events, $t$ may be equal to $\Bar{t}$ making the intensity function equivalent to zero.

\citet{trivedi2019dyrep} argue that different types of relations evolve at different rates; e.g., liking posts in a social network occurs more frequently than becoming friends. They model the dynamics of the graph by considering two types of relations: 1- \emph{communications} corresponding to node interactions (e.g., liking someone's post in social media), 2- \emph{associations} corresponding to topological evolutions (e.g., forming a new friendship). They propose to use different TPPs for these two types of relations. 
Towards this goal, they assume the embedding function provided by the encoder gives $\embedding{EMB}(\vertex{v})=(\vctr{z}^t_\vertex{v})$ and $\embedding{EMB}(\relation{r})=(\psi_\relation{r}, \vctr{z}^t_\vertex{r})$ and define the intensity function of their TPP as follows: 
\begin{align}
    \lambda_{\vertex{v},\relation{r},\vertex{u}}(t|\history{H}_{t-}) = \psi_{\relation{r}}\function{log}(1+\function{exp}(\frac{\vctr{z}^{t'}_\relation{r}[\vctr{z}^t_\vertex{v};\vctr{z}^t_\vertex{u}]}{\psi_\relation{r}}))
\end{align}
where $[\vctr{z}^t_\vertex{v};\vctr{z}^t_\vertex{u}]$ is the concatenation of $\vctr{z}^t_\vertex{v}$ and $\vctr{z}^t_\vertex{u}$. 
Notice that the above intensity function does not have the $(t-\Bar{t})$ term used in Equation~\eqref{eq:know-evolve-tpp}. Instead, different rates of evolution ($\psi_{\relation{r}}$) for relations of different types are introduced. Similar to \citet{trivedi2017know}, during training \citet{trivedi2019dyrep} also need to compute the integral corresponding to the survival function $|\vertices{V}|^2|\relations{R}|$ times; they approximate the computations through sampling. 

\citet{zuo2018embedding} use the intensity function of a Hawkes process~(\cite{hawkes1971spectra,mei2017neural}). The intensity of the interaction is obtained by the Euclidean distance between the interacting nodes and an exponentially discounted interaction history of the neighbors.

For interpolation in KGs, to predict the timestamp for a triple $(\vertex{v}, \relation{r}, \vertex{u}, ?)$, \citet{leblay2018deriving} and \citet{dasgupta2018hyte} replace the missing timestamp with all timestamps observed in the KG, find the score for all produced temporal triples using a time-conditioned decoder (see Section~\ref{sec:time-conditioned-decoders}), and select the timestamp resulting in the highest score. Note that this approach may not scale to KGs with many timestamps.

\subsection{Time-Conditioned Decoders} \label{sec:time-conditioned-decoders}
Time-conditioned decoders are decoders whose goal is to make predictions for specific timestamps given as input (as opposed to predicting \emph{when} something will happen). These decoders can be used for extrapolation (e.g., predict who will be the CEO of Apple two years from now) or interpolation (e.g., predict who was the CEO of Apple on \emph{2006-04-01}, assuming this piece of information is not explicitly stored in the KG). In other words, time-conditioned decoders predict what happened (or will happen) at some time $t$ where $t$ can be different in different queries. Time-conditioned decoders are mainly similar to static decoders. In what follows, we describe some works that aim at making predictions for a specific timestamp and the decoders they employ.

\citet{dasgupta2018hyte} develop a model for interpolation in KGs. They use the encoder in Equation~\eqref{eq:hyte} to produce node and relation embeddings that are functions of time. To make a prediction about whether some edge $(\vertex{v}, \relation{r}, \vertex{u})$ existed at time $t$ or not, they use TransE (see Section~\ref{sec:static-kg-decoders} for details) as the decoder:
\begin{align}
    || \vctr{z}^t_\vertex{v} + \vctr{z}^t_\vertex{r} - \vctr{z}^t_\vertex{u} ||
\end{align}
\citet{leblay2018deriving} develop a model for interpolation where a shallow encoder provides node, relation, and timestamp embeddings and the decoder is defined as follows:
\begin{align}
    || \vctr{z}_\vertex{v} + \vctr{z}_\vertex{r} + \vctr{z}_t - \vctr{z}_\vertex{u} ||
\end{align}
The above decoder can be viewed as an extension of TransE by adding the embedding of the timestamp to the score function. \citet{leblay2018deriving} provide other extensions of TransE as well.
\citet{ma2018embedding} develop several models for interpolation by using shallow encoders that provide node, relation, and timestamp embeddings and they extend DistMult, ComplEx, RESCAL, and several other decoders by incorporating the timestamp embedding into their score functions. Note that there exists a close connection between these models and the tensor decomposition models (with order 4 tensors) discussed in Section~\ref{sec:temporal-decomposition}. 

If a shallow encoder is used for the timestamp embeddings (which is the case in the works described so far), then an embedding can only be learned for the timestamps that have been observed in the train set. Therefore, these approaches may not generalize well to the timestamps not observed in the train set as a vector representation has not been learned for these timestamps. For instance, if a KG does not contain any information with timestamp $08/07/2012$, the above approaches do not learn an embedding for this date and may not be able to make predictions about that date.  For the same reason, these models cannot be used effectively for predicting something in a future timestamp (i.e. extrapolation) as the training data does not contain any future timestamps. Moreover, these models require many parameters and are prone to overfitting when the number of different timestamps in the training data is large.

The model proposed by \citet{garcia2018learning} addresses the above issues. They use the encoder in Equation~\eqref{eq:ta-distmult} to generate an embedding $\vctr{z}^t_\relation{r}$  for each relation $\relation{r}$ at time $t$; for nodes, they use a shallow encoder. Then, having the vector representations $\vctr{z}_\relation{v}$, $\vctr{z}^t_\relation{r}$, and $\vctr{z}_\relation{u}$, they use TransE and DistMult as the decoder. Since they use shallow embeddings for each character in the timestamp (not for the timestamp itself), their model can be potentially applied to timestamps unseen during training.

The approaches described so far learn a single static representation ($\vctr{z}_{\vertex{v}}$) for nodes and use this representation to make predictions about the nodes at any time. \citet{goel2020diachronic} argue that learning a static representation for nodes may result in the loss of important information. That is because to make a prediction about a node $\vertex{v}$ in some timestamp $t$ (e.g., predicting the movies $\vertex{v}$ liked on $1990$ assuming $\vertex{v}$ is a person), one needs to know the specific properties of $\vertex{v}$ around time $t$, whereas a static representation only provides an aggregation of $\vertex{v}$'s properties over time. To address this issue, \citet{goel2020diachronic} use the encoder in Equation~\eqref{eq:demb} to learns node features at any specific time. For the decoder, they use TransE, DistMult, and SimplE. They prove that using sine as the activation function for Equation~\eqref{eq:demb} and SimplE as the decoder results in a model that is fully expressive for link prediction for temporal KGs.

\paragraph{Making Predictions for a single timestamp:} 
In cases where all predictions are to be made for a single timestamp or a single time interval, (e.g.,  predicting what happens in the next snapshot of a DTDG, or predicting what happens in near future \emph{without} predicting when it will happen), the existing approaches mostly use a static decoder from Section~\ref{sec:static-decoders}. 
A notable exception is the work of \citet{zhou2018dynamic} for link prediction in DTDGs where a point process based on triadic closure is employed. Let $\vertex{v}_i$, $\vertex{v}_j$, and $\vertex{v}_k$ be three nodes in a graph at snapshot $t$. $\vertex{v}_i$, $\vertex{v}_j$, and $\vertex{v}_k$ form a \emph{closed triad} if all of them are pair-wise connected, and they form an \emph{open triad} if all but one pair of the nodes are connected. In an open triad, the node that is connected to the other two nodes is called the \emph{center node} of the triad. A fundamental mechanism in
the formation and evolution of dynamic networks known as \emph{triad closure} is the process of closed triads being created from open triads~(\cite{coleman1994foundations,huang2015triadic}).

\begin{example}
Consider the graph in Figure~\ref{graphs-fig}(b). In this graph, $\vertex{v}_1$, $\vertex{v}_2$ and $\vertex{v}_3$ form a closed triad. $\vertex{v}_2$, $\vertex{v}_3$ and $\vertex{v}_4$ form an open triad with $\vertex{v}_2$ being the center node of this open triad.
\end{example}

For two nodes $\vertex{v}$ and $\vertex{u}$, let $\vctr{z}^t_\vertex{v}$ and $\vctr{z}^t_\vertex{u}$ represent the embedding of the two nodes at the $\Th{t}$ snapshot respectively. \citet{zhou2018dynamic} model the probability of $\vertex{v}$ and $\vertex{u}$ forming an edge in the next snapshot to be proportional to the number of open triads this edge will close and the similarities of $\vctr{z}^t_\vertex{v}$ and $\vctr{z}^t_\vertex{u}$ to the embeddings of the center nodes in the open triads involving $\vertex{v}$ and $\vertex{u}$.

\section{Other Relevant Models and Problems}
\label{sec:other-models}
While we mainly provided an overview of the models conforming to an encoder-decoder framework, there are other active lines of work on modeling dynamic (knowledge) graphs. Here, we briefly review some of the other related works and some similar problems.

\subsection{Statistical Relational Models} 
Statistical relational models~(\cite{raedt2016statistical,koller2007introduction}) aim at marrying logic and probability to build probabilistic relational models. Examples include Markov logic networks (\cite{richardson2006markov}), Problog (\cite{de2007problog}), probabilistic soft logic (\cite{kimmig2012short}) and relational logistic regression (\cite{kazemi2014relational}). These models typically use soft rules such as:
\begin{align}
    <w: \function{Friends}(x, y) \wedge \function{Friends}(y, z)\Rightarrow \function{Friends}(x, z)>
\end{align}
where the rule implies ``friends of friends are likely to be friends'' and the weight $w$ of the rule is a measure of confidence for the rule. A model is created using a combination of such soft rules and predictions are made using logical and probabilistic inference. Different models differ in how they interpret these rules and weights. The rules and the weights may be learned from data. One may consider several hidden variables for each node, where each hidden variable corresponds to a feature, and define or learn the correlation between hidden and observed variables using logical rules. The hidden variables in this setting resemble the node representations learned by encoder-decoder architectures with the main difference being that these models learn a set of random variables per node, whereas encoder-decoder architectures learn vectors and matrices of numbers.

In comparison to the encoder-decoder framework, statistical relational models naturally capture uncertainty about facts and relations, which is critical in applications where relations are derived from noisy measurements or ambiguous interpretations such as natural language processing. Furthermore, statistical relational models permit joint inference in a principled and interpretable way over the entire graph while taking into account the uncertainty of the facts.  However, this comes at a computational price and therefore it is often needed to restrict the expressivity of the model by only using model structures that are known to be tractable~(\cite{VHaarenMLJ15,broeck2011completeness,kazemi2016new}) or to approximate the inference~(\cite{van2013complexity,kersting2009counting,bui2013automorphism}). In contrast, inference with encoder-decoder models usually scales to large datasets and although the operators used for inference are in some cases questionable, end-to-end learning allows the weights of the operators to be adjusted to yield the best predictions possible for desired tasks. 

Statistical relational models have been extended to dynamic cases as well (see, e.g., \cite{gabbay1998handbook,sadilek2010recognizing,papai2012slice,dylla2013temporal,huber2014applying,chekol2017marrying,chekol2018rule}). In their simplest form, these models can be extended to dynamic cases by adding time as an argument similar to the following soft rule:
\begin{align}
    <w: \function{Friends}(x, y, t) \wedge \function{Friends}(y, z, t) \wedge \neg\function{Friends}(x, z, t)\Rightarrow \function{Friends}(x, z, t+1)>
\end{align}
where the rule may increase the probability of the worlds where friends of friends become friends in the next snapshot. The amount of increase in the probability is controlled by $w$. Note how this rule is similar to the triadic closure procedure of \citet{huang2015triadic} as it models how closed triads are created from open triads.

Besides the approaches based on soft rules, there exists a family of approaches based on walks where the aim is to learn probabilistic walks on the graphs (or KGs), which, starting from any node $\vertex{v}$, ends up (probabilistically) at the nodes that have a desired relation with $\vertex{v}$. These probabilistic walks are different from the random walk approaches discussed in this survey. Examples of such approaches include the works of \cite{lao2010relational,lao2011random,das2018go}. These approaches are quite similar to the approaches based on soft rules corresponding in some cases to a subset of the soft rule approaches~(\cite{kazemi2018bridging}).

\subsection{Spatiotemporal Graphs}
For applications where it is possible to define temporal relations between the nodes, several papers take a DTDG and combine the snapshots through connecting the nodes in different snapshots to get a spatiotemporal graph (ST graph), i.e., a graph that spans both space and time (links across different time steps are known as temporal links whereas links within a time step are known as spatial links). Then, instead of learning on an evolving graph, a model is learned on the resulting (static) ST graph.
An ST graph can be considered as a KG with spatial and temporal relations.

Structured time series problems such as video activity recognition and segmentation, and traffic flow prediction are examples of applications that have benefited from a graph-theoretic formulation by creating ST graphs. In the video domain, for instance, a graph is extracted from each video frame and combined with the graphs extracted from other frames. \citet{pandhre2018stwalk} utilize random walk encoders for ST graphs in two ways: 1- creating random walks on the ST graph (which snaps both space and time), 2- first creating spatial random walks at the current snapshot and then temporal random walks over the temporal graph obtained by only keeping the edges between nodes in different snapshots. Their two models show superior results compared to several baselines on temporal prediction problems such as trajectory classification. 

For activity recognition, \citet{brendel2011learning} learn a structured activity model from the ST graphs obtained from videos. For recognition, they match the ST graph of the given video with the per-class learned activity models.  \citet{jain2016structural} use RNNs in a structured setting by modeling each node and edge in an ST graph using an RNN. To make the learning feasible, they partition the nodes (and edges) using semantic similarities and share the RNNs amongst the nodes (and edges) within the same partition. \citet{wang2018videos} pose activity recognition as a graph classification problem. They extract two kinds of graphs from a video, namely the similarity and ST graph. The similarity graph learns connections between objects that are semantically related to each other whereas the ST graph learns connections between objects that overlap in space and time. Then, they utilize graph classification models on the constructed graphs for activity classification. Action recognition has been also modeled as reasoning over a dynamic graph without creating an ST graph. \citet{li2018spatio} and \citet{ghosh2018stacked} use skeleton-based datasets for action recognition. They consider an evolving human skeleton during the course of the action as a DTDG, contrary to the above approaches that use heuristics to combine the individual frame-level graphs. Then they perform graph convolutions in both the temporal as well as the spatial domain. While \citet{li2018spatio} only use human pose features, \citet{ghosh2018stacked} also leverage additional contextual cues such as object features, functional relationships, etc.

Another challenging problem formulated as an ST graph is that of traffic flow prediction. \citet{yu2018spatio} propose a spatiotemporal convolution block (ST ConvBlock) that consists of temporal gated convolutions and spatial gated convolutions. These ST ConvBlocks are stacked to obtain feature representations for each node in the graph and for the traffic speed prediction. The traffic flow prediction problem has been also modeled as reasoning over the dynamic graph using RNN-based approaches (as discussed in Section~\ref{sec:rnn-dtdg}) without creating an ST graph (see, e.g., \cite{li2017diffusion,yu20193d}). 

\subsection{Constructing (Dynamic) KGs from Text}
For the case of the KGs, while we focused in this survey on methods for using a (dynamic) KG to make predictions about its past, current, or future state, there is a large body of research on how to construct a (dynamic) KG from text~(\cite{carlson2010toward,dong2014knowledge,zhang2015deepdive,das2018building}). These approaches typically rely on information extractors to obtain new (probabilistic) facts from text and then add to the KG the new facts that pass a confidence threshold. Besides using the probabilities produced by the information extractors, some works (e.g., \cite{dong2014knowledge}) also leverage the predictive models discussed in this survey to obtain a prior probability for the new facts solely based on what is already in the graph (and not based on textual data). A detailed discussion of these approaches is out of the scope of this paper.
    
\section{Applications, Datasets \& Codes}
\label{sec:datasets}
In this section, we provide an overview of the main applications of (dynamic) graphs. We describe some of the datasets that are widely used in the community for representation learning for (knowledge) graphs. We also provide links to online code for some of the works we discussed in the paper.

\subsection{Applications}
\label{subsec: application}

\textbf{Dynamic Link prediction:}
A natural problem for a graph is to predict if there is a link (with label $r$ in case of a KG) between two nodes $\vertex{v}$ and $\vertex{u}$. In the dynamic case, one may be interested in predicting if such a link existed at time $t$ in the past or if it will appear sometime in the future. An example of this would be to predict whether Donald Trump will \textbf{visit} China in the next year or not and we query for this link in a KG. Example applications include temporal KG completion, friend recommendation, finding biological connections among species, and predicting obsolete facts in a KG. 
The most common evaluation metrics for this task include: AOC (area under ROC curve), corresponding to the probability that the predictor gives a higher score to a randomly chosen existing link than a randomly chosen nonexistent one, GMAUC, a geometric mean of AOC, and PRAUC (area under precision-recall curve) - see, for example, \cite{chen2019lstm}.  Another metric named error rate was considered in \citet{chen2019lstm} corresponding to the ratio of the number of mispredicted links to the total number of truly existing links. 

\noindent \textbf{Dynamic Entity/relation prediction:}
One of the fundamental problems in KGs is to predict missing entities or relations. It is a classical problem where we want to predict either a missing head entity $(?,\relation{r},\vertex{u})$, a missing tail entity $(\vertex{v},\relation{r},?)$, or a missing relation $(\vertex{v},?,\vertex{u})$. In the context of a DTDG, one may want to predict the missing entity or relation in the next snapshot. In the case of a CTDG, one may want to predict the missing entity or relation at a specific timestamp $t$ (e.g., $(?,\relation{r},\vertex{v},t)$). An example mentioned by \citet{trivedi2017know} is to predict who Donald Trump will mention next?
\citet{leblay2018deriving} and \citet{dasgupta2018hyte} considered the task of predicting a missing entity in the temporal case. They rank all entities that can potentially be the missing entity and then find the rank of the actual missing entity. They used mean rank (MR), and the percentage of cases where the actual missing entity is ranked among the top K (known as Hit@K) to compare the quality of the results. In addition to the above metrics, \citet{garcia2018learning} and \citet{goel2020diachronic} also computed mean reciprocal rank (MRR). MRR is generally reported under two settings: raw and filtered (see \cite{bordes2013translating} for the details).

\noindent \textbf{Recommender systems:}
 The design of dynamic recommender systems is an important applied dynamic graph problem faced by a myriad of e-commerce companies in online retail, video streaming, and music streaming, to name a few examples. In a dynamic recommender-system problem, we have a set of users, a set of items, and a set of timestamped interactions between users and items, and we seek to recommend items to users based on their current tastes (\cite{wu2017recurrent, kumar2018learning}). The key is that tastes may exhibit cyclic or trend behavior which should be adapted to or anticipated inasmuch as this is possible. As a coarse approximation, we can view a dynamic recommender-system problem as one of dynamic link prediction (or dynamic entity prediction) and attempt to recommend to users the items they are likely to autonomously choose.  However, a more fine-grained analysis reveals several complications present in recommender systems that may be absent in other dynamic link prediction problems. First, the actual output of a recommender system for a given user is a sequence of slates of recommended items.  If the items on an output slate are highly similar, then their utility to the user may be strongly correlated, so that there is a non-negligible risk of the slate being useless to the user. This risk is mitigated with a more diverse slate, even if the diverse slate has a lower sum of expected utilities than the uniform slate (see \cite{kaminskas2017diversity})\footnote{This reasoning resembles  Markowitz's portfolio theory (\cite{markowitz1952portfolio}).}.  Second, from the point of view of an e-commerce company, the purpose of a recommender system is typically to maximize profit in the long term. Therefore, it may be desirable to recommend to the user items in which the user has no immediate interest, but which are expected to cause the user to purchase profitable items or click on profitable advertisements.  From this perspective, the design of a dynamic recommender system can be seen as a reinforcement learning problem on a dynamic graph.  One seeks to dynamically control an evolving graph to maximize some profit-related objective.

\noindent \textbf{Time Prediction:}
For dynamic graphs, besides predicting which event will happen in the future, an interesting problem is to also predict when that event will happen. As compared to other tasks, this task only exists for dynamic networks. For instance, we saw the example in Section~\ref{sec:time-pred-decoders} of predicting when \emph{Bob} will visit \emph{Montreal}. A similar time prediction problem is the temporal scoping problem in KG completion where the goal is to predict missing timestamps (e.g., answering queries such as $(\vertex{v}, \relation{r}, \vertex{u}, ?)$, where $\vertex{v}$ is known to have had a relation $\relation{r}$ with $\vertex{u}$ in the past).
\citet{sun2012will} and \citet{trivedi2017know,trivedi2019dyrep} used the mean absolute error between the predicted time and ground truth to measure the quality of the results. \citet{dasgupta2018hyte} ordered the predicted timestamps in the decreasing order of their probabilities and selected the rank associated with the correct timestamp. They computed the mean rank (MR) to compare the results.  

\noindent \textbf{Node classification:}
Node classification is the problem of classifying graph nodes into different classes. An example of a node classification problem is to predict the political affiliation of the users of a social network based on their attributes, connections, and activities. In particular, one may be interested in making such predictions in a streaming scenario where the classification scores keep updating as new events happen or as a user activity is observed. Node classification is often studied under two settings: transductive and inductive. In the transductive setting (also known as semi-supervised classification), given the labels of a few nodes, we want to predict the labels of the other nodes in the graph.
In the inductive setting, the label is to be predicted for new nodes that have not been seen during training.
The problem of node classification becomes challenging in the dynamic case as the distribution of the class labels may change over time. There are not many publicly-available datasets for dynamic node classification. \citet{pareja2019evolvegcn} used Elliptic a network of bitcoin transactions, for temporal node classification. \citet{sato2019dyane} considered a dataset of face to face proximity for temporal node classification. They used the SIR model for node classification, which is a popular framework to model the spread of epidemics~(\cite{hethcote2000mathematics}). At each timestamp, each node can be one of three possible states: susceptible (S), infectious (I),
and recovered (R). Classification accuracy is a widely used metric for this task. For datasets where the task is a multi-class classification, micro-F1 and macro-F1 scores are also used to measure the performance~(\cite{cui2018survey}).

\noindent \textbf{Graph classification:}
Graph classification is the problem of classifying the whole graph into one class from a set of predefined classes. This task can be useful in domains like bioinformatics and social networks. In bioinformatics, for instance, one important application is the protein function classification where proteins are viewed as graphs. As another example, \citet{taheri2019Learning} modeled activity state classification for a troop of GPS-tracked baboons as a dynamic graph classification problem where the labels are different activity states such as sleeping, hanging-out, coordinated non-progression, and coordinated progression. Common graph classification benchmarks include COLLAB~(\cite{yanardag2015structural}), PROTEINS~(\cite{borgwardt2005protein}) and D\&D~(\cite{dobson2003distinguishing}). The performance is typically measured in terms of classification accuracy.

\noindent \textbf{Network clustering:}
Network clustering or detecting communities in graphs such as social networks, biological networks, etc. is an important problem. A network cluster/community typically refers to a subset of nodes that are densely connected, but loosely connected to the rest of the nodes. One challenge in community detection for dynamic graphs is that one needs to model how communities evolve.
A common performance measure used for this task is the overlap between the predicted and the true cluster assignments.
\citet{xin2016adaptive} utilize the random walk encoders to find the closely associated nodes for a given node and cluster the global network into overlapping communities. Furthermore, the closely associated nodes are updated when impacted by dynamic events, giving a dynamic community detection method. \citet{bruna2019supervised} recast community detection as a node-wise classification problem and present a family of graph neural networks for solving the community detection problem in a supervised setting. \citet{crawford2018cluenet} considered a dynamic communication network between 75 patients and health workers in France with four different clusters representing doctors, administrators, nurses, and patients.

\noindent \textbf{Dynamic Question/query answering:} 
The way search engines respond to our questions has evolved in the last few years. While traditionally search engines were aiming at suggesting documents in which the answer to a query question can be potentially found, these days they try to directly answer the question.  This has become possible in part due to question answering over KGs (QA-KG). \citet{jia2018tequila} further extend this work to temporal questions. They defined a temporal question as a question that has a temporal expression (date, time, interval, and periodic events) or temporal signal (before, after, during, etc.) in the question or whose answer is temporal.
 For QA-KG, the most popular metrics are precision, recall, F1 score, AUC, and accuracy.  An example of a temporal question given by \citet{jia2018tequila} was, ``Which teams did Neymar play for before joining PSG?" 
\citet{hamilton2018embedding} propose a way of mapping a query formulated as a conjunctive first-order logic formula into a vector representation (corresponding to a query embedding) through geometric operations. The geometric operations leading to query embeddings are jointly optimized with node embeddings such that the query embedding is close in the embedded space to the embeddings of the entities corresponding to the correct answer of the query.  For QA-KG, the most popular metrics are precision, recall, F1 score, AUC, and accuracy. 

\subsection{Datasets}
 There is a large collection of datasets used for research on static and dynamic graphs in the community. For brevity, we will only survey a representative sample of these datasets. To explore more network datasets, we refer readers to several popular network repositories such as Stanford Large Network Dataset Collection (\url{https://snap.stanford.edu/data/index.html}), Network Repository (\url{http://networkrepository.com/index.php}), Social Computing data repository (\url{http://socialcomputing.asu.edu/pages/datasets}), LINQS (\url{https://linqs.soe.ucsc.edu/data}), UCI Network Data Repository (\url{https://networkdata.ics.uci.edu/}), CNetS Data Repository (\url{http://cnets.indiana.edu/resources/data-repository/}) and Koblenz Network Collection (\url{http://konect.uni-koblenz.de/networks/}). Section~7 of \cite{cui2018survey} and Section~7.1 of \cite{zhang2018network} are also good starting points to explore other datasets. 
 
Table~\ref{table:datasets} gives a brief summary of the datasets. In Section~\ref{subsec:temporal}, we describe some evolving and temporal datasets. In Section~\ref{subsec:non-temporal}, we survey some of the popular traditional datasets that have been widely used in the community but are not necessarily temporal or dynamic.
 
\begin{table}[t]
\setlength{\tabcolsep}{2pt}
 \scriptsize
         \centering
        \begin{tabular}{ p{23mm} | l | l | p{40mm} | l }
             Dataset & Type & Nodes & Edges & Granularity \Bstrut\\
             \hline
             Social\newline Evolution & Social network & 83 & 376-791~Associations \newline 2,016,339~Communications & 6 mins \Tstrut\\
             \hline
             Github & Social network & 12,328 & 70,640-166,565~Associations \newline 604,649~Communications & - \Tstrut\\
             \hline
             HEP-TH & Citation network & 1,424-7,980 & 2,556-21,036 & Monthly \Tstrut\\
             \hline
             Autonomous\newline systems & Communication network & 103-6,474\footnote{\label{footnote:snap}\url{http://snap.stanford.edu/data/\#as}} & 243-13,233 & Daily \Tstrut\\
             \hline
             GDELT & Events knowledge graph & 14,018 & 31.29M & 15 mins \Tstrut\\
             \hline
             ICEWS & Events knowledge graph & 12498 & 0.67M & Daily \Tstrut\\
             \hline
             YAGO & Knowledge graph & 15,403 & 138,056 & Mostly yearly \Tstrut\\
             \hline
             Wikidata & Knowledge graph & 11,134 & 150,079 & Yearly \Tstrut\\
             \hline
             Reddit & Social network & 55,863 & 858,49 & Seconds \Tstrut\\ 
             \hline
             Enron & Email network & 151 & 50.5K & Seconds \Tstrut\\
             \hline
             FB-Forum & Social network & 899 & 33.7K & Seconds \Tstrut\\
             \hline
             
             Blog & Social network & 5,196 & 171,743 & - \Tstrut\\ \hline
             Cora\footnote{\url{https://linqs.soe.ucsc.edu/node/236}} & Citation network & 2708 & 5429 & - \Tstrut\\ \hline
             Flicker & Social network & 1,715,256 & 22,613,981 & - \Tstrut\\ \hline
             UCI & Communication network & 1,899 & 59,835 & Seconds \Tstrut\\
             \hline
             Radoslaw\footnote{\url{http://networkrepository.com/radoslaw-email.php}} & Email network & 167 & 82.9K & Seconds \Tstrut\\
             \hline
             DBLP\footnote{\url{https://dblp.uni-trier.de/xml}}\footnote{\url{https://snap.stanford.edu/data/com-DBLP.html}} & Citation network & 315,159 & 743,70 & -\Tstrut\\
             \hline
             YELP & Bipartite ratings & 6,569  & 95,361 & Seconds \Tstrut\\
             \hline
             MovieLens-10M & Bipartite ratings & 20,537  & 43,760 & Seconds \Tstrut\\
             \hline
             CONTACT & Face-to-face proximity & 274 & 28,200 & Seconds \Tstrut\\
             \hline
             HYPERTEXT09 & Face-to-face proximity & 113 & 20,800 & Seconds \Tstrut \\
             \hline
             Elliptic\footnote{\url{https://www.kaggle.com/ellipticco/elliptic-data-set}} & Bitcoin transactions & 203,769 & 234,355 & 49 time steps
             \Bstrut\\
           
        \end{tabular}
         \caption{A summary of the datasets used in dynamic (knowledge) graph publications, the type of data they contain, the number of nodes and edges they contain, and their temporal granularity.}
         \label{table:datasets}
     \end{table}

\subsubsection{Temporal Datasets}
\label{subsec:temporal}
    
\textbf{Social Evolution Dataset\footnote{\url{http://realitycommons.media.mit.edu/socialevolution.html}}:} The social evolution dataset was released by the MIT Human Dynamics Lab~(\cite{madan2012sensing}) and is used by \citet{trivedi2019dyrep}. The dataset is collected between Jan 2008 to Sep 2008 and has 83 nodes. As mentioned in Section~\ref{sec:time-pred-decoders}, \citet{trivedi2019dyrep} consider two categories of relations: associations (or topological evolution) and communications (or interactions). The number of associations evolves from 376 initial edges to 791 final edges. The number of communication events (proximity, calls, and SMS) in the dataset is 2,016,339. 
    
\textbf{Github Dataset:} 
Github is a web-based hosting service for codes. 
\citet{trivedi2019dyrep} collected a dataset from Github archives between Jan 2013 and Dec 2013. They consider ``following a user'' as an associative event and ``starring'' or ``watching'' a repository as a communicative event. The dataset has 12,328 nodes. The number of associations evolves from 70,640 initial edges to 166,565 final edges. There are 604,649 communication events between the users in this dataset.  
    
\textbf{HEP-TH:} 
\citet{gehrke2003overview} created a dataset of arXiv papers in the High Energy Physics Theory conference from January 1993 to April 2003. The graph is a citation network where nodes represent papers and directed edges represent the citations. \citet{goyal2017dyngem} made this an evolving graph by considering all published papers up to that month. Their graph evolves from 1424 nodes to 7980 nodes, and from 2,556 edges to 21,036 edges.  \citet{goyal2018dyngraph2vec} and \citet{yu2018netwalk} also conducted experiments on variants of this dataset. 
    
\textbf{Autonomous Systems\footnote{\url{http://snap.stanford.edu/data/as-733.html}}:} Autonomous systems graph~(\cite{leskovec2005graphs}) is a communication network from the Border Gateway Protocol logs. The graph has 733 daily snapshots from Nov 1997 to Jan 2000. The graph grows from 103 to 6,474 nodes and from 243 to 13,233 edges. One unique thing about this dataset is that while most other graphs have only the addition of the nodes and edges, this graph has instances of both addition and deletion of nodes and edges. \citet{goyal2017dyngem} use the first 100 snapshots, and \citet{goyal2018dyngraph2vec} use the last 50 snapshots for their experiments. 
    
\textbf{GDELT:} Global Database of Events, Language, and Tone (GDELT)~(\cite{leetaru2013gdelt}) is an initiative to construct a database of all the events across the globe connecting people, organizations, events, news sources, and locations. \citet{trivedi2017know} collected a subset of this data from April 1, 2015, to Mar 31, 2016, with the temporal granularity of 15 mins. This subset contains 14,018 nodes, 20 types of relations, and 31.29M edges. \citet{goel2020diachronic} use this dataset with modified train, validation, and test sets for temporal KG completion.
    
\textbf{ICEWS\footnote{\url{http://www.icews.com/}}:}  Integrated Crisis Early Warning System (ICEWS) dataset~(\cite{boschee2015icews}) contains information about political events with their timestamps. Here the entities represent important political people (presidents, prime ministers, bureaucrats) and countries, and the relations are political scenarios such as negotiate, sign a formal agreement, criticize, etc. The data\footnote{\url{https://github.com/rstriv/Know-Evolve/tree/master/data/icews}} used by (\cite{trivedi2017know}) is collected from Jan 1, 2014 to Dec 31, 2014 with the temporal granularity of 24 hours and has 12,498 nodes, 260 types of relations, and 0.67M facts. \citet{garcia2018learning} created two KGs based on the ICEWS dataset. One of these KGs\footnote{\url{https://github.com/nle-ml/mmkb/tree/master/TemporalKGs/icews14}} contains information from 2014 and has 6,869 nodes, 230 types of relations, and 96,730 facts. The other one is a longer-term KG\footnote{\url{https://github.com/nle-ml/mmkb/tree/master/TemporalKGs/icews05-15}} containing the events occurring between 2005-2015 with 10,094 nodes, 251 types of relations, and 461,329 facts.

\textbf{YAGO:} YAGO\footnote{\url{https://www.mpi-inf.mpg.de/departments/databases-and-information-systems/research/yago-naga/yago/}}~(\cite{hoffart2013yago2}) is a spatially and temporally enriched version of the Wikipedia knowledge base. The nodes represent people, groups, artifacts and events while the relations represent facts such as \relation{wasBornIn}, \relation{playsFor}, \relation{isLocatedIn}, etc. YAGO contains temporal information in the form of ``occursSince'' and ``occursUntil''. The dataset\footnote{\url{https://github.com/nle-ml/mmkb/tree/master/TemporalKGs/yago15k}} created by \citet{garcia2018learning} has 15,403 nodes, 34 types of relations and 138,056 facts. \citet{jiang2016towards}'s dataset has 9,513 nodes, 10 types of relations, and 15,914 facts. 
     
\textbf{Wikidata:} \citet{leblay2018deriving} created a temporal knowledge base\footnote{\url{https://staff.aist.go.jp/julien.leblay/datasets/}} using Wikidata.  \citet{garcia2018learning} considered a subset of this dataset by selecting the most frequent entities along with the relations that include these entities. In their dataset\footnote{\url{https://github.com/nle-ml/mmkb/tree/master/TemporalKGs/wikidata}}, they have 11,134 nodes, 95 types of relations, and 150,079 facts.
     
\textbf{Reddit HyperLink Network\footnote{\url{http://snap.stanford.edu/data/soc-RedditHyperlinks.html}}:} Subreddit hyperlink network~(\cite{kumar2018community}) is a directed network extracted from the posts that create hyperlinks from one subreddit to another. Each edge has temporal information, the sentiment of the source towards the target, and the text of the source post. The dataset also comes with subreddit embeddings for 51,278 subreddits. There are 55,863 nodes and 858,490 edges in the graph. 
     
\textbf{Enron\footnote{\url{https://en.wikipedia.org/wiki/Enron_Corpus}}:} Enron email dataset~(\cite{klimt2004introducing}) is the network of email exchanges among the employees of Enron. This data was originally released by the Federal Energy Regulatory Commission as part of their investigation. There are several variants\footnote{\url{https://snap.stanford.edu/data/email-Enron.html},~\url{https://www.cs.cmu.edu/~enron/},~\url{http://networkrepository.com/},~\url{https://www.kaggle.com/wcukierski/enron-email-dataset}} of this dataset available and it has been widely used in the community~(\cite{nguyen2018continuous,de2018combining,chen2018gc,sankar2018dynamic}). 
     
\textbf{FB-FORUM\footnote{\url{http://networkrepository.com/fb-forum.php}}:} This dataset comes from a Facebook-like online community of students at the University of California, Irvine and was collected in 2004~(\cite{opsahl2011triadic,nr}). This is a bipartite graph where the nodes represent students and groups while the edges represent students' broadcast messages on the groups. The dataset used by \citet{nguyen2018continuous} for their experiments had 899 nodes and 33.7K edges along with timestamp in Unix time.
     
\textbf{UCI\footnote{\url{http://konect.uni-koblenz.de/networks/opsahl-ucsocial}}:} This dataset~(\cite{Kunegis:2013:KKN:2487788.2488173}) is obtained from the same social network as the one for FB-FORUM. The dataset is a communication network among users along with timestamps. This network has 1,899 nodes and 59,835 edges~(\cite{ma2018dynamic,sankar2018dynamic,yu2018netwalk}).

\textbf{YELP\footnote{\url{https://www.yelp.com/dataset}}:} 
The YELP dataset is a subset of YELP's businesses, reviews, and user data. It was originally made public as a Kaggle contest. The dataset consists of 6,685,900 reviews by 1,637,138 users for 192,609 businesses. \citet{sankar2018dynamic} use a part of the YELP dataset where they select the businesses in the state of Arizona and retain businesses that have more than 15 reviews.
    
\textbf{MovieLens-10M\footnote{\url{https://grouplens.org/datasets/movielens/10m/}}:} The MovieLens dataset~(\cite{harper2016movielens}) is a dynamic user-tag interactions dataset. It consists of 10 million ratings and 100,000 tag applications applied to 10,000 movies by 72,000 users. The dataset shows the tagging behavior of users on the movies they rated. \citet{sankar2018dynamic} utilize a subset of this dataset with 20,537 nodes and 43,760 links.
    
\textbf{CONTACT:} The CONTACT dataset introduced by~(\cite{chaintreau2007impact}) is a dynamic network for face-to-face proximity collected through wireless devices carried by people; a link is created between two people whenever they interact. It contains data from 274 people and 28.2K interactions as described in~\cite{chen2018gc}. 
    
\textbf{HYPERTEXT09:} The HYPERTEXT09 dataset described in~\cite{isella2011s} is a proximity network of attendees at the ACM Hypertext 2009 conference. The dataset contains 113 nodes each corresponding to an attendee and 20,800 edges each corresponding to an interaction between two attendees.
    
\subsubsection{Static Datasets}
\label{subsec:non-temporal}
    
\textbf{Blog\footnote{\url{http://socialcomputing.asu.edu/datasets/BlogCatalog}}:} This dataset was collected from the Blog Catalog website. Bloggers follow other bloggers forming graph edges and they categorize their blog under some predefined classes. The graph used by \citet{liu2019streaming} has 5,196 nodes, 171,743 edges, and 6 types of node classes. Other works using this dataset include \cite{li2017attributed} and \cite{ma2018depthlgp}. 
    
\textbf{CiteSeer\footnote{\url{https://linqs.soe.ucsc.edu/node/236}}:} This is a citation network where papers are considered as nodes and citations are considered as the edges. The broad category of papers is used as the class labels. It has 3,312 nodes, 4,732 edges, and 6 types of node classes. This dataset is widely used including in \cite{liu2019streaming}, \cite{kipf2017semi} and \cite{liao2019lanczosnet}. 
    
\textbf{Flickr\footnote{\url{http://socialnetworks.mpi-sws.org/data-imc2007.html}} \footnote{\url{https://snap.stanford.edu/data/web-flickr.html}}:} The Flickr dataset~(\cite{Tang:2009:RLV:1557019.1557109}) is obtained from a network of users on a photo-sharing website. There are class labels that correspond to the groups that users have subscribed to on the website. The instance used by \citet{tang2015line} contains 1,715,256 nodes and 22,613,981 edges.

\begin{table}[t]
    \centering
    \tiny
    \begin{tabular}{r|l|c|c}
        Reference & Code & Nature & Type\\ \hline
        \citet{jin2019recurrent} & \url{https://github.com/INK-USC/RE-Net} & Dynamic & KG\\
        \citet{goel2020diachronic} & \url{https://github.com/borealisai/DE-SimplE} & Dynamic & KG\\
        \citet{wu2019efficiently} & \url{https://github.com/lienwc/DKGE/} & Dynamic & KG \\
        \citet{trivedi2017know} & \url{https://github.com/rstriv/Know-Evolve} & Dynamic & KG\\
        \citet{dasgupta2018hyte} & \url{https://github.com/malllabiisc/HyTE} & Dynamic & KG \\
        \citet{bian2019network} & \url{https://github.com/Change2vec/change2vec} & Dynamic & KG \\
        \citet{seo2018structured} & \url{https://github.com/youngjoo-epfl/gconvRNN} & Dynamic & G  \\
        \citet{wu2017recurrent} & \url{https://github.com/RuidongZ/Recurrent_Recommender_Networks} & Dynamic & G\\
        \citet{zhou2018dynamic} & \url{https://github.com/luckiezhou/DynamicTriad} & Dynamic & G\\
        \citet{zhang2018timers} &  \url{https://github.com/ZW-ZHANG/TIMERS} & Dynamic & G\\
        \citet{sajjad2019efficient} & \url{https://github.com/shps/incremental-representation-learning} & Dynamic & G\\
        \citet{kazemi2018simple} & \url{https://github.com/Mehran-k/SimplE} & Static & KG \\
        \citet{dong2017metapath2vec} & \url{https://ericdongyx.github.io/metapath2vec/m2v.html} & Static & KG\\
        \citet{zhang2018metagraph2vec} & \url{https://github.com/daokunzhang/MetaGraph2Vec} & Static & KG \\
        \citet{trouillon2016complex} & \url{https://github.com/ttrouill/ComplEx} & Static & KG \\
        \citet{lacroix2018canonical} & \url{https://github.com/facebookresearch/kbc} & Static & KG \\
        \citet{nickel2016holographic} & \url{https://github.com/mnick/holographic-embeddings} & Static & KG\\
        \citet{lerer2019pytorch} & \url{https://github.com/facebookresearch/PyTorch-BigGraph} & Static & KG \\
        \citet{kipf2017semi} & \url{https://github.com/tkipf/gcn} & Static & G\\
        \citet{hamilton2017inductive} & \url{https://github.com/williamleif/GraphSAGE} & Static & G \\
        \citet{chen2015net2net} & \url{https://github.com/soumith/net2net.torch} & Static & G \\
        \citet{grover2016node2vec} & \url{https://github.com/aditya-grover/node2vec} & Static & G \\
        \citet{velivckovic2018graph} & \url{https://github.com/PetarV-/GAT} & Static & G\\
        \citet{liao2019lanczosnet} & \url{https://github.com/lrjconan/LanczosNetwork} & Static & G 
        
    \end{tabular}
    \caption{Link to open-source software for static and dynamic (knowledge) graphs. ``G'' indicates that the code can handle only simple homogeneous graphs, whereas ``KG'' indicates that the code can handle knowledge graphs.}
    \label{tab:open-source-code}
\end{table}

\subsection{Open-Source Software}
There are several open-source libraries providing implementations for the papers discussed in the survey. \citet{pytgeo19} provide an implementation for many GCN-based papers (e.g., \cite{kipf2017semi,hamilton2017inductive}) in PyTorch~(\cite{pytpaszke2017automatic}). \citet{dgl18} also provide implementations for a variety of graph-based models. Along with graph-structured models, they also make tensor-based models like transformer~(\cite{vaswani2017attention}) available with the intention of facilitating the development of new models combining the two categories. They provide both PyTorch and MXNet~(\cite{chen2015mxnet}) backends for this library. Apart from these libraries, the implementation of several techniques covered in the survey is available in independent code repositories. Table~\ref{tab:open-source-code} provides links to where these implementations can be found. 

\section{Future Directions \& Conclusion}
\label{sec:conclusion}
A wide range of real-world problems can be formulated as reasoning over graphs (or networks). Traditionally, graph analytic methods have been mostly focused on static graphs, while in a large number of applications the graphs are dynamic and evolve. In the past few years, there has been a surge of works on dynamic graphs. In this paper, we surveyed the recent approaches for representation learning over dynamic graphs. We described these approaches according to an encoder-decoder framework, a framework that has gained popularity within several communities.

Our survey sheds light on several ways in which learning from and reasoning with dynamic graphs can be done. Here, we mention some directions to improve learning and reasoning with dynamic graphs.

\begin{itemize}
    \item Current representation learning algorithms have been mostly designed for discrete-time dynamic graphs (DTDGs), with only a few works on learning from continuous-time dynamic graphs (CTDGs). Even the few existing works for CTDGs are quite limited in the types of observations they can handle as they mainly handle the addition of new edges. A promising direction for future research is to extend the existing models for representation learning over CTDGs, or develop new ones, to deal with other types of observations such as edge deletion, node addition, node deletion, node splitting, node merging, etc. \citet{ma2018depthlgp} take some initial steps towards handling node addition, but their proposal provides an embedding for a new node considering only the current state of the graph (not the evolution of the graph).
    \item While some of the existing encoders work with certain types of graphs, it is not trivial how they can be used for other types of graphs. For instance, random walk approaches have been mainly designed for non-attributed graphs. Extending these approaches to the case of attributed graphs is not straightforward. The same goes for autoencoder-based encoders where it is not trivial how these approaches can be extended to KGs, and several other models discussed in the survey. An interesting direction for future research would be to extend these models to be applicable to more types of graphs.
    \item The approaches for CTDGs that can be used for the streaming scenario are mainly based on RNNs. In other sequence modeling domains (e.g., natural language processing), new sequence modeling approaches have been developed some times showing superior performance compared to RNNs. One example is the transformer architecture~(\cite{vaswani2017attention}) where recurrence has been replaced with self-attention. Designing new models for the streaming scenario in CTDGs based on self-attentions is a promising direction for future research. Other possibilities include designing new models based on neural ordinary differential equations (see, e.g., \cite{chen2018neural}).
    \item In a CTDG, many observations may be made at the same time. For instance, in an email communications network, a sender may send an email to many receivers at the same time. Current RNN-based encoders consider a random ordering for such observations. This naive approach may hinder an RNN from learning to generalize to other possible orderings of these simultaneous observations. A future direction would be to extend RNN approaches for dealing with simultaneous observations.
    \item While there exists a body of research on classifying static graphs and some of them may be (to some extent) applicable to dynamic graphs, the literature on classifying dynamic graphs is still at its infancy. When the classification of a dynamic graph is required (e.g., for activity recognition from videos), current approaches often convert the dynamic graph into a static graph and then run a graph classification algorithm on the static graph. Designing dynamic graph classification models is an interesting direction for future research.
    \item Expressiveness is an important property to be taken into account when selecting/designing a model. A model that is not expressive enough is doomed to underfitting at least for some applications. While a few recent works study the expressiveness of some models for (knowledge) graphs~(\cite{trouillon2017knowledge,kazemi2018simple,xu2019powerful,morris2019weisfeiler}), a more detailed and in-depth study of the expressiveness and its empirical impact may be a promising direction for future research.
    \item The existing models for dynamic graphs only consider edges connecting two nodes in the graph. However, in real-world applications, some edges may connect more than two nodes. These edges are known as hyperedges. In a KG, for instance, an edge corresponding to a \emph{purchase} relation may connect a person as the \emph{buyer} to another person as the \emph{seller} and also to an item that is being purchased. \citet{kazemi2018representing} argues that representation learning algorithms may fail if these hyperedges are converted into (several) binary edges through reifying new entities as, during test time, an embedding does not exist for the reified entities. Some recent works study ways of handling such hyperedges for static (knowledge) graphs~(\cite{wen2016representation,feng2018hypergraph,yadati2018hypergcn,bai2019hypergraph,fatemi2019knowledge}). A future research direction would be to develop models for dynamic (knowledge) graphs that are capable of handling hyperedges.
    \item Recently published papers on modeling dynamic graphs are each tested on different datasets, making it difficult to compare these models. It would be quite helpful to create some standard benchmarks with train, validation, and test splits so future models can be tested on the same benchmarks and splits.
\end{itemize}

\vskip 0.2in
\bibliography{MyBib}

\end{document}